%% file: PhDThesis_PedroMachado.tex
\documentclass[oneside,12pt]{Classes/customLaTeX}

\usepackage{amsmath}
\usepackage{amsfonts}
\usepackage[ruled,vlined,linesnumbered,noresetcount,algo2e]{algorithm2e}
\usepackage{algorithm}
\usepackage{graphicx,wrapfig}
\usepackage{algorithmic}
\usepackage{multirow}
\usepackage{colortbl}
\usepackage{color}
\usepackage[table]{xcolor}
\usepackage{epigraph}
\usepackage{caption}
\usepackage{subcaption}
\PassOptionsToPackage{hyphens}{url}\usepackage{hyperref}
\usepackage{tabularx}
\usepackage{float}
\restylefloat{table}
\usepackage{longtable}
\usepackage{pdfpages}
\usepackage{pdflscape}
\usepackage[acronym,toc]{glossaries}
\usepackage{setspace}
\usepackage{color,soul}
\usepackage{enumerate}
\usepackage{tablefootnote}
\usepackage{todonotes}
\usepackage{enumitem}
\usepackage{lscape}
\hypersetup{
 colorlinks=true,
 linkcolor=blue,
 filecolor=blue, 
 urlcolor=blue,
 citecolor=blue,
}
\SetKwInput{KwConst}{Constants}
\SetKwInput{KwAlg}{Main Algorithm}

\SetKwInput{KwConst}{Constants}
\SetKwInput{KwAlg}{Main Algorithm}
\SetKwInput{KwCircuits}{Parallel Circuits}

\newcommand{\SWITCH}[1]{\STATE \textbf{switch} (#1)}
\newcommand{\ENDSWITCH}{\STATE \textbf{end switch}}
\newcommand{\CASE}[1]{\STATE \textbf{case} #1\textbf{:} \begin{ALC@g}}
\newcommand{\ENDCASE}{\end{ALC@g}}

\ifpdf
 \pdfcatalog {/PageMode (/UseOutlines)
 /OpenAction (fitbh) }
\fi

\title{\Large{Computational models of object motion detectors accelerated using FPGA technology}}

\author{Pedro Miguel Baptista Machado}
\collegeordept{Department of Computer Science}
\school{School of Science and Technology}
\university{Nottingham Trent University}
\logo{\includegraphics[width=30mm]{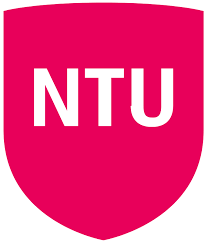}}

\renewcommand{\submittedtext}{A thesis submitted in partial fulfilment of the requirements of \\The Nottingham Trent University
for the degree of Doctor of Philosophy}

\degreedate{July, 2021}

\loadglsentries{glossary}
\makeglossaries
\begin{document}

% A page with the abstract and running title and author etc may be
% required to be handed in separately. If this is not so, comment
% the following 3 lines:
% \begin{abstractseparate}
% \input{Abstract/abstract}
% \end{abstractseparate}
\begin{spacing}{1}
\maketitle
\end{spacing}

% add an empty page after title page
% \newpage\null\thispagestyle{empty}\newpage

The copyright in this work is held by the author. You may copy up to 5\% of this work for private study, or personal, non-commercial research. Any re-use of the information contained within this document should be fully referenced, quoting the author, title, university, degree level and pagination. Queries or requests for any other use, or if a more substantial copy is required, should be directed to the author.

% set the number of sectioning levels that get number and appear in the contents
\setcounter{secnumdepth}{3}
\setcounter{tocdepth}{3}

\frontmatter
\include{Dedication/dedication}
\include{Acknowledgement/acknowledgement}
\include{Abstract/abstract}

\tableofcontents
\listoffigures
\listoftables
\printglossary[title=List of Acronyms,type=\acronymtype]
%\printglossary % Print the nomenclature (WAY TOO COMPLEX FOR ME NOW!)
\addcontentsline{toc}{chapter}{Nomenclature}

\mainmatter
\include{Chapter1/chapter1}
\include{Chapter2/chapter2}

\include{Chapter3/chapter3}
\include{Chapter4/chapter4}

\include{Chapter5/chapter5}
\include{Chapter6/chapter6}

\chapter*{Appendix}

The \gls{mhsnn} architecture presented in Chapter~\ref{Ch3:mhsnn} was published on the Conference proceedings of the International Joint Conferences on Neural Networks 2018 \cite{Machado_2018}.\\
P. Machado, A. Oikonomou, G. Cosma and T. M. McGinnity, "Bio-Inspired Ganglion Cell Models for Detecting Horizontal and Vertical Movements" 2018 \gls{ijcnn}, 2018, pp. 1-8, doi: 10.1109/IJCNN.2018.8489439.\\

The \gls{hsmd} algorithm presented in Chapter~\ref{Ch4:object_motion_detection_cells} was published on the \gls{ieee} Access \cite{Machado2021}.
P. Machado, A. Oikonomou, J. F. Ferreira and T. M. McGinnity, "HSMD: An object motion detection algorithm using a Hybrid Spiking Neural Network Architecture," in \gls{ieee} Access, doi: 10.1109/ACCESS.2021.3111005.\\

The \gls{neurohsmd} algorithm presented in Chapter~\ref{Ch5:neuromorphic_object_motion_detector} was submitted to the ACM Transactions on Reconfigurable Technology and Systems.

\bibliographystyle{ieeetr}
\renewcommand{\bibname}{References}

\bibliography{References/references} 
\addcontentsline{toc}{chapter}{References} 
\end{document}

%% file: glossary.tex
\newacronym{mhsnn}{MHSNN}{multi-hierarchical spiking neural network}
\newacronym{resume}{ReSuMe}{Remote Supervised Method}
\newacronym{stdp}{STDP}{Spike Timing Dependent Plasticity}
 
\newacronym{hsmd}{HSMD}{Hybrid Sensitive Motion Detector}
\newacronym{neurohsmd}{NeuroHSMD}{Neuromorphic Hybrid Sensitive Motion Detector}
\newacronym{ijcnn}{IJCNN}{International Joint Conference on Neural Networks}

\newacronym{fps}{fps}{Frames Per Second}

\newacronym{cdnet2012}{CDnet2012}{2012 Change Detection}
\newacronym{cdnet2014}{CDnet2014}{2014 Change Detection}

\newacronym{adas}{ADAS}{Advanced-Driver Assist Systems}
\newacronym{ns}{NS}{Navigational Systems}
\newacronym{cots}{COTS}{Commercial-Off-The-Shelf}

\newacronym{bs}{BS}{Background subtraction}
\newacronym{dbs}{DBS}{Dynamic Background subtraction}
\newacronym{roi}{ROI}{Region of Interest}
\newacronym{ptz}{PTZ}{Pan, Tilt and Zoom}

\newacronym{tp}{TP}{True Positives}
\newacronym{tn}{TN}{True Negatives}
\newacronym{fp}{FP}{False Positives}
\newacronym{fn}{FN}{False Negatives}
\newacronym{nmet}{nMet}{Number of Metrics}
\newacronym{pcc}{PCC}{Percentage of Correct Classifications}
\newacronym{pwc}{PWC}{Percentage of Wrong Classifications}

\newacronym{re}{Re}{Recall}
\newacronym{sp}{Sp}{Specificity}
\newacronym{fpr}{FPR}{False Positive Rate}
\newacronym{fnr}{FNR}{False Negative Rate}
\newacronym{wcr}{WCR}{Wrong Classifications Rate}
\newacronym{ccr}{CCR}{Correct Classifications Rate}
\newacronym{pr}{Pr}{Precision}
\newacronym{f1}{F1}{F-measure}
\newacronym{r}{R}{Average Ranking}
\newacronym{arc}{RC}{Average Ranking across all Categories}

\newacronym{ml}{ML}{Machine Learning}
\newacronym{rf}{RF}{Receptive Field}
\newacronym{hmi}{HMI}{Human Machine Interaction}
\newacronym{pca}{PCA}{Principal Component Analysis}
\newacronym{ica}{ICA}{Independent Component Analysis}
\newacronym{rpca}{RPCA}{Robust Principal Component Analysis}
\newacronym{ram-nn}{RAM-NN}{Random Access Memory Neural Network}
\newacronym{ram}{RAM}{Random Access Memory}
\newacronym{rbf-nn}{RBF-NN}{Radial Basis Function Neural Network}
\newacronym{sobs}{SOBS}{Self Organizing Background Subtraction}
\newacronym{hsv}{HSV}{Hue, Saturation and brightness Value}
\newacronym{dnn}{DNN}{Deep Neural Network}
\newacronym{rbm}{RBM}{Restricted Boltzman Machine}
\newacronym{sdae}{SDAE}{Stacked Denoising Auto-Encoder}
\newacronym{nerem}{NeREM}{Neural Response Mixture}
\newacronym{hog}{HOG}{Histogram of Oriented Gradients}
\newacronym{gloh}{GLOH}{Gradient Location and Orientation Histogram}
\newacronym{surf}{SURF}{Speeded Up Robust Features}
\newacronym{sift}{SIFT}{Scale-Invariant Feature Transform}
\newacronym{brief}{BRIEF}{Binary Robust Independent Elementary Features}
\newacronym{orb}{ORB}{Oriented FAST and Rotated BRIEF}
\newacronym{capoa}{CAPOA}{Content-Adaptive Progressive Occlusion analysis}
\newacronym{dlt}{DLT}{Deep Learning Tracker}
\newacronym{codd}{CODD}{Custom object direction detection}
\newacronym{stl}{STL}{Standard Template Library}

\newacronym{dsgc}{DSGC}{Direction-Selective Ganglion Cells}
\newacronym{ann}{ANN}{Artificial Neural Network}
\newacronym{rsnn}{RSNN}{Recurrent Spiking Neural Network}
\newacronym{cnn}{CNN}{Convolutional Neural Network}
\newacronym{rbcnn}{RBCNN}{Region Based Convolutional Neural Network}
\newacronym{csnn}{CSNN}{Convolutional Spiking Neural Network}
\newacronym{noc}{NoC}{Network-on-Chip}
\newacronym{rtss}{RTSS}{Real-Time Semantic Segmentation}
\newacronym{sbs}{SBS}{Semantic Background Segmentation}
\newacronym{mfcn}{MFCN}{Multiscale Fully Convolutional Network}

\newacronym{nha}{NHA}{NeuronHSMD Host Application}
\newacronym{ndk}{NDK}{NeuronHSMD device kernels}
\newacronym{sdk}{SDK}{Software Development Kit}

\newacronym{snn}{SNN}{Spiking Neural Network}

\newacronym{gsoc}{GSOC}{Google Summer of Code}
\newacronym{mog}{MOG}{Mixture of Gaussians}
\newacronym{mog2}{MOG2}{Gaussian Mixture Probability Density}
\newacronym{knn}{KNN}{Mixture of Gaussians K Nearest Neighbours}
\newacronym{lsbp}{LSBP}{Local Single Value Decomposition Binary Pattern}
\newacronym{ieee}{IEEE}{Institute of Electrical and Electronics Engineers}

\newacronym{cv}{CV}{Computer Vision}
\newacronym{opencv}{OpenCV}{Open Computer Vision}
\newacronym{opencl}{OpenCL}{Open Computer Language}
\newacronym{hdl}{HDL}{Hardware Description Language}
\newacronym{hls}{HLS}{High Level Synthesis}
\newacronym{rtl}{RTL}{Register Transfer Level}
\newacronym{uav}{UAV}{Unmanned Aerial Vehicle}
\newacronym{ipl}{IPL}{Inner Plexiform Layer}
\newacronym{sac}{SAC}{starburst amacrine cells}

\newacronym{nn}{NN}{Neural Network}
\newacronym{oms-gc}{OMS-GC}{Object Motion Sensitive Ganglion Cells}
\newacronym{omd}{OMD}{Object Motion Detection}
\newacronym{gc}{GC}{Ganglion Cells}
\newacronym{nuc}{NUC}{Next Unit of Computing}

\newacronym{sram}{SRAM}{Static Random Access Memory}
\newacronym{sdram}{SDRAM}{Synchronous Dynamic Random Access Memory}
\newacronym{vhsic}{VHSIC}{Very High Speed Integrated Circuit}
\newacronym{vhdl}{VHDL}{Very High Speed Integrated Circuit Hardware Description Language}
\newacronym{asic}{ASIC}{Application-Specific Integrated Circuits}
\newacronym{vlsi}{VLSI}{Very Large Scale of Integration}
\newacronym{dvs}{DVS}{Dynamic Voltage Scaling}
\newacronym{aer}{AER}{Address Event Representation}
\newacronym{davis}{DAVIS}{Dynamic Active Pixel Vision Sensor}
\newacronym{nna}{NNA}{Neural Network Accelerators}

\newacronym{fpga}{FPGA}{Field-Programmable Gate Array}
\newacronym{gpu}{GPU}{Graphical Processing Unit}
\newacronym{cpu}{CPU}{Central Processing Unit}
\newacronym{ic}{IC}{Integrated Circuit}
\newacronym{os}{OS}{Operating Systems}

\newacronym{cpg}{CPG}{Central Pattern Generator}

\newacronym{lif}{LIF}{Leaky-Integrate-and-Fire}
\newacronym{iaf}{IAF}{Integrate-and-Fire}
\newacronym{hh}{HH}{Hodgkin \& Huxley}
\newacronym{rc}{RC}{Resitance and Capacitance}
\newacronym{izk}{IZK}{Izhikevich}
\newacronym{ai}{AI}{Artificial Intelligence}
\newacronym{dog}{DoG}{Difference-of-Gaussians}
\newacronym{rgb}{RGB}{Red, Green and Blue}
\newacronym{fir}{FIR}{Finite Impulse Response}
\newacronym{iir}{IIR}{Infinite Impulse Response}

\newacronym{lab}{LAB}{logic array blocks}
\newacronym{mlab}{MLAB}{memory logic array blocks}
\newacronym{alut}{ALU}{adaptive look-up-table}
\newacronym{alm}{ALM}{adaptive logic module}
\newacronym{dsp}{DSP}{digital signal processing}

\newacronym{hdd}{HDD}{Hard Disk Drive}
\newacronym{ssd}{SSD}{Solid State Drive}
\newacronym{usb}{USB}{Universal Serial Bus}
\newacronym{sata}{SATA}{Serial Advanced Technology Attachment}
\newacronym{pcie}{PCIe}{Peripheral Component Interconnect Expres}
\newacronym{cuda}{CUDA}{Compute Unified Device Architecture}
\newacronym{ps}{PS}{Processor System}

\newacronym{som}{SOM}{Self Organizing Map}
\newacronym{bsp}{BSP}{Board Support Package}
\newacronym{ode}{ODE}{Ordinary Differential Equation}

%% file: Dedication/dedication.tex
%%%%%%%%%%%%%%%%%%%%%%%%%%%%%%%%%%%%%%%%%%%%%%%%%%%%%%%%%%%%%%%%%%%%%%%%%%%%%%%%
%2345678901234567890123456789012345678901234567890123456789012345678901234567890
%        1         2         3         4         5         6         7         8
% THESIS DEDICATION

\begin{dedication}

% quote
\textit{I would like to dedicate this PhD thesis to my wife Laura, children Elisa and Ana, my parents, sister and my grandmother Helena Machado. Laura has been by my side, giving me support and encouragement during this long PhD journey. I am truly thankful for having you in my life, and please remember that this PhD degree is also yours. Elisa, I tried my best to be there and play with you while you were growing up. Ana, thank you for all the support you gave me and for stepping up and helping mom looking after Elisa when I was working on my PhD.\\ Obrigado m\~ae pelo apoio e for\c{c}a que sempre me deste. \`{A} mem\'{o}ria da minha av\'{o} que sempre me apoiou no meu percurso acad\'{e}mico.}

\end{dedication}

% ----------------------------------------------------------------------

%%% Local Variables: 
%%% mode: latex
%%% TeX-master: "../thesis"
%%% End: 

%% file: Acknowledgement/acknowledgement.tex
%%%%%%%%%%%%%%%%%%%%%%%%%%%%%%%%%%%%%%%%%%%%%%%%%%%%%%%%%%%%%%%%%%%%%%%%%%%%%%%%
%2345678901234567890123456789012345678901234567890123456789012345678901234567890
%        1         2         3         4         5         6         7         8
% THESIS ACKNOWLEDGEMENTS

% Use the following style if the acknowledgements are long:
%\begin{acknowledgementslong}
%\end{acknowledgmentslong}

\begin{acknowledgements}

%“I don't know half of you half as well as I should like; and I like less than half of you half as well as you deserve.”
 %J.R.R. Tolkien, The Fellowship of the Ring 

I want to thank my supervisory team and independent assessor; Professor Martin McGinnity, for his immeasurable wisdom, being my mentor, extended support, encouragement and invaluable guidance during the PhD journey; Dr Andreas Oikonomou for being there to support me when I needed the most and for the long, very productive brainstorms and guidance; Professor Eiman Kanjo for her support and guidance, Dr Jo\~ao Filipe Ferreira who was my $2^{nd}$ Supervisor; Professor Ahmad Lotfi for his support, encouragement and guidance.\\
\bigskip
I also want to thank my PhD thesis' non-official reviewers and friends; Dr Robert Ranson (mentor and friend), Dr John Wade (long-time friend), Dr Farhad Fassihi Tash, Dr (to be) Francisco Lemos (lifetime friend), Dr Lorenzo Ferrara, Dr Isibor Kennedy Ihianle and Samuel Brandenburg. Their feedback and suggestions were crucial to finalise the PhD thesis. \\
\bigskip
A big thanks to my colleagues and friends; \textit{Dr David Adama, Dr Kayode Owa, Dr Salisu Yahaya, Dr Kofi Appiah, Dr Nikesh Lama, Dr Peter FitzGerald, Dr Neil Sculthorpe, Dr Filipe Neves dos Santos, Dr Micael Couceiro, Dr David Portugal, Dr Raquel Santos (close friend), Dr Alexey Petrushin, Jorge Ros\'ario (close friend), Lu\'is Costa (lifetime friend and mentor), Nuno Semedo (lifetime friend), Jo\~ao Reis (lifetime friend), Hugo Faria (lifetime friend) and Bruno Santos (lifetime friend)} and many other names that were not enumerated here. Each one of them have contributed in their way to help me to reach this important milestone.

\end{acknowledgements}

%% file: Abstract/abstract.tex
%%%%%%%%%%%%%%%%%%%%%%%%%%%%%%%%%%%%%%%%%%%%%%%%%%%%%%%%%%%%%%%%%%%%%%%%%%%%%%%%
%2345678901234567890123456789012345678901234567890123456789012345678901234567890
%        1         2         3         4         5         6         7         8
% THESIS ABSTRACT

% Use the following style if the abstract is long:
%\begin{abstractslong}
%\end{abstractslong}

\begin{abstractslong}
The detection of moving objects is a trivial task when performed by vertebrate retinas, yet a complex computer vision task. This PhD research programme has made three key contributions, namely: 1) a \gls{mhsnn} architecture for detecting horizontal and vertical movements, 2) a \gls{hsmd} algorithm for detecting object motion and 3) the \gls{neurohsmd} , a real-time neuromorphic implementation of the \gls{hsmd}  algorithm.

The \gls{mhsnn} is a customised 4 layers \gls{snn} architecture designed to reflect the basic connectivity, similar to canonical behaviours found in the majority of vertebrate retinas (including human retinas). The architecture, was trained using images from a custom dataset generated in laboratory settings. Simulation results revealed that each cell model is sensitive to vertical and horizontal movements, with a detection error of 6.75\% contrasted against the teaching signals (expected output signals) used to train the \gls{mhsnn}. The experimental evaluation of the methodology shows that the \gls{mhsnn} was not scalable because of the overall number of neurons and synapses which lead to the development of the \gls{hsmd}.

The \gls{hsmd}  algorithm enhanced an existing \gls{dbs} algorithm using a customised 3-layer \gls{snn}. The customised 3-layer \gls{snn} was used to stabilise the foreground information of moving objects in the scene, which improves the object motion detection. The algorithm was compared against existing background subtraction approaches, available on the \gls{opencv} library, specifically on the \gls{cdnet2012} and the \gls{cdnet2014} benchmark datasets. The accuracy results show that the \gls{hsmd} was ranked overall first and performed better than all the other benchmarked algorithms on four of the categories, across all eight test metrics. Furthermore, the \gls{hsmd} is the first to use an \gls{snn} to enhance the existing dynamic background subtraction algorithm without a substantial degradation of the frame rate, being capable of processing images $720\times480$ at 13.82 \gls{fps} (\gls{cdnet2014}) and $720\times480$ at 13.92 \gls{fps} (\gls{cdnet2012}) on a High Performance computer (96 cores and 756 GB of RAM). Although the \gls{hsmd} analysis shows good \gls{pcc} on the \gls{cdnet2012} and \gls{cdnet2014}, it was identified that the 3-layer customised \gls{snn} was the bottleneck, in terms of speed, and could be improved using dedicated hardware.

The \gls{neurohsmd} is thus an adaptation of the \gls{hsmd} algorithm whereby the SNN component has been fully implemented on dedicated hardware [Terasic DE10-pro \gls{fpga} board]. \gls{opencl} was used to simplify the \gls{fpga} design flow and allow the code portability to other devices such as \gls{fpga} and \gls{gpu}. The \gls{neurohsmd} was also tested against the \gls{cdnet2012} and \gls{cdnet2014} datasets with an acceleration of 82\% over the \gls{hsmd} algorithm, being capable of processing $720 \times 480$ images at 28.06 \gls{fps} (\gls{cdnet2012}) and 28.71 \gls{fps} (\gls{cdnet2014}).

\end{abstractslong}

%% file: Chapter1/chapter1.tex
%%%%%%%%%%%%%%%%%%%%%%%%%%%%%%%%%%%%%%%%%%%%%%%%%%%%%%%%%%%%%%%%%%%%%%%%%%%%%%%%
%2345678901234567890123456789012345678901234567890123456789012345678901234567890
%        1         2         3         4         5         6         7         8
% THESIS INTRODUCTION

\chapter{Introduction}
\label{Ch1:introduction}
\ifpdf
    \graphicspath{{Introduction/Figures/PNG/}{Introduction/Figures/PDF/}{Introduction/Figures/}}
\else
    \graphicspath{{Introduction/Figures/EPS/}{Introduction/Figures/}}
\fi

% quote

%\setlength{\epigraphwidth}{.35\textwidth}
%\epigraph{Research is formalized curiosity.}{ Zora Neale Hurston, 1942}

% examples of sections

\section{Background} \label{Ch1.1:background}
In Computer Vision, object motion detection is the process of detecting moving objects relative to their surroundings \cite{Chapel2020}. Object motion detection is required in many real applications such as video surveillance of human activities, monitoring of animals, optical motion detection, multimedia application, \gls{adas}, and autonomous systems \cite{Chapel2020,Garcia-Garcia2020}. There are several challenges (e.g. objects camouflage, illumination variation, motion blur) when performing object motion detection. Several object motion detection methods have been published over the years \cite{Chapel2020,Garcia-Garcia2020}, some methods are more accurate than others but are also more computationally intensive, while others are less accurate and computationally intensive. The majority of the object motion detection methods only focus on the accuracy of proposed methods solving a group of challenges and do not measure the speed \cite{Chapel2020,Garcia-Garcia2020}. Nevertheless, there is a significant demand in terms of real-time robust object motion detection is required in applications such as autonomous systems' navigation, object tracking and changes detection \cite{Garcia-Garcia2020}.\\

The detection of moving objects in different directions and the motion detection of objects are done very efficiently by vertebrate retinas. Different retinal circuits trigger different functionalities such as light detection, motion detection and discrimination (i.e. interpretation of spatio-temporal patterns triggered by the retinal photoreceptors), object motion (i.e. detection of objects moving in the scene), identification of approaching motion (looming), anticipation, motion extrapolation, and omitted stimulus-response \cite{Gollisch2010}. Vertebrate retinas are notable for i) incorporating millions of these retinal circuits, ii) being extremely efficient (the whole human brain consumes approximately 20 Watts) and iii) currently still displaying the capability to outperform any state-of-the-art computer \cite{TrujilloHerrera2020}. \\

\section{Current approaches} \label{Ch1.2:current_approaches}
The human brain is very efficient at performing computations, as it only takes about 25 watts for 86 billion neurons \cite{Bouvier2019}. The brain computational efficiency is a consequence of massive parallelism. The retina, an extension of the brain itself, is responsible for performing the first stages of visual pre-processing, including the detection of movement \cite{Kolb2003,Gollisch2010}. Although the detection of the directions described by moving objects and object motion detection are trivial tasks for vertebrate retinas \cite{Kolb2003,Gollisch2010}, these are still complex computational tasks using exiting \acrfull{bs} methods \cite{Chapel2020}.

On computers, the detection of movements is normally achieved through the use of \gls{bs} methods. \gls{bs} have been one of the most active research topics in computer vision and have been widely studied for the last 30 years \cite{Chapel2020,Bouwmans2019,Illahi2020}. \cite{Chapel2020,Bouwmans2019,Illahi2020}. In \gls{bs} algorithms, object motion detection is obtained through the extraction of the foreground (composed of moving objects) from the background (composed of static or semi-static movements) image. \gls{bs} are used in applications such as intelligent surveillance of human activities in public spaces, traffic monitoring, industrial machine vision applications, etc \cite{Bouwmans2019}. More recently, \gls{bs} algorithms have been improved using \gls{ml} algorithms  \cite{Chapel2020}. \gls{ml}  is a research field that focuses on the development of algorithms and methods for solving complex problems in a generic way \cite{Rebala2019}. These algorithms are characterised by learning the detailed design from a set of labelled data and learn a model or a set of rules from a labelled dataset so that the \gls{ml} model can correctly predict the labels of data points in unforeseen datasets (i.e. datasets that were not used to train the model) \cite{Rebala2019}. Among other research topics, \gls{ml} includes \glspl{ann},  \glspl{cnn} and \acrfullpl{snn} \cite{Bouvier2019,Illahi2020}. 

\glspl{ann} have been widely used for classification and pattern recognition tasks, but at the same time, \glspl{ann} lack biological plausibility \cite{Bouvier2019}. Although \glspl{cnn} are as recent as \glspl{ann}, their popularity has increased in recent years as a consequence of the evolution of the computational capabilities \cite{Bouwmans2019,Illahi2020}. \glspl{cnn} are known by its capability to accurately classify objects and can be used for tracking of known objects but similar to \glspl{ann} ,they lack of biological plausibility \cite{Bouwmans2019,Illahi2020}. Unlike \glspl{ann} and \glspl{cnn} that have been studied for more than thirty years, \glspl{snn} emerged about twenty years ago and have gained growing interest in researchers because of their biological plausibility. \cite{Bouvier2019}.\\

\glspl{snn} are well known for their biological plausibility, but also for the complexity inherited from biological systems, which are characterised by massive parallelism, according to \cite{Bouvier2019}. Modern computation platforms rely heavily on \gls{cpu} to provide compatibility and security with other devices/applications. Although \glspl{cpu} have been evolving in recent years, \glspl{cpu} still relies on the von Neumann architecture proposed by John von Neumann proposed in 1945 \cite{von-neurmann1945,zhang2018}. Blank \cite{Blank2018} wrote in 2018 that Moore's law, which states that the number of transistors in dense \glspl{ic} doubles every eighteen to twenty-four months, ended around 2008. Furthermore, the clock speed has reached technological limitations, preventing \glspl{cpu} from working with frequencies above 4 GHz, which also introduces memory barriers, and power dissipation challenges \cite{Blank2018}. Therefore, the design of the \gls{cpu} paradigm has shifted into multicore and multiprocessor to overcome the limitations associated with the clock speed \cite{zhang2018}. The scientific community agrees that the multicore and multiprocessor strategy will shortly meet technological limitations related to the increase in power consumption of these solutions. \cite{zhang2018}.\\

Neuromorphic engineering aims to develop hardware devices/platforms that mimic biologic nervous systems. Such neuromorphic solutions must be compatible with other devices and applications, which is a challenge because compatibility and security are provided via standard  \gls{os} (e.g. Microsoft Windows, macOS and Linux) that only run on \glspl{cpu}. The solution is heterogeneous computing, which combines \glspl{cpu} with one or more processing technologies such as graphical processing units \glspl{gpu}, \glspl{fpga}, or other \glspl{asic} connected via high-speed external (e.g. PCIe) or internal (e.g. AXI and Avalon) buses \cite{Intel2021, Xilinx2011}. Heterogeneous computing and/or neuromorphic applications are ideal for hosting customised \gls{snn}.\\

\section{Research gaps} \label{Ch1.3:gaps}
The detection of the direction performed by moving objects and the \gls{omd} are used in many fields, including traffic and human activity monitoring, \gls{adas}, object tracking, etc. Unlike computers, vertebrate retinas are highly efficient at sensing object motion and its direction. In this PhD research programme, the following gaps were identified:

\begin{itemize}
    \item Modern \gls{bs} and object motion detection algorithms use \gls{ml} approaches such as \glspl{ann} and \glspl{cnn} but are not suitable for real-time applications and lack biological plausibility.
    \item \gls{snn} are biologically plausible and can be used for implementing basic functionalities observed in retinal cells, such as in the \acrfullpl{dsgc} and \glspl{oms-gc}
    \item \gls{snn} are massively parallel and therefore not suitable for being processed by \glspl{cpu}
    \item \glspl{fpga} are specialised and flexible hardware devices that offer freely-reconfigurable logic, desirable for accelerating massively parallel algorithms
    \item Although \glspl{fpga} are normally reconfigured using \gls{hdl} which are hard to master, \gls{opencl} is a C-like programming language, it is simpler to master and more suitable for programming devices like \glspl{cpu} and \glspl{fpga}

\end{itemize}

This research was built on the premise of the aforementioned gaps identified.

\section{Aims and objectives} \label{Ch1.4:aims_objectives}
In challenging circumstances (e.g. low visibility, blurred vision, moving cameras with moving objects, occlusion, etc.), no \gls{bs} or \gls{omd} outperforms vertebrate retinas. Is it possible to model or enhance existing object motion detection algorithms? Can such methods be accelerated using dedicated algorithms? The PhD research programme aimed to explore how object motion detection could be improved using \gls{snn} to mimic some of the \glspl{dsgc} and \glspl{oms-gc} basic functionalities and accelerate \glspl{snn} using dedicated hardware.\\

The PhD research programme had the following objectives:

\begin{enumerate}[label=\alph*)]

\item Research into \glspl{dsgc} and \glspl{oms-gc} and replicate their basic functionalities (such as detecting horizontal and vertical movements and more generic object motions). The research lead to the development of the \gls{mhsnn} architecture of a customised 4-layer \gls{snn}, inspired on \gls{dsgc}  available on vertebrate retinas, which is sensitive to vertical and horizontal movements. The results show that the \gls{mhsnn} performed the correct detection of leftwards, rightwards, downwards, and upwards movements in 93.6\%, 92.4\%, 93.9\% and 93.1\%, respectively  when tested against the custom semisynthetic dataset.

\item Explore the use of spiking neural networks to model and/or enhance existing object motion detectors. The \gls{hsmd} algorithm combined an existing \gls{bs} algorithm named \acrshort{gsoc}, available on the \acrshort{opencl} library, with a customised 3-layer \gls{snn}. The \gls{hsmd} algorithm was ranked first overall against the \gls{cdnet2012} and \gls{cdnet2014} benchmark datasets against the competing \gls{bs} algorithm. Furthermore, the \gls{hsmd} did not introduce a substantial degradation of the frame rate, being capable of processing $720\times480$ at 13.82 \gls{fps} (\gls{cdnet2014}) and 13.92 \gls{fps} (\gls{cdnet2012}).

\item Optimise object motion detectors using \glspl{fpga} to improve the latency by accelerating \gls{snn} with minimal or no deterioration of the accuracy. The \gls{hsmd}'s customised 3-layer \gls{snn} {was accelerated on an} \gls{fpga} device using \gls{opencl} and was tested against the \gls{cdnet2012} and \gls{cdnet2014} benchmark datasets. The \gls{neurohsmd} has produced an acceleration of 82\% over the \gls{hsmd} and it is capable of processing $720\times480$ at 28.06 \gls{fps} (\gls{cdnet2012}) and 28.71 \gls{fps} (\gls{cdnet2014}).

\end{enumerate}

\section{Summary of the thesis} \label{Ch1.5:summary}
The PhD research programme focused on modelling a \gls{mhsnn} for detecting horizontal and vertical movements, design of \gls{snn} to enhance an existing and efficient \gls{bs} algorithm using a custom 3-layer \gls{snn} called \gls{hsmd} and \gls{neurohsmd}, which is the neuromorphic implementation of the \gls{hsmd} algorithm, for optimising the computation speed of the custom 3-layer \gls{snn}. The thesis structure and a summary of each Chapter are presented below:\\

\bigskip

\large\textbf{Chapter ~\ref{Ch2:lr}: Literature Review:}\\

This chapter reviews the relevant common aspects to most vertebrate retinas' physiology, emphasising the \glspl{oms-gc}. This chapter also reviews the most relevant \gls{snn} works, highlighting their main advantages and disadvantages. The most relevant \gls{bs} works, challenges, and limitations are also reviewed in this chapter. The Chapter ends with a revision of relevant neuromorphic implementations of retinas and other brain functions.\\

\bigskip

\large\textbf{Chapter~\ref{Ch3:mhsnn} - Initial exploration of Spiking Neural Networks on motion detection:}\\ The initial exploration of \gls{snn} lead to the development of a novel \gls{mhsnn} architecture capable of detecting leftwards, right-wards, upwards, and downwards movements. It is also explained how the supervised learning \gls{resume} was used to train the output layer neurons and how synapses with different propagation can be used to create local buffers. The \gls{mhsnn} is capable of classifying simple movements using a semisynthetic dataset.\\

\bigskip

\large\textbf{Chapter~\ref{Ch4:object_motion_detection_cells}: \gls{hsmd} - Hybrid Spiking Motion Detection:} \\

The \gls{hsmd} proposes the enhancement of the \gls{gsoc} \gls{bs} algorithm \cite{Samsonov2017} available on the \gls{opencv} library \cite{OpenCV2021}, and that has performed better on the \gls{cdnet2012} \cite{Goyette2012}, and \gls{cdnet2014} \cite{Wang2014} which are two of the reference datasets for benchmarking \gls{bs} algorithms using customised 3-layer \gls{snn}. The 3-layer \gls{snn} was optimised to ensure that the \gls{hsmd} algorithm could perform background subtraction on the fly. The \gls{hsmd} ranked first when benchmarked against all the \gls{bs} available on the \gls{opencv} library \cite{OpenCV2021} using the eight metrics proposed in \gls{cdnet2012} \cite{Goyette2012}.\\

\bigskip

\large\textbf{Chapter~\ref{Ch5:neuromorphic_object_motion_detector}: \gls{neurohsmd} - Neuromorphic Hybrid Spiking Motion Detection:} \\ A state-of-the-art \gls{neurohsmd} is presented in this chapter. The 3-layer \gls{snn} used to enhance the \gls{gsoc} algorithm, which was the bottleneck of the \gls{hsmd} algorithm, was fully implemented on an Intel \gls{fpga} board \cite{Intel2019} using OpenCL \cite{Kronos2021}.

\bigskip

\large\textbf{Chapter~\ref{Ch6:conclusions}: Conclusions and Future Work}\\

A summary of the work and achievements obtained during the PhD research programme is given in this chapter. A list of publications and future publications emerging from the results obtained in Chapter~\ref{Ch5:neuromorphic_object_motion_detector} is also presented in this chapter. The last section discusses future work and how the systematic methodology developed in this research programme can efficiently model other bio-inspired retinal cells.

%% file: Chapter2/chapter2.tex
%%%%%%%%%%%%%%%%%%%%%%%%%%%%%%%%%%%%%%%%%%%%%%%%%%%%%%%%%%%%%%%%%%%%%%%%%%%%%%%%
%2345678901234567890123456789012345678901234567890123456789012345678901234567890
% 1 2 3 4 5 6 7 8
% THESIS Chapter

\chapter{Background Research} \label{Ch2:lr}
This chapter reviews works relevant to the PhD research program. A brief introduction to the anatomy of the eye and retina is given in Section \ref{Ch2.1:biological} to set the context and to introduce the \glspl{oms-gc}, which are the foundations of the PhD research programme. Spiking neuron models and \gls{snn} are discussed and in Section~\ref{Ch2.2:snn}. Existing \gls{snn} simulators and their advantages/disadvantages are summarised in Section~\ref{Ch2.3:snn_simulators}. Relevant works in terms of \gls{snn} architectures for image processing are analysed in Section~\ref{Ch2.4:snn-arcitectures}. The reader will have the opportunity to understand the research gaps in terms of \gls{snn} architectures which justify the need for this PhD research programme. Challenges and use-case scenarios of object motion detection using classical computer vision methods are discussed in Sections~\ref{Ch2.5:object_motion_detection}. Neuromorphic and heterogeneous computing are analysed in Section~\ref{Ch2.6:hardware_implementations} covers the importance of hardware implementations in the acceleration of \glspl{snn}. The research gaps that led to this PhD research programme are discussed in section \ref{Ch2:revised-literature}.

\section{Anatomy of the Eye and Retina}\label{Ch2.1:biological}
\begin{figure}[H] 
 \centering
	\includegraphics[width=0.5\textwidth]{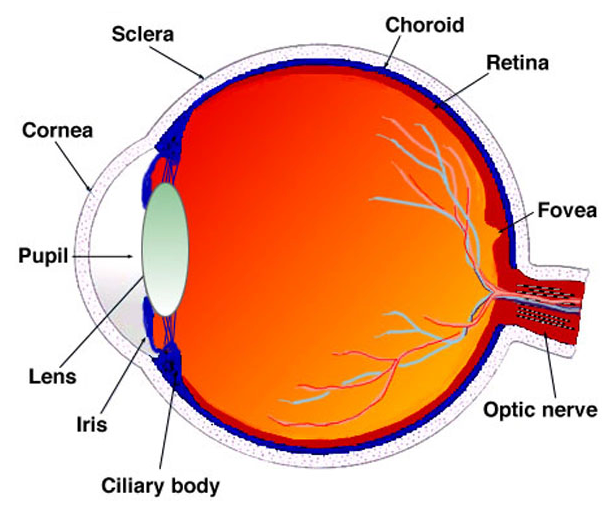}
	\caption{
 Anatomy of the eye. Adopted from \protect\cite{Kolb2011}. } \label{fig:eye}
\end{figure}

The eye (see Figure~\ref{fig:eye}) is a fundamental part of the vision system, and it is composed by the following parts: \textbf{Iris} which is the coloured part of the eye that is responsible for controlling the amount of light that reaches the retina that lays at the back of the eye; \textbf{Pupil} is a circular membrane in the centre of the Iris that dilates and contracts to increase or reduce the quantity of light that reaches the retina; \textbf{Cornea} is the transparent circular membrane covering the front of the eyeball which refracts the light to the Lens;

\textbf{Lens} is the transparent structure behind the Pupil that refracts the light into the retina; \textbf{Choroid} is the middle layer of the eye between the retina and the Sclera that contains pigments to absorb excess light to prevent image blurring; \textbf{Ciliary body} connects the Choroid to the iris; \textbf{Sclera} is the white and though part of the eye for maintaining the spherical configuration of the eye and offers resistance to internal and external forces; \textbf{Retina} a tiny tissue that lays at the back of the eyeball and is composed of thousands of neurocircuits responsible for converting the light intensities into visual information to be further processed by the brain;\textbf{Fovea} is a tiny depression in the centre of the Macula (a yellow spot on the retina) and it is considered a highly-specialised region of the retina responsible for producing the sharpest vision with the highest colour discrimination; and \textbf{Optic nerve} which provides the multichannel connectivity between the retina and the brain.

\begin{figure}[H] 
 \centering
	\includegraphics[width=0.8\textwidth]{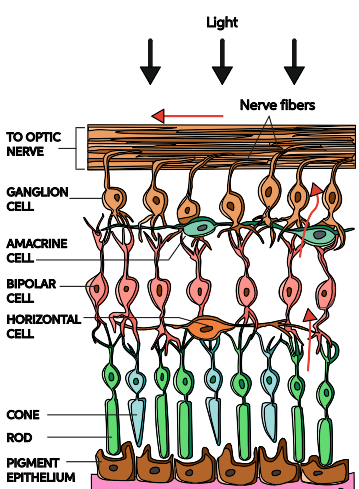}
	\caption{
The structure of the retina comprises rods, cones, and horizontal, Amacrine, and ganglion cells. Light passes through all the retinal layers and hits the pigment epithelium responsible for protecting the outer retina from excessive light. Rods detect grey gradients and dim light, while cones detect red, green, and blue light. Horizontal cells regulate the visual information from rods, cones, and bipolar cells. Bipolar cells transport the visual information to ganglion cells. Amacrine cells regulate the visual information from bipolar and ganglion cells. Finally, ganglion cells receive and process the visual information from bipolar cells and transmit the post-processing information to the visual cortex via the optic nerve. Adopted from \protect\cite{Chegg2021}. } \label{fig:retina}

\end{figure}

The retina is considered an extension of the brain and has been widely studied since Cajal (1892) \citep{Cajal1892}, \cite{Piccolino1988}. Vertebrate retinas vary in type, shape, size, connectivity, and number of cell types. All vertebrate retinas are composed of photoreceptors (cones and rods), bipolar, horizontal, amacrine, and ganglion cells (Figure~\ref{fig:retina}). Each of these types of cells comprises a wide range of functional subtypes \cite{Masland2001}.

Rods are sensitive to low intensity light, while cones sense light and colour. Cones can be subdivided into three categories based on their ability to detect short, medium, and long light wavelengths \cite{Bosze2020}. Bipolar cells forward signals triggered by rods and cones to ganglion cells \cite{Bosze2020}. Horizontal cells are responsible for processing visual information received from bipolar cells, rods, and cones \cite{Kolb2011, Bosze2020}. Amacrine cells are responsible for modulating and integrating visual signals from bipolar, and ganglion cells \cite{Kolb2011, Bosze2020}. Ganglion cells receive visual information from bipolar cells and analyse shapes, contrast, motion, and colour, and forward them to the visual cortex via the optic nerve \cite{Kolb2011, Bosze2020}.

Each retinal cell has a specific role in the vision system. The types of cells vary between animal species, and the number of retinal cells also varies according to the surrounding environment where each species lives. The organisation of the retina is shown in Figure~\ref{fig:retina}

Object motion detection, light sensitivity, and looming (i.e. object approaching the eye) detection are three of the visual tasks that are observed in all vertebrate retinas, such as the light sensitive and looming \gls{gc}. The \glspl{oms-gc} active respond when a local patch on the centre of receptive field moves with a trajectory different from the background; light sensitive \gls{gc} respond to light intensity variations; and the looming \gls{gc} respond to approach and recede motions \cite{Gollisch2010}. More complex cells have been identified, including the fast response and the predictive \gls{gc}. The fast response cells respond to fast motion variations, while the predictive cells trigger automatic responses to specific stimuli that were previously learnt \citep{Garvert2013}. The focus of this PhD research programme is to model and emulate the \gls{oms-gc} basic functionalities for improving existing motion detection algorithms, targeting a wide range of environmental conditions from dim light to challenging weather conditions. \\

\section{Spiking Neuron Models} \label{Ch2.2:snn}

Models of the retina are generally either focused on single neuron cells, or on complex networks of neurons \cite{Guo2014,Roberts2016}. Advances in retina research \cite{Kolb2018,Gollisch2010, Vance2018,Fu2019,Kuhn2019} have enabled researchers to understand better the anatomical and neurophysiological retinal function to design computational models that mimic some retinal aspects. Retinal responses are related to action potential initiation, dendritic processing, many levels of effects (such as ionic channels, physical properties, extracellular stimulation), motion detection, and motion anticipation \cite{Guo2014}. Computational models have been used to augment the understanding of single neuron response dynamics and computation and their functional contributions in multi-hierarchy neural networks \cite{Guo2014}. Typically, retinal models are modelled with a high level of abstraction, focusing on the individual action potentials \cite{Huo2014}. The significant differences between approaches are the degree to which the authors can model the actual retina's behaviours \cite{Gerstner2002}.

\begin{figure} [H]
	\begin{center}
	\includegraphics[width=0.5\textwidth]{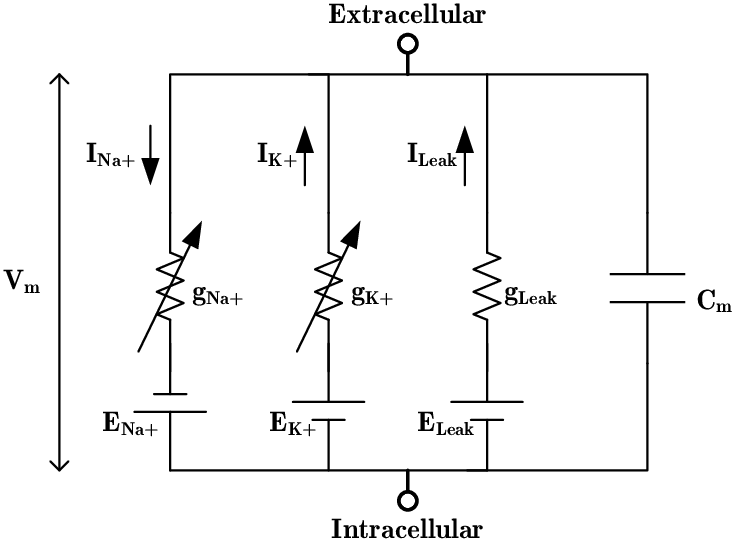}
	\caption{Schematic diagram of the Hodgkin-Huxley neuron model.$V_m$ is the membrane potential and $C_m$ is the membrane capacitance; $I_{Na^+}$, $I_{K^+}$ and $I_{Leak}$ are the currents associated to each channels. $C_m$ is the membrane potential; $g_{Na^+}$, $g_{K^+}$ are the non-linear electrical conductances that control the voltage-gated ion channels; $g_{Leak}$ is the linear conductance. $E_{Na^+}$, $E_{K^+}$ and $E_{Leak}$ are the equilibrium potentials. Adopted from \protect\citep{Hodgkin1952}. \label{fig:HH}}
	\end{center}
\end{figure}

In 1952, \gls{hh} \cite{Hodgkin1952} proposed the first biologically plausible and also the most computationally intensive spiking neuron model. The \gls{hh} model consists of a semipermeable cell membrane that splits the internal cell, that behaves as a capacitor, from the extracellular fluids (see Figure~\ref{fig:HH}) \cite{Hodgkin1952}.

\begin{figure} [H]
	\begin{center}
	\includegraphics{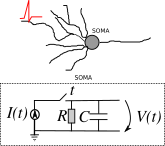}
	\caption{Schematic diagram of the leaky-integrate-and-fire neuron model. The base circuit is the module inside the grey circle on the right-hand side. A current I(t) charges the \protect\gls{rc} circuit. If the voltage V(t) across the capacitance reaches the threshold $V_t$ then is a spike generated and V(t) is set to the reset voltage during a refractory period. Adapted from \protect\citep{Gerstner2002}. \label{fig:LIF}}
	\end{center}
\end{figure}

Currents are injected through the three different types of channels (sodium, potassium, and leakage), each with a distinct conductivity \cite{Hodgkin1952}. More specialised neuron models inspired on the \gls{hh} and also based on differential equations were proposed such as the FitzHugh-Nagumo \cite{nagumo1962}, Morris-Lecar \cite{morris1981}, Hindmarsh-Rose \cite{rose1989},Komendantov-Kononenko \cite{komendantov1996} and Wilson \cite{wilson1999}.\\
A more simplistic model was proposed by Gerstner and Kistler \cite{Gerstner2002} proposed the \gls{iaf} and some variations, including the \acrfull{lif}. The main difference between the \gls{lif} and \gls{iaf} is that the \gls{lif} is more realistic than the \gls{iaf} because it includes a leakage resistor to lower the membrane potential voltage when no pre-spikes are received. \gls{lif} neurons can be represented using an electronic \gls{rc} circuit, as shown in Figure~\ref{fig:LIF}.

The action potential is governed by Equation \ref{eq:2.1}.
\begin{eqnarray}
\label{eq:2.1}
\dfrac{\tau \delta V \left(t\right)}{\delta t} =- V \left( t \right) + RI \left( t \right)\\
\nonumber
\end{eqnarray}
where $\tau = RC$ is the time constant, R the membrane resistance, C the membrane capacitance, V(t) the membrane voltage at a given time t and I(t) is the current at time t.

\begin{figure} [h]
	\begin{center}
	\includegraphics[width=0.8\textwidth]{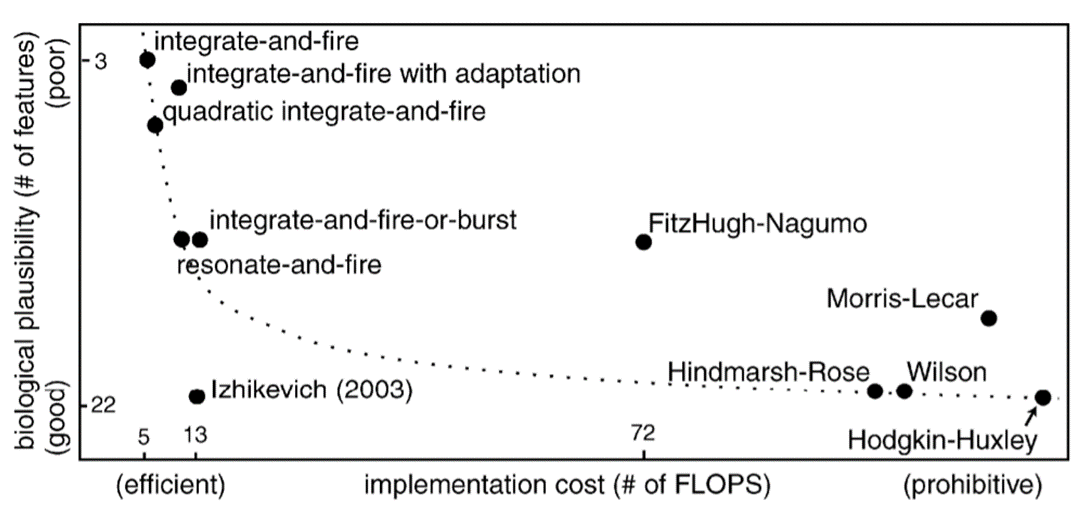}
	\caption{Spiking neural models and its performance. Adopted from \protect\cite{Izhikevich2004}.} \label{fig:neurons}
	\end{center}
\end{figure}

Izhikevich \cite{Izhikevich2004} proposed the \gls{izk} model that is as biologically plausible as the \gls{hh} model, but with a computational efficiency comparable to the \gls{iaf}/\gls{lif} models. In his work, Izhikevich \cite{Izhikevich2004} presented a comparative study between some of the most relevant spiking and bursting neural models, discussing the use of each one in large-scale simulations. Figure~\ref{fig:neurons} depicts the implementation of the computational cost versus biological plausibility of the spiking and burst neural models.

Figure~\ref{fig:neurons} shows that the simplest model, in terms of implementation cost, is the \gls{iaf}. The \gls{iaf} is also the least realistic neural model; however, some variations of \gls{iaf} (\gls{iaf} with adaptation, quadratic \gls{iaf} resonate-and-fire and the \gls{iaf}-or-burst models) have a higher level of biological realism \cite{Gerstner2002,Izhikevich2004}. The Izhikevich model proves to have a good balance between biological plausibility and implementation cost \cite{Gerstner2002}. Finally, the \gls{hh}, FitzHugh-Nagumo, Morris-Lecar, Hinmarsh-Rose and Wilson which are the most realistic models, also require more floating-point operations per second (FLOPS) \cite{Izhikevich2004}. The \gls{lif} neuron model has a good balance between processing cost and biological compatibility dynamics (see Chapter~\ref{Ch3:mhsnn} and \ref{Ch4:object_motion_detection_cells}) and suitable for implementation in large-scale on  (see Chapter~\ref{Ch5:neuromorphic_object_motion_detector} for further details). In this PhD research programme, the \gls{lif} model was selected over other Spiking Neuron models; because it requires less computational resources, which is desirable for implementing on dedicated hardware (see Chapter~\ref{Ch5:neuromorphic_object_motion_detector} for more details).

\section{Spiking Neural Networks simulators} \label{Ch2.3:snn_simulators}
\glspl{snn} with different spiking neuron models and synaptic models, both parameterised with custom parameters and freely interconnected. The Retiner \cite{Pelayo2003,Pelayo2004} is a framework designed for testing the retina model, designed and implemented by the Cortivis consortium \cite{CORDIS2021}. The Retiner simulates nine types of cells and includes edge detection, motion detection, and colour discrimination in real-time. The primary computational colours, \gls{rgb}, and light intensity are filtered by Spatio-temporal retina-like filters \gls{dog} and the laplacian-of-gaussian operators. The output of these filters is weighted and combined to produce a unique intensity matrix. A leaky integrator is used to convert the intensity matrix into a voltage activity matrix. The activity matrix is then processed by \gls{lif} neurons, and the result is presented in the form of spike events.

\begin{figure} [htb!]
	\begin{center}
	\includegraphics[width=0.8\textwidth]{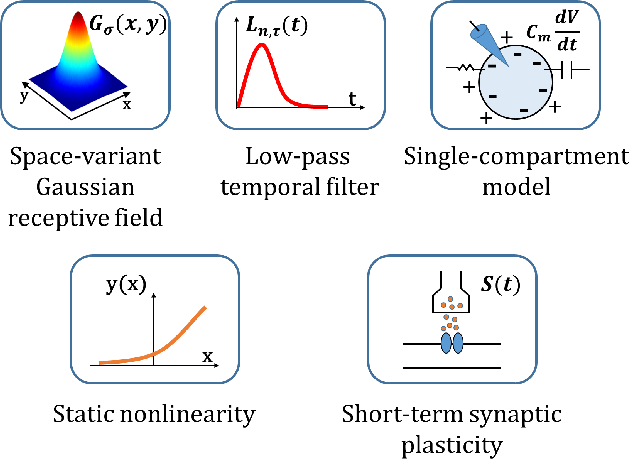}
	\caption{Computational retinal microcircuits that are used as basic building blocks within COREM. Adopted from \protect\cite{Martinez-Canada2016}} \label{fig:retinal_micro_circuits}
	\end{center}
\end{figure}

Martinez-Canada \textit{et al.} proposed the COREM computational framework for realistic retina modelling \cite{Martinez-Canada2016}. In this work, five computational retinal microcircuits were used as building blocks to model different retina mechanisms. The five computational microcircuits described in \cite{Martinez-Canada2016} are the space-variant Gaussian receptive field, low-pass temporal filter, single-compartment model, static nonlinearity, and short-term synaptic plasticity (see~\ref{fig:retinal_micro_circuits}).

\begin{figure} [htb!]
	\begin{center}
	\includegraphics[width=1.0\textwidth]{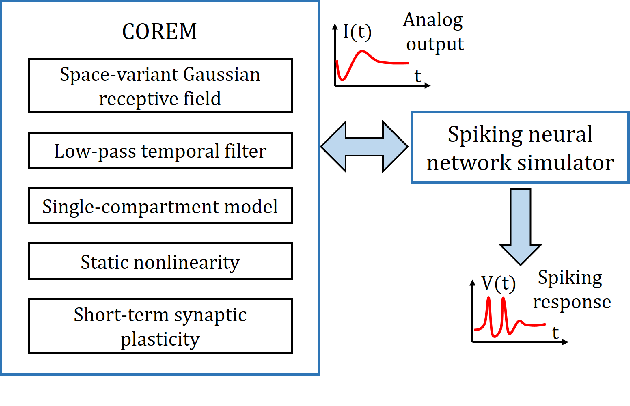}
	\caption{Interface of COREM with NEST. Adopted from \protect\cite{Martinez-Canada2016}} \label{fig:COREM}
	\end{center}
\end{figure}

The accuracy of the computational models was validated by fitting published electrophysiological recordings. Martinez-Canada \textit{et al.} modelled the adaptation to the mean light intensity, fast and slow temporal contrast adaptation, and the object motion-sensitive cells \cite{Martinez-Canada2016}. The five computational microcircuits, proposed by Martinez-Canada \textit{et al.} \cite{Martinez-Canada2016}, can be combined to reproduce single-cell and large-scale retina models at different abstraction levels; and can be classified as block-structured, block-compartment, or single-compartment models. The retinal model was then connected with NEST \cite{NEST2019} to simulate ganglion cells using \gls{lif} neurons. Although Martinez-Canada \textit{et al.} reported that it was possible to replicate retinal cell functionalities using COREM, neither accuracy nor speed details were provided \cite{Martinez-Canada2016}. The COREM framework functionality is represented in Figure~\ref{fig:COREM}.

Although the COREM and Retiner enable researchers to design Neural Networks, the three primary and widely used \gls{snn} simulators are Brian, NEURON and NEST \cite{Tikidji-Hamburyan2017}. Brian \cite{Goodman2009} is one of the most popular \gls{snn} simulators written in Python that delivers a user-friendly environment for users to write code quickly using vector-based computation to deliver efficient simulations. Furthermore, Brian provides flexibility to design custom \gls{snn} with spiking neurons freely interconnected between them. Brian simulator was replaced by the new Brian2 that delivers a more flexible and simplified environment for users to write new neuron models and more complex \glspl{snn} \cite{Stimberg2019}. Brian 2 simulator which has been used in this research programme is described in Chapter~\ref{Ch3:mhsnn}. The neuron models are written as differential equations in standard mathematical notation \cite{Goodman2009}. NEURON is a \gls{snn} simulation toolbox that enables users to design, implement, and run efficient discrete-event \gls{snn} simulations, with easy to integrate hybrid simulations composed of spiking neuron models and cells with voltage gate conductance. The \gls{snn} simulator NEST \cite{NEST2019} is also a simulation environment written in C++ with Python bindings, specially designed for models that focus on the dynamics, size, and structure of \glspl{snn} and offers a wide range of built-in tools to monitor internal variable states and spike events. Many other packages and adaptors for Brian 2, NEURON and NEST, are available (such as PyNN, Cypress, NengoDL, etc.) but will not be discussed in this thesis; the reader can find more details in \cite{Tikidji-Hamburyan2017,Lobo2019,Rasmussen2018}.

\section{Spiking Neural Networks architectures suitable Computer Vision Processing} \label{Ch2.4:snn-arcitectures}
\glspl{snn} have been used for performing different tasks in Computer Vision. Long et al. \cite{Long2008} proposed the JSpike framework that delivers a catalogue of algorithms suitable for designing large-scale \glspl{snn} designed to require minimal memory and processing per synapse and therefore suitable for computer vision processing. The JSpike was one of the first \gls{snn} simulators optimised for computer vision processing. Nevertheless, there are no recent publications nor an active JSpike project, and therefore, it could be concluded that the JSpike project is inactive.

Wu et al. proposed bio-inspired \gls{snn} for segmenting objects and bind their pixels to construct object forms using excitatory lateral connections \cite{Wu2008}. In a follow-up work, Wu et al. \cite{Wu2008b} proposed another \gls{snn} for detecting moving objects in a visual image sequence. The \gls{snn} were trained for extracting the boundaries of moving objects from grey images \cite{Wu2008b}. Cai \textit{et al.} \cite{Cai2012} expanded the work in \cite{Wu2008} and mimicked the basic functionality of motion detection with axonal delays. Despite the two \gls{snn} architectures \cite{Wu2008,Cai2012} being able to detect moving objects, neither is able to process moving objects in real-time.

Zylberberq et al. \citep{Zylberberg2011} used a biologically inspired sparse coding model using \glspl{snn} for emulating the types of responses that are found in cortical neurons in the primate visual cortex (also known as Gabor functions). Zylberberg et al. used a population of inhibitory neurons and a second population of excitatory neurons. The inhibitory neurons are used for producing lateral inhibition, which is required to generate the same patterns generated by Gabor functions. They also introduce a new unsupervised learning technique, an adaptation of Oja's learning rule, \citep{Oja1982} for training the weights of both populations of neurons. Nevertheless, Gabor filters are especially useful in classification and recognition applications, which is not the main focus of this PhD research program.

Kerr et al. \cite{Kerr2013} proposed a custom \gls{snn} to model bio-inspired photoreceptors receptive fields and integrator neurons for extracting important features from intensity and range images, to perform edge-detection. Although the bio-inspired edge-detection proposed by Kerr et al. perform this visual task, the authors did not perform an exhaustive analysis to assess the quality of edge detection. Furthermore, in 2015, Kerr et al. \cite{Kerr2015} proposed a four-layer hierarchical neural network for the extraction of complex features from natural images. The input image is convolved with \gls{dog} filters, and the result is converted by ganglion cells into spike events. The processing stages include edge detection, orientation detection, end-stopped detection and interest point detection. The proposed \gls{snn} was implemented in the Brian simulator \cite{Goodman2009} and is capable of processing $800\times600$ pixels images in 3.64s. Although 3.64s is a good processing speed for a \gls{snn}, this speed is far from ideal because the typical commercial camera frame rate is around 30 frames per second (33 ms).

Tavnaei et al. \citep{Tavanaei2016} who proposed a biological-inspired \gls{csnn} composed of a convolution layer, followed by a pooling layer and a fully connected layer (feature discovery). On classification tasks on the MNIST digit dataset\footnote{The MNIST database of handwritten digits is widely used to measure the accuracy of machine learning and artificial intelligence algorithms. \protect\url{http://yann.lecun.com/exdb/mnist/}, last accessed: 27/02/2018.}, the proposed architecture showed an accuracy of 98\% for clean (without any additive noise) images. Rueckert et al. \citep{Rueckert2016} proposed a \gls{rsnn} for planning tasks that detect movements. The proposed architecture is composed of 2 layers of \gls{lif} neurons, one for saving the current state and another for keeping the context. The \gls{rsnn} is inspired by hippocampal neurons found in rats; however, such recurrence is not found in vertebrate retinas. Therefore, \gls{csnn} and \gls{rsnn} are optimised to perform classification and not object motion detection.

Sun et al. \cite{Sun2017} proposed a \gls{snn} that combines \gls{lif} neurons in hierarchical layers with excitatory/inhibitory pathways to describe receptive fields used to extract colour features for object recognition. Sun et al. made use of the unsupervised learning \gls{stdp} to train the excitatory synapses and therefore improve the classification accuracy on the 5 public datasets \cite{Sun2017}. The authors claim an accuracy of about 90\% in four of the 5 datasets used. Machado et al. \cite{Machado_2019} proposed the NatCSNN, a 3-layer convolutional spiking neural network for the classification of objects extracted from natural images. The authors \cite{Machado_2019} suggested the use of \gls{stdp} unsupervised learning for training the middle layers and the \gls{resume} supervised learning algorithm for teaching the output layer neurons using teacher signals. Although the NatCSNN scored an accuracy of 84.7\%, the complex structure (use of a variation of the \gls{lif} with adaptive threshold neuron model for preventing over-spiking, \gls{stdp} and \gls{resume} synapse models to interconnect neurons via many-to-many connectivities) of the NatCSNN introduces severe limitations in speed performance. Therefore, the NatCSNN is not suitable nor scalable for applications targetting real-time or near real-time processing. Although both works \cite{Sun2017} \cite{Machado_2019} are based on hierarchical \glspl{snn} that use a combination of excitatory/inhibitory pathways to describe receptive fields, none of these architectures is easily scalable nor suitable for real-time applications.

The works proposed and covered in this section share the following similarities: (i) use of \gls{lif} neurons, organised in a hierarchical network; (ii) neural networks are composed of 3 or more layers which are interconnected via excitatory and/or inhibitory synapses, forming receptive fields; (iii) inclusion of final layer trained using \gls{stdp} and/or \gls{resume}. Neither of the studies addressed emulation of \glspl{oms-gc} and the reviewed \gls{snn} architectures are not easily scalable, nor are suitable for processing images in real-time. Although only the \gls{snn} architecture that was proposed by Wu et al. \cite{Wu2008b} was capable to perform \gls{omd}, the proposed \gls{snn} was not thoroughly tested against any public \gls{omd} datasets (e.g. \gls{cdnet2012} \cite{Goyette2012} or \gls{cdnet2014} \cite{Wang2014}). Therefore, this PhD research programme explored the use of \gls{snn} for performing \gls{omd} in real-time applications.

\section{Object Motion Detection} \label{Ch2.5:object_motion_detection}
\begin{figure} [h]
	\begin{center}
	\includegraphics[width=1.0\textwidth]{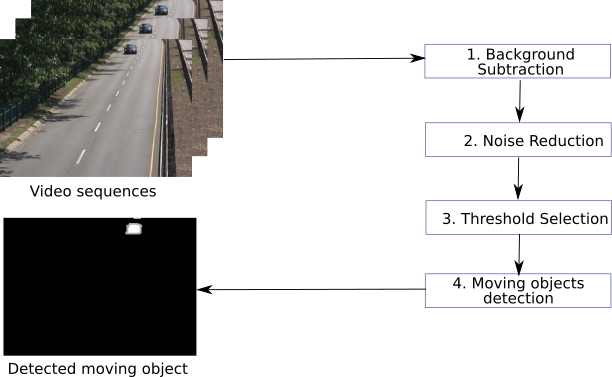}
	\caption{\acrlong{omd} steps.} \label{fig:omd-steps}
	\end{center}
\end{figure}

The detection of moving objects from video frame sequences is a trivial visual task performed by vertebrate retinal \gls{gc} \cite{Gollisch2010,Kolb2003} and yet a challenge in the \gls{cv} research field. \gls{omd} is one of the most researched fields in computer vision and has been studied for more than 30 years \cite{Chapel2020}. \gls{omd} in videos captured from static and/or moving cameras is essential for a wide range of computer vision applications such as video surveillance, object collision avoidance, \gls{adas}, etc \cite{KumarBheemavarapu2021,Garcia-Garcia2020,Chapel2020}. Although the initial \gls{omd} models were designed for static cameras, the advances in sensor technology and the accessibility to portable devices fitted with cameras is triggering more challenging scenes where both cameras and objects can move at the same time \cite{Chapel2020}. \gls{omd}  includes the following tasks: 1) \acrlong{bs}, 2) noise reduction, 3) threshold selection and 4) moving objects detection (see Figure~\ref{fig:omd-steps}).

Several challenges have been identified in various works \cite{bouwmans2010,bouwmans2014,bouwmans2014book,Bouwmans2019,Garcia-Garcia2020,Chapel2020} and can be summarised as follows :
\begin{itemize}
 \item \textbf{Bootstrapping}: the sequence of images includes objects in both the background and foreground. 
 \item \textbf{Camouflage}: the objects in the foreground are either obstructed by background objects or are composed of similar colours.
 \item \textbf{Dynamic background}: the objects in the background include parasitic movements such as surface water movement, branches and leafs shaking in trees, flags on windy days, etc.
 \item \textbf{Camera aperture}: blurred background and foreground as a consequence of the incorrect opening in a lens through which light passes to enter the camera.
 \item \textbf{Variation of illumination}: instant variations of illumination will increase the number of false-positive detections (i.e. pixels that should belong to the background are classified as foreground).
 \item \textbf{Low frame rate}: the temporal distance between image frames prevents instant updates of the background and illumination changes, which reduces the accuracy and increases the number of false positives.
 \item \textbf{Motion blur}: caused by rapid camera movements or jittering, which blurs the image.
 \item \textbf{Parallax}: the apparent displacement of an object as a consequence of the camera movement. The parallax will have implications on the background modelling and its compensation.
 \item \textbf{Moving camera}: moving cameras introduce complexity because the static objects seem to be moving, and objects moving at a similar speed in the same direction of the camera will seem to be static.
 \item \textbf{Background objects movement}: although static objects can be added to and removed from the background, such objects should still be considered static.
 \item \textbf{Night videos}: night videos have dim light, lower contrast and reduced colour information.
 \item \textbf{Noisy images}: low-quality sensors, dust exposure, dirty lens, bright lights and low resolution are examples of factors that cause noisy images.
 \item \textbf{Shadows}: shadows created by objects when exposed to light sources (e.g. sun rays and artificial illumination) should not be part of the foreground models.
 \item \textbf{Stationary foreground objects}: a foreground that has stopped moving for a short period should not become part of the background model; 
 \item \textbf{Challenging weather}: weather conditions (such as fog, rainstorms, strong winds, intense sun rays) have a major impact on the image quality and reduce the quality of the image drastically.
\end{itemize}

Furthermore, Garcia et al. \cite{Garcia-Garcia2020} and Chapel et al. \cite{Chapel2020} identified \gls{omd} use-cases that can address many challenges discussed above and can be summarised as follows:
\begin{itemize}
 \item \textbf{Visual analysis of human activities} fixed, or movable cameras used for monitoring human activities. Human activities can include highway maintenance (e.g. traffic density estimation, vehicle tracking, detection of dangerous manoeuvres)\cite{monteiro2008a,monteiro2008b,monteiro2008c,monteiro2008d,batista2006,hadi2017,virginas2014,lin2009}; tracking people in public places (e.g. airports, train stations, seaports) \cite{masoud2001,yi2020,qureshi2009,wang2008}; monitoring specific people/objects in mass events (e.g. sports games, music concerts, manifestations, and gatherings) \cite{chen2014,leong2014,guler2006}; and indoor/outdoor behaviour analysis (e.g. track people in closed public spaces, body-cameras installed police forces) \cite{li2014,cermeno2018,dotter2021}.
 \item \textbf{Visual observation of animals behaviours} the observation of animals enables one to better understanding the health status of individual and/or colonies of animals. Animals behaviours may include monitoring of livestock (monitor cows, pigs, and diseases detection through the analysis of atypical movements) \cite{ren2021,pasupa2015,Guo2014,okura2019};
 gathering understanding about complex colonies of insects (e.g. bees and ants have communication mechanisms that enable them to work together to solve complex problems) \cite{fasciano2014,kimura2014,yang2018,wu2020}; and monitoring wildlife (e.g. track movements of shoals of fish or pods of whales) \cite{zhao2016,de2018,srividya2014,qin2014,hossain2016,seese2016}.
 \item \textbf{Visual observation of natural environments} the detection of foreign objects in natural environments is crucial for such environments. Some examples of natural environments are forests, lakes, rivers, oceans, and glaciers that require active human intervention to protect biodiversity in terms of fauna and flora \cite{salman2019,wawrzyniak2019,fattahzadeh2020,ghaffarian2021,cao2021}. 
 \item \textbf{Visual hull computation} object motion detection is currently being used in many sports (such as Football, Tennis, and Athletics) to perform athletes movements analysis \cite{reno2015,liu2015,barhoumi2015,kim2018,meghana2019,kim2019}. 
 \item \textbf{\gls{hmi}} gesture recognition enables users to interact with machines. \gls{hmi} is currently being used in many fields such as in games (gestures are mapped into game instructions); health (capture of facial motions to enable patients with severe muscular degenerative diseases to communicate using computers); and augmented/virtual reality (enable interaction between users and mixture of real and virtual objects for completing specific tasks) \cite{chakraborty2018}.
 \item \textbf{Content-based video coding/decoding} the foreground can be extracted from the background and encoded and streamed to the destination. At the destination, the foreground can be decoded and added to the pre-existing background model \cite{podder2015,zhao2017}.
 \item \textbf{Background substitution} state-of-the-art conference platforms enable users to blur or substitute the background in conference calls. The background substitution is being widely used to protect users' privacy \cite{huang2017,grega2017}.
 \item \textbf{\gls{adas} and \gls{ns}} require constant updates of the foreground models to ensure that autonomous/semi-autonomous systems can safely navigate without colliding with a multitude of objects and living beings moving at random speeds and variable trajectories \cite{del2014,dixit2015,vasuki2016}.
\end{itemize}

\subsection{Background Subtraction}

Several surveys about \gls{bs} have been published in the literature focused on static or semi-static (i.e. cameras fixed in a given position exhibiting pan-tilt-zoom movements) scenes. McIvor \cite{mcivor2000} published, in 2000, one of the first \gls{omd} surveys where nine \gls{bs} methods which were only described in detail but not compared. Piccardi \cite{piccardi2004} presented, in 2004, a comparative study between seven algorithms cording to speed, memory resources utilisation and accuracy (see Table~\ref{tab:bs_analysis}). Piccardi's study \cite{piccardi2004} aims to facilitate the \gls{bs} selection based on speed, memory requirements and accuracy requirements. 

\begin{table}[h] \caption{\gls{bs} methods and their performance analysis} \label{tab:bs_analysis}
\resizebox{14.5cm}{!}{\begin{tabular}{|l|c|c|c|}
\hline\hline
Method & Speed & Memory & Accuracy \\ \hline
Running Gaussian average \cite{wren1997,koller1994}& high & low & acceptable \\ \hline
Temporal median filter \cite{lo2001,cucchiara2003} & high & low & acceptable \\ \hline
Mixture of Gaussians \cite{Stauffer1999} & low & high & very good \\ \hline
Kernel density estimation \cite{han2004}  & low & high & very good \\ \hline
Sequential kernel density approximation \cite{elgammal2000} & low & acceptable & good \\ \hline
Cooccurence of image variations \cite{seki2003} & acceptable & acceptable & good \\ \hline
Eigenbackgrounds \cite{oliver2000} & acceptable & acceptable & good \\ \hline
\end{tabular}}
\end{table}
Cheung et al. \cite{cheung2005} proposed a method for validating foreground regions (blobs) using a slow-adapting Kalman filter and compared the proposed method against six other methods using the recall and precision metrics. Elhabian et al. \cite{elhabian2008} covered several background removal algorithms and identified that all the \gls{bs} algorithms follow four significant steps, namely, pre-processing, background modelling, foreground extraction, and validation. Although the review was very comprehensive, the focus was on recursive and non-recursive approaches, which are suitable for background maintenance but less suitable for background modelling. Cristiani et al. \cite{cristani2010} reviewed \gls{bs} methods that can be applied to data captured from different sensor channels (including audio). Elgammal \cite{elgammal2014} reviewed more than 100 papers about object motion detection for static and moving cameras, highlighting the challenges and suggesting which method to use in each case. Bouwmans et al., Garcia et al. and Chapel et al. published comprehensive surveys \cite{bouwmans2010,bouwmans2014,bouwmans2014book,Bouwmans2019,Garcia-Garcia2020,Chapel2020} focusing on traditional, recent, and prospective object motion detection methods.

Consecutive frame difference, background modelling and optical flow are the main categories for \gls{bs}. Consecutive frame difference methods are the simplest to implement and require less computational resources, but are also the most sensitive to the challenges listed above \cite{Chapel2020,Garcia-Garcia2020}. In contrast, optical flow methods are the most robust but require more computational resources and, consequently, are not suitable for real-time applications \cite{Chapel2020,Garcia-Garcia2020}. Therefore, background modelling methods are commonly used methods for extracting the foreground from the background in real-time applications \cite{Chapel2020,Garcia-Garcia2020}.

\begin{figure} [h]
	\begin{center}
	\includegraphics[width=1.0\textwidth]{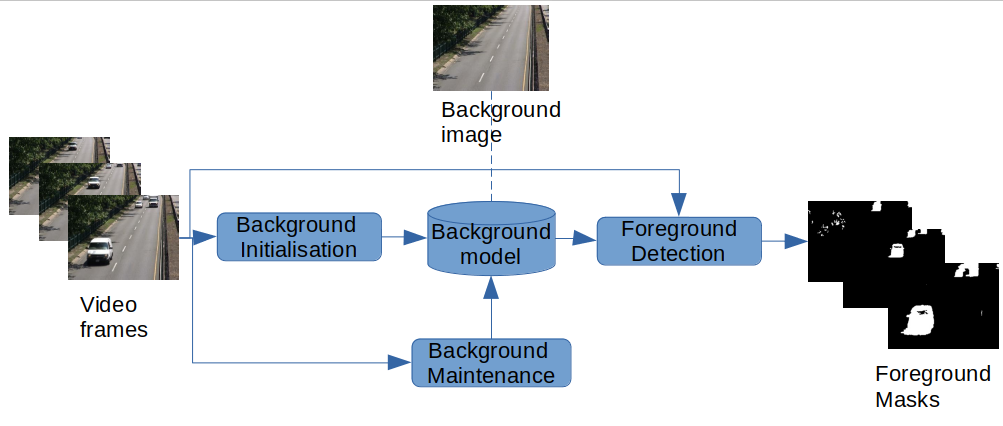}
	\caption{\acrlong{bs} steps.} \label{fig:bs-steps}
	\end{center}
\end{figure}

\gls{bs} generic steps are detailed in Figure~\ref{fig:bs-steps}.

Stauffer \& Grimson \cite{Stauffer1999}, and KaewTraKulPong \& Bowden \cite{KaewTraKulPong2002} suggested modelling each pixel as a \gls{mog} where the Gaussian distributions of the adaptive mixture model are analysed for determining which ones are likely to belong to the background process. All the pixel values that do not fit in the background distributions are considered foreground \cite{Stauffer1999}. Zivkovic \cite{Zivkovic2004} proposes an efficient adaptive algorithm using the \gls{mog2} for enhancing the \gls{mog} algorithm. \gls{mog2} selects automatically the number of components per pixel, which results in complete adaptation to the observed scene. Zivkovic \& Heijden \cite{Zivkovic2006} identified recursive equations for updating the parameters of the \gls{mog} and proposed \gls{knn} for the automatic selection of the pixel components. The Gaussian mixture based algorithms (\gls{mog}, \gls{mog2} and \gls{knn}) show good robustness when exposed to noise and losses due to image compression but lack sensitivity to intermittent object motion, moving background objects and abrupt illumination changes.

In 2016, Sagi Zeevi \cite{CNT-2016} proposed the CNT algorithm, which performed better on the \gls{cdnet2014} dataset \cite{Wang2014} and targets embedded platforms (e.g. Raspberry PI\footnote{Available online, \protect\url{https://www.raspberrypi.com/products/raspberry-pi-4-model-b/}, last accessed 28/03/2022}). The CNT uses minimum pixel stability for a specified period for modelling the background; this can vary from 170 ms, default for swift movements, up to hundreds of seconds, where 60s is the default value \cite{Zeevi2016}. Godbehere \textit{et al.} \cite{Godbehere2014} suggested a single-camera statistical segmentation and tracking algorithm named the GMG by combining per-pixel Bayesian segmentation, a bank of Kalman filters, and Gale-Shapley matching for the approximation of the solution to the multi-target problem. The proposed GMG algorithm is limited when processing video streams susceptible to camouflage, losses due to image compression, and noise.

Guo \textit{et al.} \cite{Guo2016} reported an adaptive \acrlong{bs} model enhanced by a local \gls{lsbp} for addressing illumination changes. The proposed \gls{lsbp} algorithm enhances the robustness of the motion detection to illumination changes, shadows, and noise. However, the \gls{lsbp} is less effective when processing video streams susceptible to camouflage, losses due to image compression, or external noise. More recently, in 2017, \gls{opencv} released an improved version of the \gls{lsbp} algorithm, also known as \gls{gsoc} \cite{GSOC2020,samsonovcode2017}, developed during the Google Summer of Code event \cite{Google2017}, which enhances the \gls{lsbp} algorithm by using colour descriptors and stabilisation heuristics for motion compensation \cite{Samsonov2017,samsonovcode2017}. The \gls{gsoc} algorithm demonstrates better performance on the \gls{cdnet2012} \cite{Goyette2012}, and \gls{cdnet2014} \cite{Wang2014} datasets \cite{Samsonov2017,OpenCV2021} when compared to other algorithms available on the \gls{opencv} library.\\

More recently, Braham et al. \cite{braham2017} proposed a \gls{sbs}that uses object-level semantics to meet a range of problematic background subtraction conditions. The proposed \gls{sbs} reduces false positive detections by integrating the output information of a semantic segmentation method, expressed as a probability for each pixel, with the output of existing \gls{bs} methods. Inspired by Braham's work \cite{braham2017}, Zeng et al. \cite{zeng2019} proposed a \gls{rtss} for performing \gls{bs}. The \gls{rtss} consists of two components: a \gls{bs} segmenter B and a semantic segmenter S that work in parallel for foreground segmentation. The \gls{rtss} achieves state-of-the-art performance among most unsupervised background subtraction methods while functioning in real-time as compared to other BS methods \cite{zeng2019}. Liang et al. \cite{liang2018} proposed a deep background subtraction method using a directed learning strategy that learns a specific \gls{cnn} model for each video without manually labelling. Zeng et al. \cite{zeng2018} proposed a \gls{mfcn} architecture for background subtraction that takes advantage of diverse layer features. The deep features learned from \gls{mfcn} improves foreground detection and that the complexity of the background subtraction process can be easily handled during the subtraction operation itself.

\gls{bs} is the first step and the most important step of the \gls{omd}. \gls{bs} algorithms are responsible for extracting the foreground from the background using the spatio-temporal information of the light variation between sequential video frames. Furthermore, \gls{bs} can be done using methods based on signal processing, \gls{ml}, \gls{dnn} or mathematical models. Although signal processing, \gls{ml} and \gls{dnn} tend to exhibit better accuracy than mathematical models, these algorithm types are also computationally intensive, introducing undesirable latencies. Nevertheless, mathematical models exhibit lower accuracy but require fewer computational resources and, therefore, are suitable for real-time applications. Moreover, the methods proposed by Braham et al. \cite{braham2017} and Zeng et al. \cite{zeng2019b} demonstrated that existing \gls{bs} algorithms can be improved when combined with semantic segmentation models. Therefore, this PhD research programme combined a customised \gls{snn} for improving existing \gls{gsoc} \gls{bs} algorithm for mimicking simple biological functionalities observed in \gls{oms-gc} of vertebrate retinas.\\

\subsection {Noise reduction}
Parks \& Fels \cite{parks2008} evaluated several noise reduction post-processing strategies and concluded that morphological operators outperform other techniques such as the median filter. This technique can repair small gaps in the foreground segmentation and erase small clusters of pixels that were incorrectly labelled as foreground. Shadows and reflections introduce severe challenges, since most moving object recognition techniques incorrectly classify their region pixels as moving objects. Although it is preferable to deal with these issues by developing an effective algorithm at the time of detection, post-processing methods based on threshold selection can be useful in removing erroneous information from the output data. Solehah et al. \cite{yaakob2012} proposed comparing the current image's histogram with the one of the warped background and thresholding it to re-classify the pixels to remove noisy pixels associated with the foreground. This PhD research programme used median filters for filtering salt-and-pepper noise \cite{chan2005} as consequence of random pixel illumination variations.

\subsection {Threshold selection}
The threshold can be set to the same value for all pixels and the series. This plan is straightforward, but it is not ideal because pixels depict various actions, necessitating the use of an adaptable threshold. This can be accomplished by determining the threshold using the local temporal standard deviation of intensity between the background and current images, and then updating it with an \gls{iir} filter as suggested by Collins et al. \cite{collins2000}. According to Wren et al. \cite{wren1995}, an adaptive threshold can also be calculated statistically from the pixel variance. Chacon-Muguia and Gonzalez-Duarte \cite{chacon2011} proposed the use of fuzzy thresholds for performing adaptive thresholds using a one-to-one \gls{som} architecture to deal with dynamic backgrounds for object detection and shadow removal. This PhD research programme used populations of spiking neurons (see Chapters \ref{Ch3:mhsnn} and \ref{Ch4:object_motion_detection_cells}) and synapses with different propagation delays (see Chapter \ref{Ch3:mhsnn}) to perform adaptive threshold selection.

\subsection{Moving object detection}
Detecting the physical movement of an object in a specific location or region is known as \gls{omd}. Unlike \gls{bs} algorithm, which compares light variation between the current frame and previous frames, the \gls{omd} detects moving objects in the scene. Moreover, the quality of the \gls{bs} (covered in section \ref{Ch2.5:object_motion_detection}) has a direct impact on the quality of the \gls{omd} which is the building-block of other advanced visual tasks (such as classification of moving objects, tracking, interpretation and description of actions, human identification, or fusion of data from multiple cameras) \cite{hu2004}. This PhD research programme was focused on the \gls{omd} and therefore, this section will briefly review some of the most relevant \gls{omd} methods available in the literature.

\subsubsection{Representation learning}
Modelling the background is a complex visual task, especially in scenarios where the camera, objects, or both are moving. Therefore, \gls{ml}/\gls{ai} methods are normally used for modelling the background \cite{Chapel2020}

Oliver et al. \cite{oliver2000} proposed the use of \gls{pca} models to lower the dimensionality and perform unsupervised learning of the illumination variation, which is referred to as subspace learning. Other subspace learning methods based on discriminative \cite{farcas2010, farcas2012} and mixed \cite{marghes2010} models were proven to be more robust for foreground detection. Nevertheless, \gls{pca} based models are very sensitive to noise, outliers and incomplete data.

\gls{rpca} with decomposition into low-rank plus sparse matrices have been frequently employed in the area to address the \gls{pca} based algorithms limitations \cite{candes2011,sobral2015,javed2016,javed2016b}. Although these approaches are tolerant to light variation as well as dynamic backgrounds, batch algorithms are required, which makes these methods unsuitable for real-time applications \cite{Chapel2020}. Dynamic \gls{rpca} and robust subspace tracking have proven to overcome the \glspl{rpca}' limitations and attain real-time performance of \gls{rpca} based algorithms \cite{he2012,guo2014b,rodriguez2016,narayanamurthy2018,vaswani2018,javed2018,prativadibhayankaram2018}.

Tensor \gls{rpca} based methods \cite{javed2015,sobral2015b,lu2019,driggs2019} allow for the consideration of spatial and temporal constraints, making them more noise resistant. Although tensor/dynamic \gls{rpca} based algorithms are more robust than \gls{pca} based algorithms, they are also computationally intensive. Therefore, \gls{pca} was used in the work presented in section~\ref{Ch3:mhsnn} for reducing the dimensionality of the datasets' image frames.

\subsubsection{Neural networks modelling}

Schofield et al. \cite{schofield1996} were the first to suggest a \gls{nn} for background modelling and foreground detection named \gls{ram-nn}. \glspl{ram-nn} are trained with a single pass of background images, which introduces limitations related to images having to accurately reflect the scene's background and no background maintenance stage. Tavakkoli \cite{tavakkoli2005} proposed a \gls{nn} for separating the background into chunks during the training process. \gls{rbf-nn} are trained with background samples corresponding to its associated block. \gls{rbf-nn} performs as a detector rather than a discriminant, generating a near border for the known class. Furthermore, \glspl{rbf-nn} can learn the dynamic background while addressing dynamic object detection as a single class problem, but also requires a large dataset to be able to generalise the background scenario. Maddalena and Petrosino \cite{maddalena2007,maddalena2008,maddalena2008b,maddalena2008c} proposed the \gls{sobs} method, which is based on a 2D self-organizing \gls{nn} designed for preserving pixel spatial relationships. The background is automatically modelled using the network's neuron weights. A neural map with $n\times n$ weight vectors is used to represent each individual pixel. The \gls{hsv} colour space is used to initialise the weight vectors of the neurons with the relevant colour pixel values. For each new image, each individual pixel value is compared to its current model to identify whether the pixel relates to the background or the foreground. Several SOBS-based variations have emerged  including (multivalued SOBS \cite{maddalena2009}, SOBS-CF \cite{maddalena2010}, SC-SOBS \cite{maddalena2012}, 3dSOBS+ \cite{maddalena2014}, Simplified \gls{som} \cite{sergio2009}, Neural-Fuzzy \gls{som} \cite{chacon2013}, and MILSOBS \cite{gemignani2015}), allowing it to remain among the top methods on the \gls{cdnet2012} \cite{Goyette2012}. \gls{sobs} also performs well in the detection of stationary objects \cite{maddalena2009b,maddalena2009c,maddalena2013}. However, one of the major drawbacks of SOBS-based approaches is that at least four parameters must be manually adjusted, and these methods lack biological plausibility. Therefore, the \gls{nn} modelling methods were not used in this PhD research programme.

\subsubsection{Deep Neural Network modelling}

\glspl{dnn} have been used for performing moving object detection for moving cameras due to the availability of large annotated video datasets such as the \gls{cdnet2012} \cite{Goyette2012} and \gls{cdnet2014} \cite{Wang2014}. Babaee et al. \cite{babaee2017} proposed a single \gls{cnn} that learns relevant features from data and predicts an appropriate background model from video. The proposed method can be utilised in applications for the detection of moving objects in a variety of video scenarios. The single \gls{cnn} method proposed by Babaee et al. \cite{babaee2017} was followed by improved versions that address: foreground detection enhancement \cite{zeng2019}, ground-truth generation \cite{wang2017}, and learning of deep spatial features \cite{lee2018b,nguyen2018,shafiee2018}. The methods above-mentioned are very accurate and address different artefacts, but they are not suitable for real-time applications.

Guo and Qi \cite{guo2013} and Xu et al. \cite{xu2015} proposed the use of \glspl{rbm} \l{modelling the background creation to achieve moving object detection. Xu et al.} \cite{xu2014,xu2014b} used deep auto-encoder networks to accomplish the same goal, whereas Qu et al. \cite{qu2016} used a context-encoder for background initialization. Zhang et al. \cite{zhang2015} proposed the \gls{sdae} to learn robust spatial features and density analysis to model the background, whereas Shafiee et al. \cite{shafiee2016} proposed the \gls{nerem} to acquire deep features for improving \gls{mog} model proposed by Stauffer \& Grimson \cite{Stauffer1999}, and KaewTraKulPong \& Bowden \cite{KaewTraKulPong2002}. Although \glspl{dnn} have proven to perform object detection with a good accuracy, these types of \gls{nn} lack from biological plausibility and therefore, such methods were not used in this PhD research programme.

\subsection{Advanced Object Motion Detection applications}
\gls{omd} methods are the building blocks for real-time applications such as trajectory classification and object tracking \cite{Yazdi2018}. Although most \gls{omd} methods work offline by analysing recorded video sequences, the need for real-time moving object detection is increasingly being recognised, particularly in fields such as sociology, criminology, suspect behaviour detection and tracking, traffic accident detection, crowd tracking, and vehicle and robot navigation \cite{Yazdi2018}. In general, the necessity for a real-time system necessitates a very short calculation time as well as minimal hardware and memory requirements \cite{Yazdi2018}. 

\subsubsection{Trajectory classification}
\begin{figure} [h]
	\begin{center}
	\includegraphics[width=1.0\textwidth]{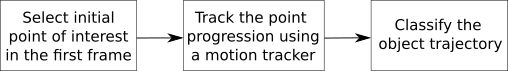}
	\caption{Trajectory classification steps. i) selection of the initial point of interest corresponding to the moving object's centre of mass in the first image frame, ii) tracking the progression of the point of interest and iii) classification of the trajectory described by that point of interest.} \label{fig:obj-trajectory}
	\end{center}
\end{figure}

Several works have used trajectory classification to find moving objects in video sequences acquired by cameras \cite{Yazdi2018,Chapel2020}. The trajectory classification generic steps (see Figure~\ref{fig:obj-trajectory}) include i) selection of the initial point of interest corresponding to the moving object's centre of mass in the first image frame, ii) tracking the progression of the point of interest, and iii) classification of the trajectory described by that point of interest.

Sheikh et al. \cite{sheikh2009} employed a scale space for the background that was formed by the fundamental trajectory bases, and then classified trajectories that were not in this area as belonging to moving objects. Although the proposed method relied just on 2D picture quantities, the homography limitations were successfully extended to 3D scenes to accommodate freely moving cameras. While the algorithm is computationally expensive, the findings revealed that it performs well \cite{sheikh2009}.

Brox \& Malik \cite{brox2010} proposed the use of optic flow for compensating long term motion in a video sequence and extracting the trajectory points. Furthermore, spectral clustering was used to classify background and foreground trajectories. The approach can handle a high number of sequenced frames and partially occluded objects, but it fails to segment the objects densely. Yin et al. \cite{yin2015} used optic flow to compensate for camera parasitic movements, and subsequently \gls{pca} was employed to reduce the number of unusual trajectories. Moreover, the watershed transforms to distinguish foreground and background trajectories. Although label inference is used to handle unlabelled pixels, it fails to create dense segmentation of moving objects \cite{yin2015}. Singh et al. \cite{singh2017} addressed trajectory detection of freely moving cameras in footage acquired while the camera was worn by a human. Instead of employing elaborate models, they use point tracking with optical flow and a bag of words classifier to discern the trajectory of moving objects in their research. The proposed method works well for recognising first-person actions \cite{singh2017}. Zhang et al. \cite{zhang2020} proposed a pre-trained \gls{cnn} to recognise moving object trajectories in an unconstrained video over time by learning adaptive discriminating characteristics from short video clips extracted from the main video. Local tracklets (short trajectories) are extracted from each video clip and linked together across sequential video clips. Even with unrestricted camera movement, the proposed method \cite{zhang2020} works effectively for multi-face tracking. However, depending on the nature of moving objects, the proposed method requires wider training data \cite{zhang2020}.

In this PhD research programme, a custom \gls{snn} architecture is proposed for trajectory classification (see Chapter~\ref{Ch3:mhsnn}). The proposed \gls{mhsnn} architecture combines excitatory and inhibitory synapses with different propagation delays for detecting the object trajectory.

\subsubsection{Object tracking}
\begin{figure} [h]
	\begin{center}
	\includegraphics[width=1.0\textwidth]{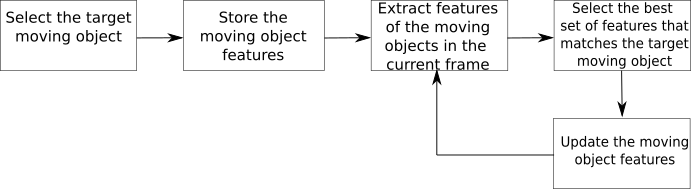}
	\caption{Object tracking steps. i) selection of the target moving object, ii) store the moving object features, iii) extract moving objects features in the current frame, iv) select the best set of features that matches the target moving objects and v) update the moving object features} \label{fig:obj-tracking}
	\end{center}
\end{figure}

Tracking is the process of locating a moving object in sequences of frames \cite{Yazdi2018}. The object tracking generic steps (see Figure~\ref{fig:obj-tracking}) include: i) selection of the target moving object, ii) store the moving object features, iii) extract moving objects features in the current frame, iv) select the best set of features that matches the target moving objects and v) update the moving object features.

The selection of the target moving object can either be done manually by the user or automatically using object detection methods using saliency detection or object segmentation and recognition from a single image \cite{Yazdi2018}. Colours, texture, edges, geometric information, frequency coefficients, simple pixel grey values, or a combination of all of these object properties that comprise a feature space \cite{hare2015, zhang2015b, danelljan2014, du2016, bouwmans2018} can be used to characterise the target moving object. Other features for appearance modelling include the colour histogram \cite{zhao2008} and the \gls{hog} \cite{dalal2005} to use a set of local histograms to characterise a moving object. Furthermore, the occurrence of gradient orientation in a local region of the object is counted in these histograms. Although \gls{hog} provide strong local information about a moving object, it is susceptible to variations in illumination.

Local feature descriptors are crucial in the image registration process and have a direct impact on the accuracy and robustness of image registration \cite{leng2018}. Some of the most common local feature descriptors algorithms utilised for representing moving objects are \gls{gloh} \cite{mikolajczyk2005}, \gls{surf} \cite{bay2006}, \gls{sift} \cite{lowe2004}, \gls{brief} \cite{calonder2010} and \gls{orb} \cite{rublee2011}. \gls{sift} is particularly desirable since it produces local features that are resistant to scale, noise, illumination, and local geometric distortion \cite{ha2011}. Nevertheless, \gls{sift}'s object tracking speed degrades as the number of falsely matched key points increase, which is not desirable when monitoring small objects \cite{ha2011}. \glspl{cnn} can be used for learning adaptive, hierarchical, and distributed features. Such features can be used to update the object's appearance in real time \cite{kim2015} and perform descriptors encoding local spatial-temporal properties \cite{leal2016}.

Statistical models can be used to match the target moving object features to the features extracted from the current frame \cite{Yazdi2018}. Subspace techniques such as \gls{pca} and \gls{ica} can be utilised to decrease the dimension of the feature space. The target moving object's location in the current frame should be determined using a matching strategy. The similarity/correlation function will select the best moving object candidate in the current frame that is the most similar to the target moving object based on appearance features taken from both the candidates and the target \cite{Yazdi2018}. The target moving object features are updated after the target is detected in the current frame, and the process repeats for the following frames \cite{Yazdi2018}. Ning et al. \cite{ning2009} propose using the statistics of the object's visual features (colour and texture) for the matching stage. Leal et al. \cite{leal2016} proposed a template-based method that uses a simple geometric shape, contour, or silhouette of the moving object. Nevertheless, the proposed basic template-base approach fails when the pose of the moving object changes. Pan et al. \cite{pan2008} proposed the \gls{capoa} method that provides a clear distinction between the target and outliers by merging the information provided by spatio-temporal context and motion constraints together.

Comaniciu et al. \cite{comaniciu2003} employed a kernel-based method using mean shift to construct a moving object's histogram based on appearance and similarity measurement using the Bhattacharyya distance to determine the best match for the moving object in the current and future frames. Bebenko et al. \cite{babenko2010} proposed a learning model to identify pixel blocks containing the moving object, and track the moving object's behaviour in subsequent frames. Bagherzadeh \& Yazdi \cite{bagherzadeh2015} proposed the use of saliency maps to extract appearance cues in the frequency domain, and a regularised least squares classifier was used to classify pixels belonging to a moving object.

Another prominent approach for tracking moving objects is tracking-by-detection \cite{breitenstein2009, Milan2015, le2016} , which uses a detection algorithm combined with a matching technique to correlate the discovered items on sequenced images. This technique excels in difficult situations such as uncalibrated moving cameras, changing backgrounds, and, most notably, occlusion. However, matching step or object association is challenging, and the outputs are a distinct set of answers that frequently result in an increase of false positives. Chen et al. \cite{chen2017c} and Sadeghian et al. \cite{sadeghian2017} proposed to use the information from future frames to improve the detection of moving objects in the current frame. Although the currently utilised hand-craft features for moving object detection and tracking offer good results, new trends are toward employing more descriptive features. Nevertheless, the learning process can be used to target particular representations as opposed to a fixed collection of pre-defined traits \cite{kuen2015}. \gls{cnn} have recently been proposed for tracking moving objects, which effectively exploit category specific features for tracking objects even in complex scenarios like moving cameras \cite{ma2015,li2014b,nam2016}. \glspl{cnn} have typically been employed for tracking in one of two ways: as a feature extractor in conjunction with a good classifier \cite{wang2013}, or as a unified deep structure for object tracking \cite{wang2015}. Wang and Ying \cite{wang2013} proposed the use of \gls{dlt} to learn genetic traits from supplemental natural images. Although \gls{dlt} being successful in scenarios with major temporal changes, the proposed method fails to learn these changes efficiently. Wang et al. \cite{wang2015b} used an improved \gls{cnn} architecture to train hierarchical features for model-free object tracking that can manage temporal variations and do online tracking quickly and in follow-up work, Wang et al. \cite{wang2015c}, employed a selection approach including two pre-defined convolutional layers to filter out noisy, irrelevant, or redundant features using \gls{cnn} features retrieved from various layers. Zhai et al. \cite{zhai2018} proposed a \gls{cnn} tracker using a Bayesian classifier as a loss layer and updated the network parameters, which enables the detection of the moving object appearance over the time. Li et al. \cite{li2018} published a comprehensive comparison of \gls{dnn}-based tracking algorithms where it is highlighted that tracking \glspl{cnn} are trained straightforwardly and effectively, resulting in good features for object tracking with the downside of requiring large annotated datasets.

Object tracking methods perform better than \gls{omd} algorithms in complex scenarios where both the cameras and objects moving freely. These approaches, however, substantially rely on the accurate initial selection of the moving object in the sequence's initial frames. Furthermore, tracking multiple objects is a complex problem to solve, especially when both the camera and objects are moving freely. Although tracking is an important task in \gls{omd}, this PhD research programme does not focus on moving object tracking.

\subsubsection{Real-time considerations}

Although most methods for detecting moving objects work offline by analysing recorded video sequences, there is a need for real-time \gls{omd}, particularly in fields such as traffic monitoring, crowd tracking and \gls{adas} \cite{Yazdi2018}. Moreover, real-time applications require low-latency, which is normally preferred by parallelisable hardware such as \glspl{gpu} and \glspl{fpga} and low-spec embedded \glspl{cpu} \cite{Yazdi2018,yu2020}.

There exist a wide variety of objects which exhibit different behaviours, which makes tracking a complex process. Comaniciu et al. \cite{comaniciu2000} proposed a method for real-time tracking of non-rigid objects captured from a moving camera, where the mean shift iterations are used to compute the target position in the current frame. Furthermore, the Bhattacharyya coefficient is used to express the dissimilarity between the target model (its colour distribution) and the target candidate. Comaniciu et al. \cite{comaniciu2000} were able to demonstrate the tracker's ability to manage partial occlusions, considerable clutter, and target scale fluctuations in real time.

The object size impacts on the quality of object's tracking, and it becomes more difficult to track small objects or objects that become progressively smaller when moving away from the centrer of the camera. Ponga \& Bowden \cite{kaewtrakulpong2003} proposed using probabilistic models for tracking small-area targets, which are common in outdoor visual surveillance scenes. The proposed model uses both appearance and motion models in the tasks of classification and tracking objects for which detailed information is not available. Moreover, the proposed method uses motion, shape cues, and colour information to distinguish between the different moving objects in the scene. The results show that the proposed method can track multiple people moving independently and maintain trajectories even when there are occlusions or background clutter.

The complexity of object tracking increases when it is required to perform multiple-object tracking because each object will describe random trajectories and might occlude or be occluded by other objects in the scene. Yang et al. \cite{yang2005} demonstrated a real-time method for tracking several objects in dynamic situations. The proposed method demonstrated that it is capable of coping with long-term and full occlusion without prior knowledge of the object's shape or motion. Extensive testing using video sequences in varied situations of indoor and outdoor environments with long-duration and complete occlusions in changing backgrounds demonstrated that the proposed method achieves good segment and tracking for images of $30\times240$ at 15~20 \gls{fps}.

Selective tracking is also a challenge, especially when the real-time tracking of the target moving object is crucial in sport applications. Grabner et al. \cite{grabner2006} proposed an online AdaBoost feature selection method for employing rapid computable features (such as Haar-like wavelets, orientation histograms, and local binary patterns) in real-time. Heinemann et al. \cite{heinemann2003} extended the work proposed by Grabner et al. \cite{grabner2006} and explored tracking in a controlled environment and proposed a method for accurately recognising and tracking the ball in a RoboCup scenario\footnote{Available online, \protect\url{https://www.robocup.org/}, last accessed: 12/04/2022}. To get a colourless representation of the ball, the authors used Haar features specified by the AdaBoost method \cite{schapire2013} and particle filter for handling the ball tracking. It was demonstrated that, even in a congested environment, the proposed method can follow the ball in real-time (i.e. at 25 \gls{fps}).

Tracking crowd movements is also a complex tracking task, because each pedestrian in the scene can describe a random trajectory and be occluded by other pedestrians. Shah et al. \cite{shah2007} propose an automated surveillance system that is suitable to be used in a wide range of real-world applications, including railway security and law enforcement. The proposed method has been utilised in a number of surveillance-related initiatives that have been supported by governments and private entities. The algorithm is capable of detecting and classifying targets, as well as tracking them across numerous cameras. It also creates a summary in terms of key frames and a textual description of trajectories for final analysis and reaction decision by a monitoring officer.

One of the most complex tracking challenges is associated with tracking unknown objects because the tracking algorithm must distinguish between visual artefacts (such as shadows or ghost objects) and real objects. Bibby et al. \cite{bibby2008} proposed a probabilistic approach for robust real-time visual tracking of previously unknown objects captured by a moving camera. A bag of pixels representation is used to solve the tracking problem, which includes stiff frame registration, segmentation, and online appearance learning. The registration compensates for rigid motion, segmentation models any residual shape distortion, and the online appearance learning refines both object and background appearance models on a continuous basis.

Night objects tracking is also a challenge for real-time object tracking applications because it is difficult to distinguish between the target moving objects and shadows generated by artificial illumination. Huang2008 et al. \cite{huang2008} proposed a real-time object recognition technique for nighttime vision surveillance through contrast analysis. The contrast in local changes over time is employed to detect moving objects in the scene, and the false positives are reduced by combining motion prediction and spatial closest neighbour data association.

Object adaptability is a characteristic of a good real-time tracking algorithm. Li et al. \cite{li2010b} proposed the use of mean shift technique-based solution for global target tracking combined with a background weighted histogram and a colour weighted histogram to represent the model and the candidate. The results show the proposed method may adaptively achieve precise object size with low computational complexity, including camera motion, camera vibration, camera zoom and focus, high-speed moving object tracking, partial occlusions, target scale variations, etc.

Intrinsic characteristics of objects may also have a positive impact on the performance and speed of the algorithm. Danelljan et al. \cite{danelljan2014} proposed a method that performs real-time object tracking using colour features. The findings show that colour features provide greater visual tracking performance and offer a low-dimensional adaptive colour attribute variation for real-time aspects. The authors also demonstrated that the proposed method is comparable to state-of-the-art tracking algorithms, with a surprising speed of up to 100 \gls{fps}.

Navigational applications are amongst the biggest challenges for real-time object tracking algorithms that must be capable of tracking objects of interest in a very short period of time to adjust the trajectory of the moving vehicle. Agarwal et al. \cite{agarwal2017} proposed a real-time multiple object tracking using a \gls{rbcnn} for object detection and a regression network for general object tracking specially design for targetting \gls{adas} applications. Minaeian et al. \cite{minaeian2018} proposed the use of local motions for detecting moving objects and extract tracking keypoints for images captured using \glspl{uav}. Their efforts yielded positive outcomes in real-world applications. 
Overall, it is possible to infer that reliable \gls{omd} algorithms are required in a wide range of applications and their quality will dictate the quality of advanced \gls{omd} applications (such as trajectory classification and object tracking). Although advanced \gls{omd} applications were covered in this section, this PhD research programme does not focus on such applications.

\section{Hardware implementations} \label{Ch2.6:hardware_implementations}

The neuromorphic computing concept was introduced by Carver Mead \cite{Mead1990} in 1990 and was described as the use of \gls{vlsi} equipped with analogue components for emulating biological neural systems. Neuromorphic architectures deliver flexibility for describing highly parallel architectures, require low-power and are typically interconnected to \gls{cpu} via high-speed interfaces (such as \gls{pcie} or intra-chip high-speed bus) \cite{Schuman2017}.\\

Neuromorphic architectures can perform parallel calculations faster, with higher power efficiency and a smaller footprint than the traditional 32/64-bit standard von Neumann architectures \cite{Schuman2017}. Furthermore, the emerging \gls{ai}/\gls{ml} methods require more flexible hardware architectures to accommodate the unique requirements of such methods\cite{Schuman2017}. Emerging neuromorphic architectures deliver higher densities of transistors per unit of space, higher intra-chip communication speed between the customisable fabric, real-time capabilities, and other application-specific processing units (e.g. \glspl{cpu} and \glspl{gpu}) \cite{Intel2021}. Neuromorphic implementations can be split into digital, analogue, and hybrid (both digital and analogue) platforms \cite{Schuman2017}. This PhD research programme aimed to develop a digital neuromorphic solution to accommodate and accelerate the customised \gls{snn} using digital systems \cite{Schuman2017}. \gls{nna} are networks of neuromorphic chips optimised to handle complex neural network workloads that require reconfigurable computational resources and a high degree of parallelism. The comparison between different types of \gls{nna} still remains a challenge, and there is no simple way to compare the performance between \glspl{ann} with \glspl{snn} nor between different types of Neural Networks (i.e. different types of \glspl{ann} or \glspl{snn}). This PhD research programme focused on \glspl{snn}, and therefore, this section only reviews \gls{nna} that were specially designed for accelerating SSNs \cite{Beigne2019}.

\subsection{General propose Neural Network accelerators} \label{Ch2.6.1:gen_purpose_nna}
SpiNNaker \cite{furber2014} is a network composed of custom \glspl{cpu} with an architecture optimised for running customised \glspl{snn} based on \gls{lif} neurons. BrainScaleS \cite{schemmel2010} is another powerful \gls{nna} composed of interconnected wafers composed of high input count analogue neural network cores targeting the rigorous emulation of brain-scale \glspl{nn}. Neurogrid \cite{benjamin2014} is an analogue \gls{nna} for emulating the structure of biological nervous systems in real-time. TrueNorth \cite{merolla2014} is a digital and low-power neuromorphic chip for simulating complex neuronal networks. Loihi \cite{davies2018} neuromorphic chip is a state-of-the-art brought by Intel specially designed for enabling on-chip learning compatible with several learning rules, complex neuron and synapses models targetting complex \glspl{snn}\\

Table~\ref{tab:big_scale} summarises the characteristics (implementation, sampling time, of each one of the \gls{nna}).\\

\begin{table}[h]\caption{Large-scale Neural Networks accelerators characteristics. Adapted from \protect\cite{Beigne2019}}\label{tab:big_scale}
\resizebox{14.5cm}{!}{\begin{tabular}{|l|c|c|c|c|c|}
\hline
Processor & BrainScaleS \cite{schemmel2010} & Neurogrid \cite{benjamin2014} & TrueNorth \cite{merolla2014} & SpiNNaker \cite{furber2014} & Loihi \cite{davies2018} \\ \hline
Implementation type & analogue & analogue & digital & digital & digital \\ \hline
Sampling Type & discrete & continuous & discrete & discrete & discrete \\ \hline
Neuron Update & continuous & continuous & Time MUX & Time MUX & Time MUX \\ \hline
Synapse Resolution & 4b & 13b shared & 1b & Variable & 1 to 64b \\ \hline
Bio-Mimicry & Not Configurable & Not Configurable & Limited to \gls{lif} & Configurable & Configurable \\ \hline
On-Chip Learning & STDP only & No & No & Yes & Yes \\ \hline
Network on Chip & Hierarchica & Tree Multicast & 2D Mesh Unicast & 2D Mesh Multicast & 2D Mesh Unicast \\ \hline
Neurons per Core & 8 $\sim$512 & 65e3 & 256 & $\sim$1e3 & max. 1024 \\ \hline
Synapses per Core & $\sim$130k & 100e6 & 65k x 1b & $\sim$1e6 & 16000 x 64b \\ \hline
Cores per Chip & 352 (wafer scale) & 1 & 4096 & 16 & 128 \\ \hline
Chip Area (mm2) & 50 (single core) & 168 & 430 & 102 & 60 \\ \hline
Technology (nm) & 180 & 180 & 28 & 130 & 14 (FinFET) \\ \hline
Energy/SOP (pJ) & 174 & 941 & 27 & 27e3 & 105.3 \\ \hline
\end{tabular}}
\end{table}

\subsection{Neuromorphic and heterogeneous devices}\label{Ch2.6.2:neuromorphic}

Lichsteiner \textit{et al.} \cite{Lichtsteiner2005} introduced the concept of \gls{aer} silicon retina chip capable of generating events proportional to the log intensity changes. Farian \textit{et al.} \cite{Farian2015} proposed an in-pixel colour processing approach inspired by the retinal colour opponency using the same \gls{aer} concept. Brandli \textit{et al.} \cite{Brandli2014} proposed a new version of the \gls{aer} camera reported by Lichsteiner \cite{Lichtsteiner2005} called the \gls{davis} which exploits the efficiency of the \gls{aer} protocol and introduces a new synchronous global shutter frame concurrently.

Kasabov et al. \cite{Kasabov2013} proposed the deSNN that combines the use of Spike Driven Synaptic Plasticity (an unsupervised learning method for learning spatio-temporal representations) with rank-order learning (supervised learning for building rank-order models). The deSNN \cite{Kasabov2013} was tested on data collected by an \gls{aer} silicon retina chip \cite{Lichtsteiner2005} (which generates events in response to changes in light intensity) for recognising moving objects. Although the deSNN was able to recognise moving objects, the deSNN was designed to work specifically with \gls{aer} cameras. More recently, Jiang \textit{et al.} \cite{Jiang2019} proposed a \gls{snn} based on the Hough Transform to detect a target object with an asynchronous event stream fed by an \gls{aer} cameras. The proposed algorithm \cite{Jiang2019} was able to process up to 40.74 frames per second on an Intel i7-4770 processor, accelerated by an Nvidia Geforce GTX 645. Nevertheless, the authors do not explain if the algorithm would work with regular \gls{cots} cameras.

Oudjail \& Martinet \cite{Oudjail2019} proposes an approach to analyse a moving pattern motion using \gls{snn} from feed captured using \gls{dvs} \gls{aer} camera. A synthetic dataset containing three different moving patterns in four directions was used to train and test the custom \gls{snn} designed by Oudjail \& Martinet \cite{Oudjail2019}. The proposed \gls{snn} \cite{Oudjail2019} is composed of \gls{lif} neurons whose weights were trained using STDP unsupervised learning \cite{Ponulak2010} and the algorithm was implemented in the Brian simulator \cite{Goodman2009}. The authors concluded that the number of output neurons (between 2 and 6 in the output layer) is insufficient to detect the movement's direction accurately. The results of Oudjail \& Martinet \cite{Oudjail2019} work raise the following questions: 1) Was the number of neurons in the input layer sufficient? Because each input neuron received the currents of 150 light intensity values encoded in currents, which would most likely trigger the middle layer neurons to over-spike, forcing the post-synaptic weight to increase until reaching the maximum allowed value; 2) Was the number of neurons on the output layer sufficient? Each output neuron received the synaptic connections from 25 of the previous layer neurons, again the 25 neurons which would most likely trigger the middle layer neurons to over-spike pushing the post-synaptic weight to increase until reaching the maximum allowed value during the training phase; and 3) Supervised learning \gls{resume} (see Chapter~\ref{Ch3:mhsnn} for further details) could have been used to train the output layer neurons because the desired output pattern was well known and could have been used to teach the output layer to recognise the desired patterns.

Jiang et al. \cite{Jiang2019} proposed a \gls{snn} based on the Hough Transform to detect a target captured using a \gls{dvs} \gls{aer} camera. The authors concluded that the speed performance of the \gls{snn} on the \gls{gpu} was better than on the \gls{cpu} \cite{Jiang2019}. Analysis of the results of Jiang's work \cite{Jiang2019} leads to the conclusion that the work is incomplete because 1) little is stated about the dataset; 2) it is not clear how the dataset was preprocessed; 3) no metrics were used to assess the accuracy of the tracking system; and 4) the comparison between the \gls{gpu} and \gls{cpu} speed performance is insufficiently rigorous because a simplified and synchronous update rule was only used on the \gls{gpu}.

Kuriyama et al. \cite{Kuriyama2021} proposed to accelerate the cerebellar scaffold model \cite{DAngelo2013} using heterogeneous computing based on \glspl{gpu}. They simulated synaptic plasticity mechanisms at parallel fibre - Purkinje cell synapses and emulated the gain adaptation of optokinetic response. The results show that the use of \glspl{gpu} enables the processing of 2s of simulation in just 750ms, which is about 100 times faster than previous simulations running on \glspl{cpu}. Although the Kuriyama et al. \cite{Kuriyama2021} model achieves an impressive acceleration of 100 times faster than the \gls{cpu}, the acceleration was obtained using 4 \glspl{gpu} which is not cost-effective (each NVIDIA Tesla V100 \gls{gpu} used in this project cost thousands of dollars). It will not be easily scalable (the majority of modern servers will only have up to 4 \gls{pcie} slots to install \glspl{gpu} and other devices). Moreover, Kuriyama did not include details about the test conditions of the \gls{cpu} implementation nor the \gls{cpu} specifications, which makes it difficult to assess if the comparison between the \gls{gpu} and \gls{cpu} was fair. The emulated circuit was not a retinal like circuit, which is the focus of this research project.

She et al. \cite{She2021} proposed a heterogeneous \gls{snn} (H-SNN) suitable for learning complex spatiotemporal patterns when using unsupervised learning STDP. The authors \cite{She2021} demonstrated analytically that the H-SNN exhibits long and short memory capabilities. It was established, in simulation, that the H-SNN is capable of classifying the object type and motion dynamics. A \gls{gpu} was used to accelerate the \gls{snn} \cite{She2021}, but little details are provided about the level of acceleration achieved. The results reported in \cite{She2021} focus on the accuracy of the H-SNN when compared to other classical CNNs, and it is not clear how accurate the H-SNN is predicting the motion dynamics. Parameshawara et al. \cite{Parameshwara2021} proposed the SpikeMS, a deep \gls{snn} encoder-decoder for motion segmentation of \gls{aer} frames captured using \gls{dvs} \gls{aer} cameras. A novel spatial-temporal loss formulation based on spike counts and classification labels is used on the SpikeMS. The SpikeMS predicts using time windows with a duration of 10 ms. Although the SpikeMS can predict the object type with high accuracy using a 10 ms time window, it is focused on \gls{dvs} \gls{aer} cameras, and it is not clear how it would perform in COTS cameras.

\subsection{FPGA implementations}\label{Ch2.6.3:fpga}
\glspl{fpga} are specialised devices that can be reprogrammable after manufacture and offer high flexibility, high degree of parallelism, high-performance and low-power platforms \cite{chen2017}. \gls{fpga} offers the desirable flexibility for accelerating \glspl{snn} which are characterised for being massively parallel \cite{yang2018a}. \glspl{fpga} are also known for being programmed using low-level \glspl{hdl} such as \gls{vhdl}, and Verilog \cite{Podobas2017}. Nowadays, \gls{fpga} development has been simplified with the introduction of \gls{hls} tools, which allow \gls{fpga} developers to implement their applications using high-level programming languages such as C and \gls{opencl} \cite{Podobas2017}.\\

Mishra et al. \cite{Misra2010} identified in their survey that many of \gls{snn} usually have about $10^4 \sim 10^8$ neurons and $10^{10} \sim 10^{14}$ synapses and that high-performance neural hardware is essential for practical application. Li et al. \cite{Li2010} proposed the implementation of visual cortex neurons on \glspl{fpga}. The implemented visual cortex neurons exhibited the same dynamics as those recorded from real neurons using multi-electrode arrays. Li et al.\cite{Li2012} implemented 256 fully connected neurons, and their performance was assessed by storing four patterns and applying similar patterns containing errors. The implemented system was capable of operating using a 100 MHz clock, which enables the acceleration of the system 40 times above its real-time operation \cite{Li2012}. Cassidy et al. \cite{Cassidy2013} proposed the use of \glspl{fpga} to accommodate spiking neurons and unsupervised STDP\gls{stdp} learning structures. In this work, Cassidy et al. \cite{Cassidy2013} demonstrated that digital neuron abstraction is preferable to more realistic analogue neurons; they also emulated the massive parallelism connectivity and high neuron density as observed in nature; the neuron states were also multiplexed to take advantage of clock frequencies and dense \glspl{sram}.

Moeys et al. \cite{Moeys2016} implemented object motion cells in \glspl{fpga} that takes about 22 clock cycles at 50 MHz to detect motion and also reported that the \gls{fpga} is at least 100 times less than an Intel \gls{nuc} to compute motion. Furthermore, the work in \cite{Moeys2016} shows that the \gls{fpga} implementation has lower latency when compared with the same implementation running on Intel \gls{nuc} at 1.30 GHz and an Intel I7-4770k at 3.50Ghz.\\ Chen et al. \cite{chen2017a} described a \gls{cpg} composed of two reciprocally inhibitory neurons. To reduce the \gls{fpga} resources usages, Chen et al. \cite{chen2017a} has optimised the \gls{cpg} to avoid using multipliers (\glspl{fpga} have a low quantity of multiplier blocks), and the non-linear parts of the Komendantov-Kononenko neuron model \cite{komendantov1996} were removed. Cheung et al. \cite{Cheung2016} proposed the NeuroFlow, a scalable \gls{snn} simulator suitable to be implemented on \gls{fpga} clusters. It was possible to simulate about 600,000 neurons and to get a real-time performance for up to 400,000 neurons simulated using NeuroFlow on 6 \glspl{fpga} \cite{Cheung2016}. Podobas \& Matsuoka \cite{Podobas2017} proposed the use of \gls{opencl}, an \gls{hls} tool, to increase productivity by facilitating the \gls{snn} design (provide a higher level of hardware abstraction) on \glspl{fpga}. Two different neuron models, their axons and synapses, were designed using \gls{opencl} and Podobas \& Matsuoka \cite{Podobas2017} claim a speed performance of up to 2.25 GSpikes/second. Sakellariou et al. \cite{Sakellariou2021} suggested a spiking accelerator based on \glspl{fpga} to enable users to develop \glspl{snn} targetting \gls{ml} applications and promise an acceleration of up to eight hundred times for inference and up to five hundred times for training compared to Software \gls{snn} simulations.\\
The works reviewed in this section demonstrated that \glspl{fpga} offer flexibility, high efficiency, low-power, and high degree of parallelism, making \glspl{fpga} suitable devices for implementing brain-like circuits. Furthermore, \glspl{fpga} enable the design of complex biologically plausible neuron models and massively parallel \glspl{snn} capable of generating complex biological like patterns. Although \gls{fpga} devices being normally programmed using complex \gls{hdl} tools, \gls{hls} tools such as \gls{opencl} can be used to increase the productivity of \glspl{snn} design process by providing hardware abstraction which reduces the implementation complexity.\\

\section{Revised Literature} \label{Ch2:revised-literature} 
The works reviewed in section \ref{Ch2.1:biological} highlight that retinal cells are organised in multi-hierarchical circuits. Unlike other retinal cells, the retinal ganglion cells trigger spike events encoded and forwarded to the visual cortex via the optical nerve. \gls{oms-gc} natural circuits include both nor-spiking with spiking cells, and therefore efficient \gls{oms-gc} computational models will most likely combine non-spiking \gls{bs} models (e.g. \gls{bs} mathematical theory models) with spiking neuron models (i.e. customised \glspl{snn}). The works revised in sections~\ref{Ch2.2:snn} and \ref{Ch2.3:snn_simulators} show that \glspl{snn} are biologically plausible, capable of producing realistic patterns, are hardware-friendly and therefore suitable to be accelerated using dedicated hardware. 
\gls{snn} have a massively parallel component because they combine incorporate spiking neurons connected via numerous synapses, with differing synaptic models. The \gls{snn} parallelism introduces speed performance limitations if implemented on traditional \glspl{cpu} based on the von Neumann architecture have a massively parallel component because they combine incorporate spiking neurons connected via numerous synapses, with differing synaptic models. This \gls{snn} parallelism introduces speed performance limitations if implemented on traditional \glspl{cpu} based on the von Neumann architecture (see Chapter~\ref{Ch4:object_motion_detection_cells} for further details). 
Several \gls{omd} methods were also reviewed in section~\ref{Ch2.5:object_motion_detection}. \acrlong{omd} includes 4 phases, namely, \acrlong{bs}, noise reduction, threshold selection and moving objects detection (see Figure~\ref{fig:omd-steps}). Three main classes of \gls{bs} were identified, namely, mathematically based theories, \gls{ml}/\gls{dnn}, signal processing models. Mathematical based models were selected for this PhD research programme because these methods require lower computational requirements when compared with \gls{ml} and signal processing models. Although \gls{bs} mathematical based models methods show a good balance between robustness and processing speed, these modules can be improved using \glspl{snn} to 1) produce bio-realistic responses for neuro-engineering applications (such as eye prosthesis), 2) enhance \gls{bs} methods to filter high-frequency parasitic light variations, and 3) reduce power by reducing the dynamic power consumption (using lower speed clocks). The enhancement of the \gls{gsoc} \gls{bs} algorithm using the \glspl{snn} is reported in Chapters \ref{Ch4:object_motion_detection_cells} and \ref{Ch5:neuromorphic_object_motion_detector}. Although the \gls{omd} steps 2) noise reduction and 3) threshold selection are relevant, this PhD research programme will also explore the use of \glspl{snn} to improve the step 4) Moving objects detection (see Chapter~\ref{Ch3:mhsnn}) by combining receptive fields (acting as filters to reduce noise) and synapses with different propagation delays (short term memory) and lateral inhibition (improving the classification). 

Neuromorphic computing using \glspl{vlsi} or \glspl{fpga} have been used to accelerate for accelerating \glspl{snn} and other brain-like functions. Modern System-on-Chip solutions make use of heterogeneous computing for accelerating \glspl{snn} using \glspl{cpu} with \glspl{gpu}, \glspl{vlsi} and \glspl{fpga}. The revised literature shows that \glspl{fpga} deliver more flexibility but with higher computational complexity because the hardware developers are generally required to have a deep understanding of \gls{fpga} devices (see Chapter \ref{Ch5.2.2:fpga_architecture} and \ref{Ch5.2.3:hdl} for further details). \glspl{gpu} are more straightforward to program (developers are abstracted of the \gls{gpu} hardware architecture) than \glspl{fpga} and are suitable for accelerating classical \gls{ml}/\gls{ai} algorithms and delivering \gls{fpga} comparable speed performances. Nevertheless, the \gls{gpu} architecture has been designed to accelerate operations with 2D matrices. \glspl{vlsi} are normally the most power-efficient devices and perhaps should be faster than \glspl{fpga} and/or \glspl{gpu}, but \glspl{vlsi} have a well-defined architecture. Moreover, state-of-the-art \gls{dnn} algorithms require flexibility for changing the number of neurons, synapses, and weights and \glspl{vlsi} do not offer this flexibility. Therefore, \glspl{fpga} were selected to be used in this PhD research programme.

This PhD research programme investigated the i) use of customised \glspl{snn} for motion detection using either \gls{mhsnn} in Chapter~\ref{Ch3:mhsnn}; (ii) enhance existing mathematical theory \gls{bs} algorithms using a customised \gls{snn} targetting real-time applications in Chapter~\ref{Ch4:object_motion_detection_cells} and iii) accelerate the customised \gls{snn} using an \gls{fpga} in Chapter~\ref{Ch5:neuromorphic_object_motion_detector}.

%% file: Chapter3/chapter3.tex
%%%%%%%%%%%%%%%%%%%%%%%%%%%%%%%%%%%%%%%%%%%%%%%%%%%%%%%%%%%%%%%%%%%%%%%%%%%%%%%%
%2345678901234567890123456789012345678901234567890123456789012345678901234567890
%        1         2         3         4         5         6         7         8
% THESIS CHAPTER

\chapter{Detection of horizontal and vertical movements using Spiking Neural Networks}
\label{Ch3:mhsnn}
The \gls{mhsnn} architecture for detecting horizontal and vertical movements using a custom dataset is proposed in this chapter. Furthermore, the dataset is composed of semisynthetic image sequences of black cylinders performing rightwards, leftwards, upwards, and downwards movements, which were generated in laboratory settings. The \gls{mhsnn} architecture was designed to reflect the connectivity, behaviour, and the number of layers found in the majority of vertebrate’s retinas, including humans. The architecture was trained using 2303 images and tested using 816 images. Simulation results revealed that each cell model is sensitive to vertical and horizontal movements, with a detection error of 6.75\%.\\

\section{Introduction} \label{Ch3.1:introduction} 
\begin{figure}[htb!]
	\begin{center}
	\includegraphics[width=1.0\textwidth]{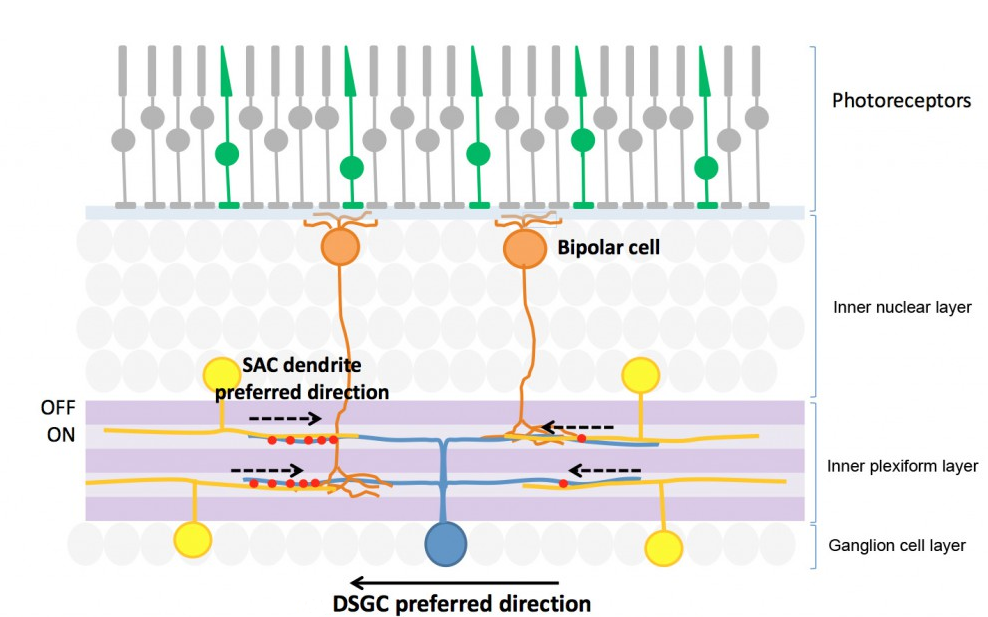}
	\end{center}
	\caption{Schematic of the \gls{dsgc} circuit. The figure shows \gls{dsgc} (in blue), \gls{sac} dendrites (in yellow), bipolar cells (in orange) and photoreceptors (green). Inhibitory synapses (in red dots) are formed on the \gls{dsgc} by the \gls{sac} dendrites (dashed arrows), which have a preferred movement in the opposite direction to the \gls{dsgc} (solid arrow) \protect\cite{Kolb2018}. \label{fig:dsgc}} 
\end{figure}

Vertebrate retinas have the ability to sense the direction of moving objects. \cite{Kolb2018}. The \gls{dsgc} detect motion direction by spiking strongly when an object moves in the preferred direction and sparsely when the identical object moves in the opposite direction \cite{im2016, Gollisch2010, kay2011}. These cells are composed of bistratified dendrites that laminate in both the On and Off of the retina's \gls{ipl}. Dendrites of glutamatergic bipolar cells supply primarily excitatory stimuli, while \acrfull{sac}, supply primarily inhibitory stimuli (see Figure~\ref{fig:dsgc}) \cite{kay2011}. 

The \gls{mhsnn} , proposed in this chapter, was designed for mimicking \gls{dsgc} basic functionalities (i.e. detection of horizontal and vertical movements) \cite{Machado_2018}. Similar to the \gls{dsgc}, the \gls{mhsnn} is also composed of 4 layers, each layer designed to deliver functionalities of specific vertebrate retinal cell (i.e. photoreceivers, bipolar, \gls{sac} and \gls{dsgc} cells). Furthermore, \gls{mhsnn} combines the use of synapses with different propagation delays for generating residual synaptic memories, and also combines inhibitory with excitatory synapses for the generation of patterns associated with the direction taken by moving objects.

\gls{codd} algorithm (described in Section \ref{codd} was designed and implemented to establish the comparison between the accuracy and computational speeds against the \gls{mhsnn} algorithm. The \gls{codd} algorithm is based on the \gls{bs} algorithms available through the \gls{opencv} library. The \gls{opencv} library is maintained by a large Open Source community (including well-known corporations like Intel, NVIDIA, Microsoft, and Google) and it is one of the most popular and reliable computer vision libraries.

The chapter structure is as follows: the proposed \gls{mhsnn} is described in Section \ref{Ch3.2:proposed_architecture}, the simulation methodology is described in section \ref{Ch3.3:methodology}, the simulation results are presented in Section \ref{Ch3.4:results} and analysis and future work are discussed in Section \ref{Ch3.5:discussion}.

\section{Proposed architecture} \label{Ch3.2:proposed_architecture}
\begin{figure}[htb!]
	\begin{center}
	\includegraphics[width=1.0\textwidth]{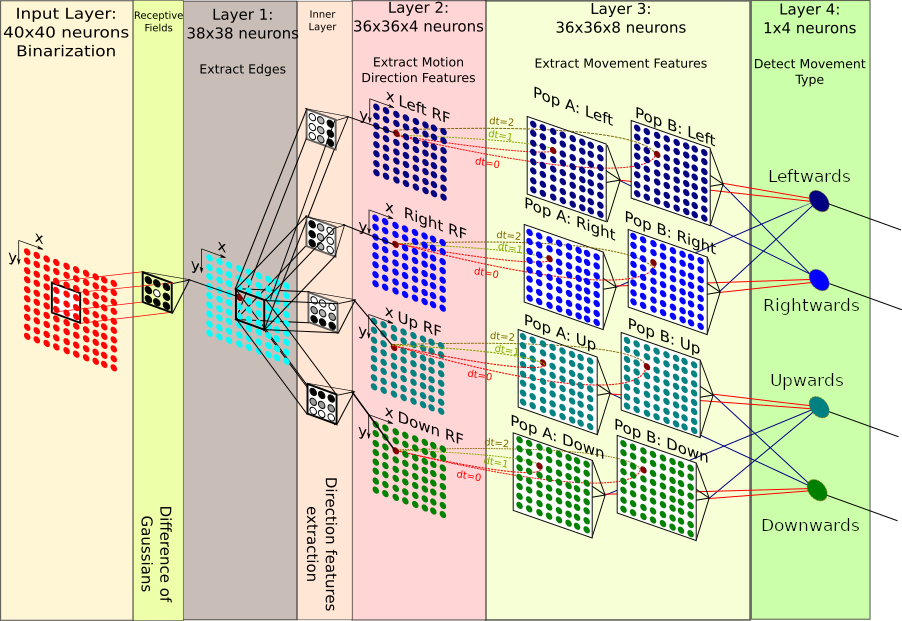}
	\end{center}
	\caption{\gls{mhsnn} with (i) $40\times40$ image input followed by the four processing Layers. Layer 1: Edge detection Layer, Layer 2: Direction features extraction, Layer 3: Movement extraction features and Layer 4: Direction-sensitive ganglion cells. \label{fig:proposed_architecture}} 
\end{figure}

The \gls{mhsnn} (see Figure~\ref{fig:proposed_architecture}) is a four-layered architecture proposed for detecting vertical and horizontal movements. The input layer converts the normalised graded pixel values (0.0 up to 1.0) to spike events (where values above 0.85\footnote{The value 0.85 was obtained empirically.} are considered a spike event); Layer 1 detects local edges; Layer 2 extracts movement direction features, Layer 3 extracts movement features, and Layer 4 detects types of movement.

\subsection{Input Layer: Binarisation via conversion from pixel grade values to spike events} \label{inLayer}
The input layer converts graded pixel values to spike events. Values equal to or above 0.85 are considered spike events, similar to the functionality of rods \citep{Kolb2003}. A 1:1 connectivity is used between each pixel and the neurons in the input layer.

\begin{figure}[htb!]
	\begin{center}
	\includegraphics[width=0.7\textwidth]{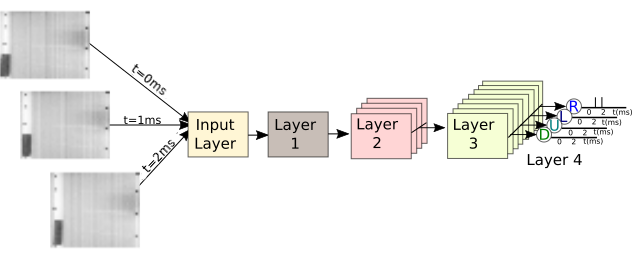}
	\end{center}
	 
	\caption{Three image frames being processed by the proposed architecture. The images are exposed to each Layer in sequence (Layer 1, 2, and 3), and finally, the movement is detected in Layer 4 by rightwards (R), leftwards (L), upwards (U) and downwards (D) \protect\gls{gc}. \label{fig:example}} 
\end{figure}

Figure~\ref{fig:example} shows an example where three sequential image frames (from a synthetic dataset with objects performing horizontal and vertical movements) of $40\times40$ pixels are presented to the Input Layer, then processed by the neurons in Layers 1 to 4. The number of required synapses and neurons is given by equations \ref{eq:1_1} and \ref{eq:1_2}.

\begin{flalign}
\label{eq:1_1}
& N_{N}=(L-2).(W-2) + 3.M_f.(L-4).(W-4) + M_f &
\end{flalign}
\begin{flalign}
\label{eq:1_2}
\notag & N_S=F_L.F_W.\left[ (L-2).(W-2) + 3.M_f.(L-4).(W-4)\right] & \\ 
 & +4.M_f.(L-4).(W-4) &
\end{flalign}

where the $N_N$ is the number of neurons, $N_S$ is the number of synapses,  L is the length of the image in pixels, W is the width of the image in pixels, $M_f$ is the number of movement features, $F_L$ is the filter length and $F_W$ is the filter width.

The decision to use small-sized images was taken based on the number of synapses and neurons that were required to implement the \gls{mhsnn}. A total of 17000 neurons and, 173700 synapses are required to build the network for processing images of $40\times40$ pixels. Overall, it takes 6 minutes and 30 seconds to build the \gls{mhsnn} architecture, 6 hours and 37 minutes for training and 7 seconds to run the simulation when using a workstation equipped with an 8-core Intel Xeon E5 2640 CPU @ 2.60GHz, 96 GB of DDR4, and 1 TB of disk space. Although 6 minutes and 30 seconds for building the network and 7 seconds for running the simulation seems to be an acceptable timing, the results shown in the Tables \ref{tab:downwards}, \ref{tab:upwards}, \ref{tab:leftwards} and \ref{tab:rightwards} show that 7 seconds is about 887\% slower than the \gls{codd} algorithm.

Figure \ref{fig:example} shows a black cylinder object moving (small displacements) rightwards. In the example, each image frame is exposed to the input layer for a simulation period of 1ms. The results are represented with vertical bars (i.e. spike events) in the output synapse of the \gls{dsgc} rightwars cell, which is represented with an R.

\subsection{Layer 1: Edge detection}\label{Layer1}
The aim of Layer 1 is to detect edges.
$40\times40$ pixels images are exposed to Layer 1 for a period of 1 simulation time step. 
The neurons in Layer 1 receive spike events from a 3$\times$3 patch where the central neuron is connected via an excitatory synapse (weight greater than 0), and the eight neighbouring neurons are connected via inhibitory synapses (weight lower than 0), originating a \gls{rf}. The \gls{rf} in the visual system comprises a 2D region in a specific visual space and may have different sizes and shapes. The latter is represented in Figure~\ref{fig:proposed_architecture} by windows composed of 8 black circumferences (acting as inhibitory synapses) and a white central circumference (acting as an excitatory synapse). The \glspl{rf} have a stride of 1 for retaining the spatial information required for performing the edge detection of a \gls{rf} weights' distribution given by the \gls{dog} function \cite{Deng1993}. The \gls{dog} function, commonly used to perform edge detection\cite{Deng1993}, in Layer 1 is used to perform the edge detection and is governed by equation \ref{eq:8}.
\begin{flalign}
\label{eq:8}
& DoG(x,y)=\dfrac{1}{2 \pi \sigma _{s}^{2}}e^{-\dfrac{x^2+y^2}{2\sigma _{s}^{2}}}-\dfrac{1}{2\pi \sigma_{c}^{2}}e^{-\dfrac{x^2+y^2}{2\sigma_{c}^{2}}} & \\
& \dfrac{\sigma_c}{\sigma_s}=1.6 &
\end{flalign}

Because of the border conditions, the number of neurons per pixel is reduced by two columns and two rows (i.e., the required number of neurons is reduced by $38 \times 38$). Zero or negative values are produced by the \gls{dog} filter when a given patch is composed of similar pixel intensities, preventing neurons from generating spike events; Positive values are produced when the neighbouring pixels and the central pixels have a considerable variation, triggering neurons to generate spike events.

\subsection{Layer 2: Horizontal and vertical features extraction} \label{Layer2}

The aim of Layer 2 is to extract horizontal and vertical features.

Layer 2 is composed of 36 rows$\times$ 36 columns$\times$4 groups of neurons (36$\times$36$\times$4). Each neuron is connected via a 3$\times$3 \gls{rf}, and the connection between neurons is performed via inhibitory and excitatory synapses. The excitatory synapses between the neurons of Layer 1 and Layer 2 vary according to the type of filter used. The goal of each filter type is to generate unique spike patterns that are required for detecting the movement type while retaining spatial information. The filters in the inner Layer are represented in Figure~\ref{fig:proposed_architecture} with nine blue circles overlapping nine neurons in Layer 1.

The neurons in Layer 2 are used to extract features related to each type of movement (rightwards, leftwards, upwards and downwards). Again, 3$\times$3 windows are used to produce features maps that capture the required details for a specific movement and distinct pattern from the homologous movement.

\begin{flalign}
\label{eq:4}
& L_f \left( x,y\right)=\displaystyle \sum_{i=0}^{x} \sum_{j=0}^{y} \frac{x}{2}-i & \\
& R_g \left( x,y\right)=\displaystyle \sum_{i=0}^{x} \sum_{j=0}^{y} \dfrac{-x}{2}+i &
\end{flalign}
\begin{flalign}
\label{eq:6}
& U_p \left( x,y\right)=\displaystyle \sum_{j=0}^{y} \sum_{i=0}^{x} \dfrac{-y}{2}+j & \\
& D_w \left( x,y\right)=\displaystyle \sum_{j=0}^{y} \sum_{i=0}^{x} \dfrac{y}{2}-j &
\end{flalign}

where the parameters $L_f$, $R_g$, $U_p$ and $D_w$ represent rightwards, leftwards, upwards, and downwards filter weight distribution. The number of neurons per patch of layer neurons is reduced by two columns and two rows (i.e.. The required number of neurons is 36$\times$36) per filter type because of the border conditions. Like the \gls{dog} filters, neurons will trigger spike events when the current values generated by the patch of neurons in Layer 1 is positive and greater or equal to the threshold.

\subsection{Layer 3: Extraction of movement features} \label{Layer3}

The aim of Layer 3 is to extract horizontal and vertical movement features. Layer 3 neurons are connected in a one-to-one (1:1) configuration to the Layer 2 neurons. There are two neuron populations, Population A and Population B. 

The Population A neurons are wired via two synapses, an excitatory synapse with no delay, and an inhibitory synapse with a 1ms delay. The use of synapses with different delays facilitates movement feature extraction \citep{Taherkhani2015}. In addition, the delay produced by synapses with varying times of propagation creates a short-term memory. E.g. given a pre-neuron that spiked at the timestep $t$ and did not spike at timestep $t-1$ causes a spike event in the post-neuron because the difference between spike events is positive and above the threshold.

Population B differs from Population A because the inhibitory synapses are delayed by two simulation time steps instead of one. The two populations are used for improving the accuracy of the movement features extraction by creating a bio-inspired buffer of 3 spike patterns. Each group of neurons comprises groups of 36$\times$36 neurons, and in the overall configuration, there are eight groups of neurons (four per population). The neurons in layer three can sense movement because they are interconnected using different propagation delays. Such delays create a local residual memory for holding past neurons' events and comparing them with current events (the simulation time-step was set to 1ms). In addition, the populations are used to improve the robustness of the motion detection because fast movements are detected by Population A, slower signs by Population B and average speed movements are seen by both populations of neurons. Therefore, a movement feature is extracted if a spike in the current frame was not detected in the previous frame (population A with a propagation delay of one) or two frames before (population B with a propagation delay of 2). 

Therefore, it is possible to generate different spiking patterns to detect differences between spike trains. The neurons in Layer 3 will spike if changes are detected.

\subsection{Layer 4: Detection of movement type} \label{Layer4}
The aim of Layer 4 is to detect horizontal and vertical movements' directions (i.e. rightwards, leftwards, downwards, and upwards) based on the movement features extracted in Layer 3. The horizontal and vertical movements were inspired by the basic functionalities exhibited by \gls{dsgc} \cite{Gollisch2010,Kolb2018}. Each neuron in Layer 4 receives connections from all the neurons of population A and B, of its specific type of movement (\textit{e.g.} right movement cell is connected via excitatory cells to the left population's A and B) and inhibitory synapses from its type of movement (\textit{e.g.} right movement cell is connected via inhibitory cells to the right population's A and B). The reason for having these connections is that the left cell must not spike when the right cell is spiking, or vice-versa. It is possible in certain circumstances to have different types of cells spiking at the same time (\textit{e.g.} a simultaneous rightwards and upwards). However, a particular cell cannot spike at the same time as its pairing cell (\textit{e.g.} the right cell cannot spike at the same time as the left cell because that would mean that a given object was moving rightwards and leftwards at the same time). In Figure~\ref{fig:proposed_architecture}, the inhibitory synapses are represented with dashed red lines, and the inhibitory synapses with blue lines. The brown and yellow dashed boxes indicate that all the neurons are connected to the movement-sensitive cells.

This layer contains four types of neurons, two of which respond to horizontal movements and the other two to vertical movements. These neurons receive connections from the neurons of both populations, A and B, of the same type and its pairing (\textit{e.g.} right/left). The connections from the same kinds of movements are excitatory, while the connections from the paring type are inhibitory.

The weights of these connections were trained using the \gls{resume} algorithm proposed in \citep{Ponulak2010}.

\gls{resume} \cite{Ponulak2010}, a supervised learning algorithm, was used to train the Layer 3 neurons' response. In \gls{resume}, teacher signals are used to produce the desired spike pattern in response to a stimulus \cite{Ponulak2010}. 

\begin{figure} [htb!]
	\begin{center}
	\includegraphics[width=1.0\textwidth]{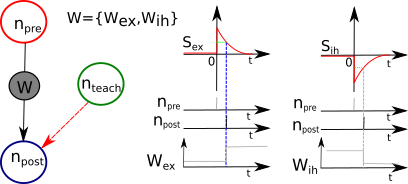}	
	\end{center}
	\caption{\gls{resume} learning: (left) Remote supervision. (right) Learning windows \protect\cite{Ponulak2010}. \label{fig:resume}}
\end{figure}

Figure~\ref{fig:resume} shows the teacher signal (desired spike pattern) $n_{teach}$ being presented to a neuron $n_{post}$ for delivering a spike pattern by adjusting the synaptic weight w between the pre-neuron $n_{pre}$ and the post-neuron $n_{post}$. The learning occurs with the modification of the weights.
 
The \gls{resume} \cite{Ponulak2010} equations are as follows:
\begin{align} [h]
\label{eq:5}
W_{ex}(s_{ex})=
  \begin{cases}
  A_{ex} e^{\left( \frac{-s_{ex}}{\tau_{ex}}\right)}, & \text{if }s_{ex}>0,\\
  0, & \text{if } s_{ex} \leq 0,
  \end{cases}\\
W_{ih}(s_{ih})=
  \begin{cases}
  A_{ih} e^{\left( \frac{-s_{ih}}{\tau_{ih}}\right)}, & \text{if }s_{ih}>0,\\
  0, & \text{if } s_{ih} \leq 0,
  \end{cases}
\end{align}

where $A_{ex}$, $A_{ih}$, $\tau_{ex}$ and $\tau_{ih}$ are constants. $A_{ex}$ and $A_{ih}$ are positive in excitatory synapses and negative in inhibitory synapses. In both cases, $\tau_{ex}$ and $\tau_{ih}$ are positive time constants \cite{Ponulak2010}.

Layer 4 neurons were trained using teacher signals. The teacher signal is generated using one teacher neuron ($n_{teach}$) connected to a Layer 4 neuron.  To generate a spike event, a high current is applied to $n_{teach}$ which makes the Layer 4 neuron generate a spike event. This interactive process enables the weights of synapses coming from Layer 3 neurons to increase over time. The training was initially done on the horizontally sensitive cells and then on the vertically sensitive cells.

\begin{figure}[H]
	\begin{center}
	\includegraphics[width=1.0\textwidth]{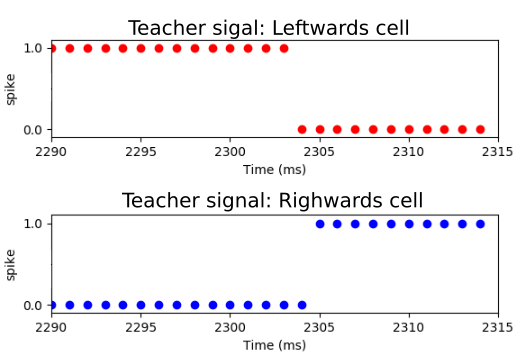}
	\end{center}
	\caption{Two teacher signals used to train the horizontal sensitive cells during the simulation time window [2290, 2315]ms.} \label{fig:teacher}
\end{figure}

In Figure~\ref{fig:teacher} the two teacher signals used to train the synaptic weights of the horizontal cells during the simulation time window [2290, 2315]ms are shown. The period [2290, 2315]ms refers to a time window where a black cylinder object is moving leftwards during the period [2290, 2303]ms rightwards during the period [2304, 2315]ms. 

\section{Implementation} \label{Ch3.3:methodology}
This section covers the details of how the dataset was generated and preprocessed, and the methodology that was followed for evaluating the performance of the proposed \gls{mhsnn} for detecting horizontal and vertical movements.

\subsection{Dataset} \label{dataset}

A simple dataset containing 100 repetitions of a black cylindrical object moving rightwards. Each repetition comprises approximately 30 images (the number of images is proportional to the object speed). A total of 3120 images were generated from one hundred trials. The dataset was augmented by performing rotations of $\frac{\pi}{2}$ rad (downward movements), $\pi$ rad (rightward movements), and $\frac{2\pi}{3}$ rad (upward movements). In addition, all the images were annotated with the type of movement being described. The original size of each figure is 640$\times$480 pixels.

\subsection{Image pre-processing} \label{preprocessing}

Natural images extracted from batches of image sequences require preprocessing to reduce the dimension of the images and lower the number of neurons per layer. Therefore, the image is first converted to greyscale and the number of channels is reduced from 3 (red, green, and blue) to one (greyscale) by applying equation \ref{eq:2} \cite{OpenCV2021a}.

\begin{eqnarray}
\label{eq:2}
& Grey=(0.299 \times R)+(0.587 \times G)+(0.114 \times B)
\end{eqnarray}

The second preprocessing step was to compute the \gls{pca} whitening to reduce the amount of redundant input. Through the application of \gls{pca} whitening, we minimise the degree of correlation between adjacent pixels or feature values, which might be highly correlated \citep{Kessy2015}.

The \gls{pca} and whitening were done using the Python 3.8 Sklearn library\footnote{Available online, \protect\url{https://scikit-learn.org/stable/modules/generated/sklearn.decomposition.PCA.html}, last accessed: 22/06/2022}. The third and final step was resizing the image to reduce the number of required neurons per layer using the OpenCV library built-in functions for Python 3.8. The image frames were resized to $40\times40$ pixels while keeping the original aspect ratio.

\begin{figure}[H]
	\begin{center}
	\includegraphics[width=1.0\textwidth]{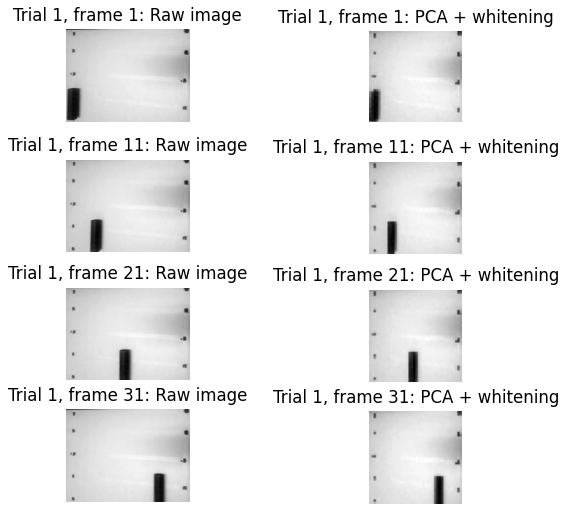}
	\end{center}
	\caption{Sequence of 4 raw images, where a black cylinder object is moving rightwards ($1^{st}$ column); image after pre-processing steps namely, conversion from RGB to greyscale, resizing, \gls{pca} and whitening ($2^{nd}$ column).} \label{fig:videoclip}
\end{figure}

For example, figure~\ref{fig:videoclip} shows a sequence of 4 images extracted from one of the trials, where the black cylindrical object is moving rightwards.

\begin{figure}[H]
	\begin{center}
	\includegraphics[width=1.0\textwidth]{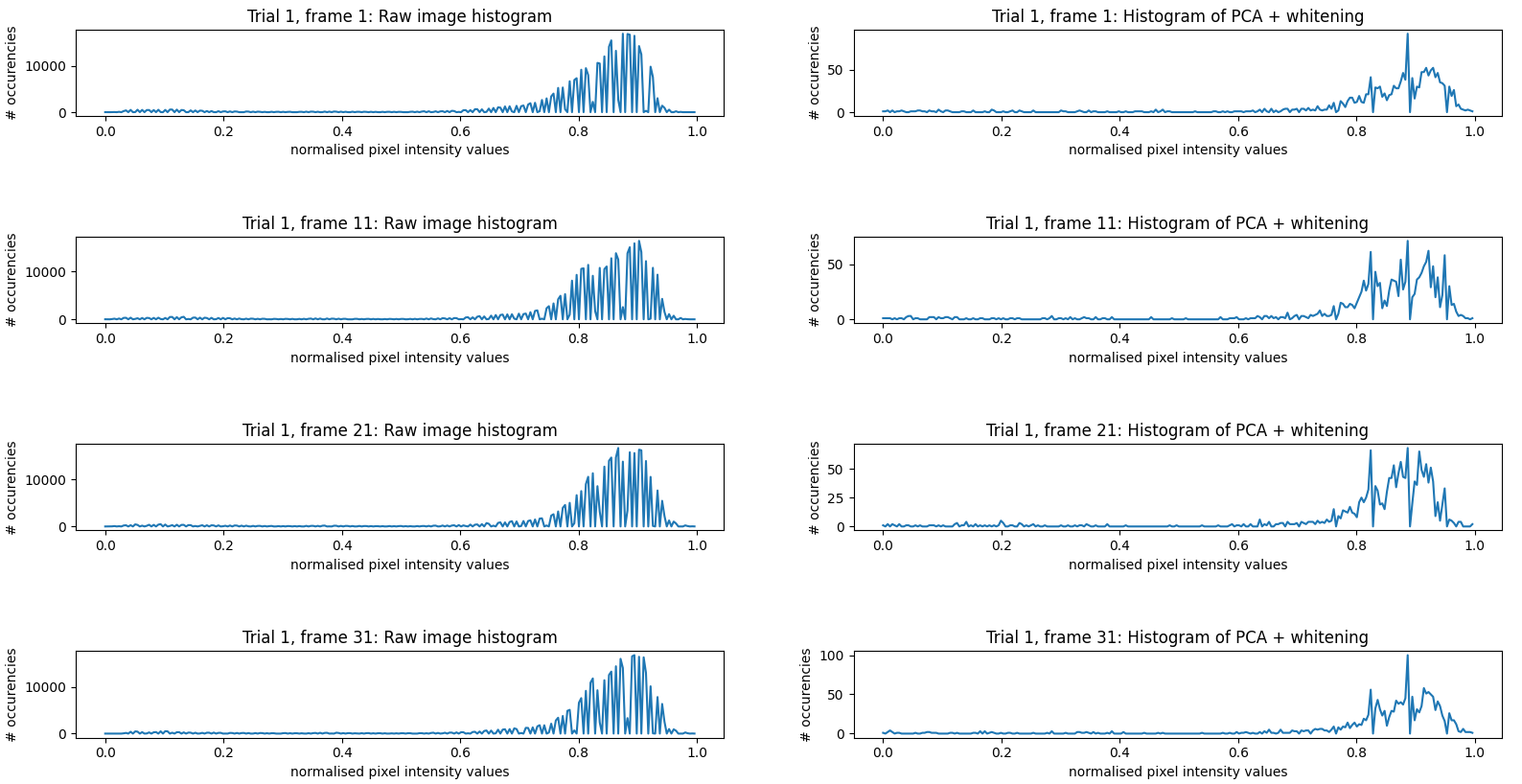}
	\end{center}
	\caption{Histograms of the sequence of the images, shown in Figure \protect\ref{fig:videoclip}. The histograms of pre-processed images are shown in the $1^{st}$ column and histograms of the post-processed images are shown in the $2^{nd}$ column.} \label{fig:histo}
\end{figure}

Figure \ref{fig:histo} depicts the histograms of the pre- and post-processed images. Compared with histograms from the first-column, the histograms from the second column show a dimensionality reduction due to using the \gls{pca} and, therefore, reducing the local correlation between pixels and speed performance.

The results labelled with MHSNNv1 were obtained when tested against the dataset after the pre-processing stage 1 (i.e., image conversion from colour to greyscale), whereas the MHSNNv2 results were obtained against the dataset after the pre-processing steps 1 and 2 (i.e., image conversion from colour to greyscale followed by the \gls{pca} whitening).

\subsection{Simulation Process} \label{simulation}

The simulation was performed using the Brian2 \gls{snn} simulator\footnote{The Brian2 SNN simulator. \url{http://brian2.readthedocs.io/en/stable/index.html}, last accessed: 31/01/2018}.

\textbf{Step 1 - Preparing the simulation:} The image sequences were randomly split into four dataset subsets, each comprising different training and testing batches. A total of 75 batches of image sequences (2303 images) were used for training and the remaining 25 batches of image sequences (816 images) for testing per movement type. The batches of image sequences were all loaded into memory before starting the training/simulation.

\textbf{Step 2 - Setting the simulation parameters:} The simulation, neuron, and synapses parameters were set in accordance with the Brian 2 documentation \cite{Brian2021}. The simulation was configured with a time step of $\tau=1 ms$. The following parameters were set for all neurons: $u_{reset}=-1mV$, $\tau_{refractory}=0$. The threshold for neurons in Layer 1, 2 and 3 was set to $u_{th}=0.0mV$ and for Layer 4 $u_{th}=30.0mV$. The weights for Layers 1, 2 and 3 were set constant. The neurons in Layer 4 had their weights trained using the \gls{resume} algorithm, with the following parameters: $\tau^d=20ms$, $\tau^l=20ms$,  $A^d=0.01$, $A^l=0.01$ (for excitatory synapses) and $A^d=-0.01$, $A^l=-0.01$ (for inhibitory synapses). 

\textbf{Step 3 - Simulation stages:} The simulation can be divided into two stages, namely training and testing. 

\begin{itemize}

\item \textbf{Training Stage:} During the simulation training stage, a teacher signal is used to train each cell's weights. The weights of all synapses connected to all the Layer 4 neurons were initialised with 1.0. The training is repeated 10,000 times\footnote{The value 10,000 was obtained experimentally.} on the training batches, while the weights are updated constantly.

\item \textbf{Testing Stage:} The testing mode is repeated once on the test batches, while the weights are kept constant. In the test mode, \gls{resume} is not used, and the outputs of Layer 4 are scored against the expected results.

\end{itemize}

\subsection{Custom Object Direction Detection algorithms} \label{codd}

\begin{figure}[H]
	\begin{center}
	\includegraphics[width=0.5\textwidth]{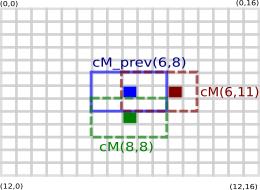}
	\end{center}
	\caption{Detection of the object movement direction. The figure represents an image of $13\times17$ where a given object (blue rectangle) performs rightwards (dashed brown rectangle) and downwards (dashed green rectangle) movements. In the case of the brown rectangle, $cM(j)=11$, $cM_{prev}(j_{prev})=8$ and therefore the object is moving rightwards because $cM(j)>cM_{prev}(j_{prev})$. While for green rectangle, $cM(i)=8$, $cM_{prev}(i_{prev})=6$ and therefore, the object is moving downwards because $cM(i)>cM_{prev}(i_{prev})$.} \label{fig:detect_mov_direction}
\end{figure}

The \gls{opencv}'s \gls{bs} algorithms (i.e. \gls{mog}, \gls{mog2}, CNT, \gls{knn}, GMG, \gls{lsbp} and \gls{gsoc}) are highly efficient algorithms designed for modelling the dynamic background changes (i.e. about two hundred frames are required to train the background model) and classifying all the outliers as foreground. The \gls{codd} algorithm makes use of the \gls{opencv}'s \gls{bs} algorithms for extracting the foreground from the background. The direction of the movement is obtained by computing the centre of mass ($cM$) from the foreground blob with the coordinates $(i,j)$ and comparing cM to the centre of mass from the previous frame ($cM_{prev}$) with the coordinates $(i_{prev}$,$j_{prev})$. When detecting vertical motion, the object moves upwards when $cM(i)<cM_{prev}(i_{prev})$, remains stationary when $cM(i)=cM_{prev}(i_{prev})$ or moves downwards when $cM(i)>cM_{prev}(i_{prev})$. Similarly, when detecting horizontal motion, the object moves leftwards when \\
$cM(j)<cM_{prev}(j_{prev})$, remains stationary when $cM(j)=cM_{prev}(j_{prev})$ or moves rightwards when\\
$cM(j)>cM_{prev}(j_{prev})$. Algorithm \ref{alg:codd} describes the  \gls{codd} algorithm (see Figure~\ref{fig:detect_mov_direction}). 

\begin{algorithm}
 \caption{\gls{codd} algorithm pseudocode}
 \label{alg:codd}
 \begin{algorithmic}[1]
 \SWITCH {backgound\_subtraction\_alg}
    \CASE {MOG}
      \STATE $bgsub=createOpenCVBackgroundSubtractorMOG$
    \ENDCASE
    \CASE {MOG2}
      \STATE $bgsub=createOpenCVBackgroundSubtractorMOG2$
    \ENDCASE
    \CASE {CNT}
      \STATE $bgsub=createOpenCVBackgroundSubtractorCNT$
    \ENDCASE
    \CASE {KNN}
      \STATE $bgsub=createOpenCVBackgroundSubtractorKNN$
    \ENDCASE
    \CASE {GMG}
      \STATE $bgsub=createOpenCVBackgroundSubtractorGMG$
    \ENDCASE
    \CASE {LSBP}
      \STATE $bgsub=createOpenCVBackgroundSubtractorLSBP$
    \ENDCASE
    \CASE {GSOC}
      \STATE $bgsub=createOpenCVBackgroundSubtractorGSOC$
    \ENDCASE
    \ENDSWITCH
    \STATE $set\_movement\_type$
    \FOR{k:=calibration \TO classification}
        \FOR{i:=0 \TO number\_images}
            \STATE $get\_image\;$
            \STATE $img=apply\_bgsub$
            \IF{classification}
                \STATE $binarise\_img\_and\_normalise\_img$
                \STATE $compute\_cM$
                \IF{$i=0$}
                    \STATE $cM_{prev}=cM$
                \ENDIF
                \IF{$movement\_type=horizontal$}
                    \STATE $direction=cM(j)-cM_{prev}(j_{prev})$
                \ELSE
                    \STATE $direction=cM(i)-cM_{prev}(i_{prev})$
                \ENDIF
               \IF{$direction=expected\_direction$}
                    \STATE $TP=TP+1$
                \ELSE
                    \STATE $FP=FP+1$
                \ENDIF
            \ENDIF
        \ENDFOR
        \IF{classification}
            \STATE store\_results\;
        \ENDIF
    \ENDFOR
\end{algorithmic}
\end{algorithm}

\subsection{Metrics} \label{metrics}
The following metrics ruled by equations \ref{eq:pcc} and \ref{eq:pwc} were used to compute the \gls{pcc} and \gls{pwc} using the \gls{tp} and \gls{fp}.

\begin{eqnarray}
& \gls{pcc}= \frac{\gls{tp}}{\gls{tp}+\gls{fp}}.100\\\label{eq:pcc}
& \gls{pwc}= \frac{\gls{fp}}{\gls{tp}+\gls{fp}}.100\label{eq:pwc}
\end{eqnarray}

\section{Results} \label{Ch3.4:results}
The voltage threshold of the Layer 1 neurons was obtained by computing the average of all the images used for training. The Brian2 simulator has a structure called "TimedArray" used to expose the images to the first layer of neurons. Images are converted into 1D vectors, and each image is a row of "TimedArray". 

A vector of neuron indexes was used to retain the spatial information between image pixels and neurons and the desired connectivity between neurons. The Brian2 exposes each row (image) after the predefined time step (The \gls{mhsnn} uses a 1ms time step).

Spike monitors were configured and used for tracking all the spikes generated by layer and plotted at the end of the simulation. The synaptic weights were updated with the output generated by the \gls{resume} training algorithm during each training mode iteration in the training mode.

\subsection{Horizontal movement test} \label{horizontal_movements}

\begin{figure}[H]
	\begin{center}
	\includegraphics[width=1.0\textwidth]{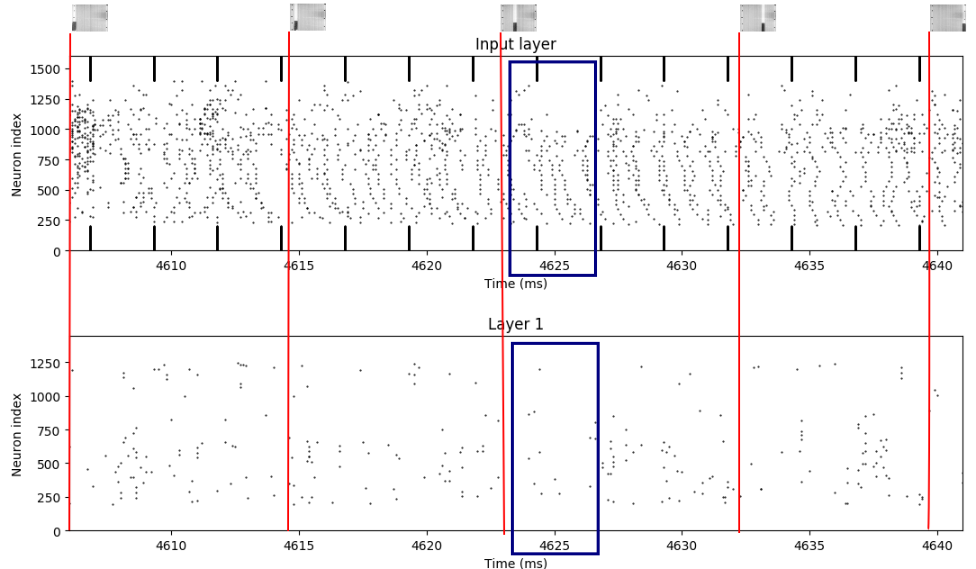}
	\end{center}
	\caption{Raster plot of the spiking pattern obtained during the period [4605,4640]ms (a black cylinder object was moving rightwards) and generated by the input layer (after converting the graded values into spikes) and Layer 1 (edge extraction) neurons. The blue rectangle is used to track the spike events generated during the period [4624,4625]ms.} \label{fig:inl1Horizontal}
\end{figure}

\begin{figure}[H]
	\begin{center}
	\includegraphics[width=1.0\textwidth]{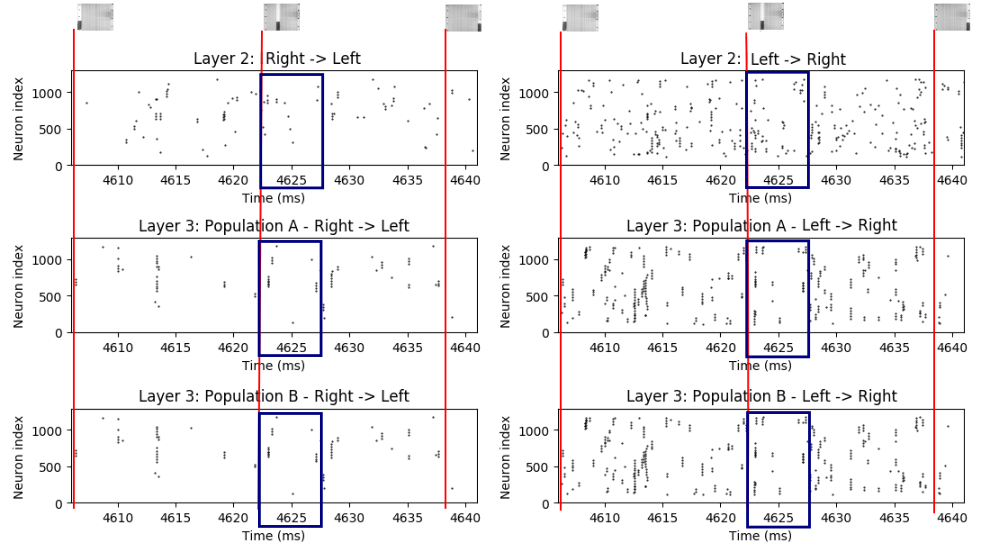}
	\end{center}
	\caption{Raster plot of the spikes obtained during the period [4605,4640]ms (a black cylinder object was moving rightwards) and generated by the neurons in Layers 2 and 3. The blue rectangle is used to track the spike events generated during the period [4624,4625]ms.} \label{fig:l2l3Horizontal}
\end{figure}

\begin{figure}[H]
	\begin{center}
	\includegraphics[width=1.0\textwidth]{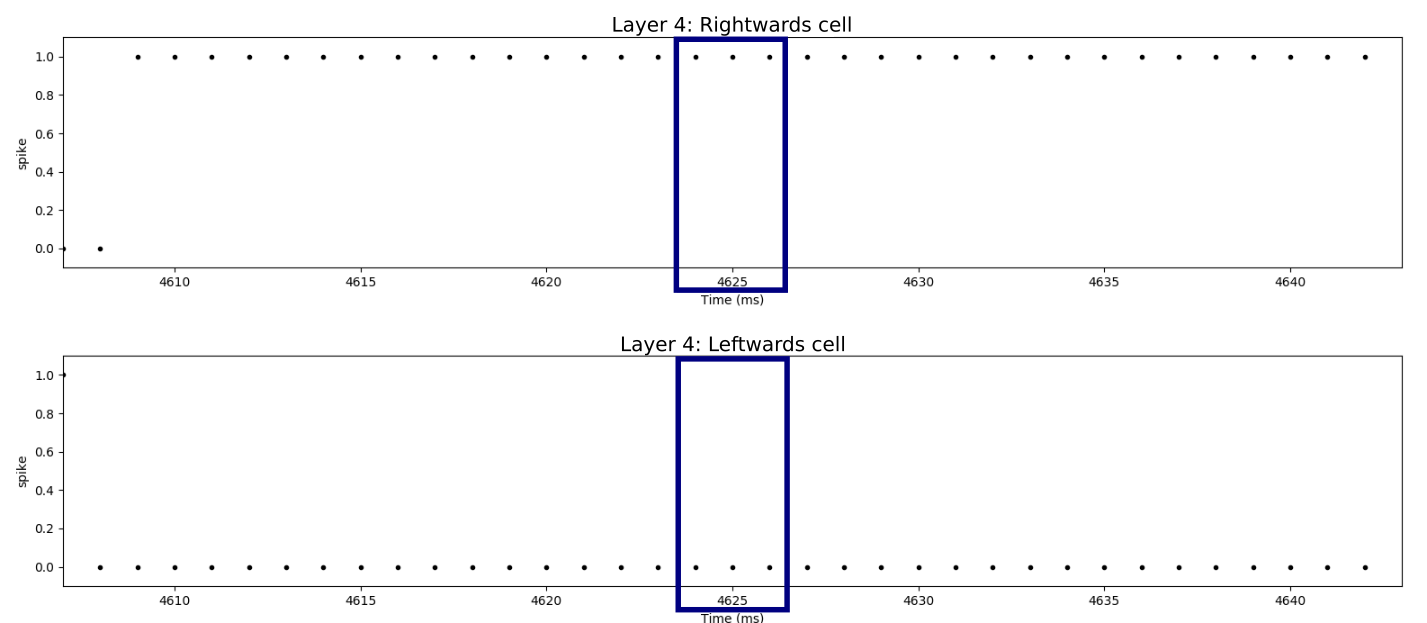}
	\end{center}
	\caption{Raster plot of the spikes pattern obtained during the period [4605,4640]ms (a black cylinder object was moving rightwards) and generated by the horizontal sensitive cells. The blue rectangle is used to track the spike events generated during the period [4624,4625]ms.} \label{fig:l4Horizontal}
\end{figure}

The results obtained from the horizontal test sequences are shown in Figures~\ref{fig:inl1Horizontal}, \ref{fig:l2l3Horizontal} and \ref{fig:l4Horizontal}.

Referring to Figure~\ref{fig:inl1Horizontal}, during the period [4605,4640]ms the image was performing a movement rightwards. The input layer illustrates the graded values after the conversion to spike events (values above 0.85 are considered spike events). Layer 1 shows the spike pattern generated from the edge extraction. The spiking pattern in Figure~\ref{fig:inl1Horizontal} was generated during test mode (after training).

Referring to Figure~\ref{fig:l2l3Horizontal}, the spike activity from the right cells is more prominent than the left cells, which is a consequence of the type of movement. The spike patterns obtained in population A ($frame[t]-frame[t-1]$) are very similar to the ones obtained from population B ($frame[t]-frame[t-2]$), which is a consequence of having slow movements that are detected by both populations of neurons.

Figure~\ref{fig:l4Horizontal} shows that each cell is spiking at the expected time. 
The \gls{pwc} was 7\% for the horizontal cells. The scoring algorithm compares the output result with the expected result, and the error counter is increased by one every time an error is detected (i.e. false positives or false negatives). The error is associated with a sudden change of image sequences. This phenomenon occurs when the last image of a given batch is followed by the first image of a new batch. This situation triggers a different spike pattern compared with the previous spike patterns (Layer 3 populations A and B).

\subsection{Vertical movement test} \label{vertical_movements}

\begin{figure}[H]
	\begin{center}
	\includegraphics[width=1.0\textwidth]{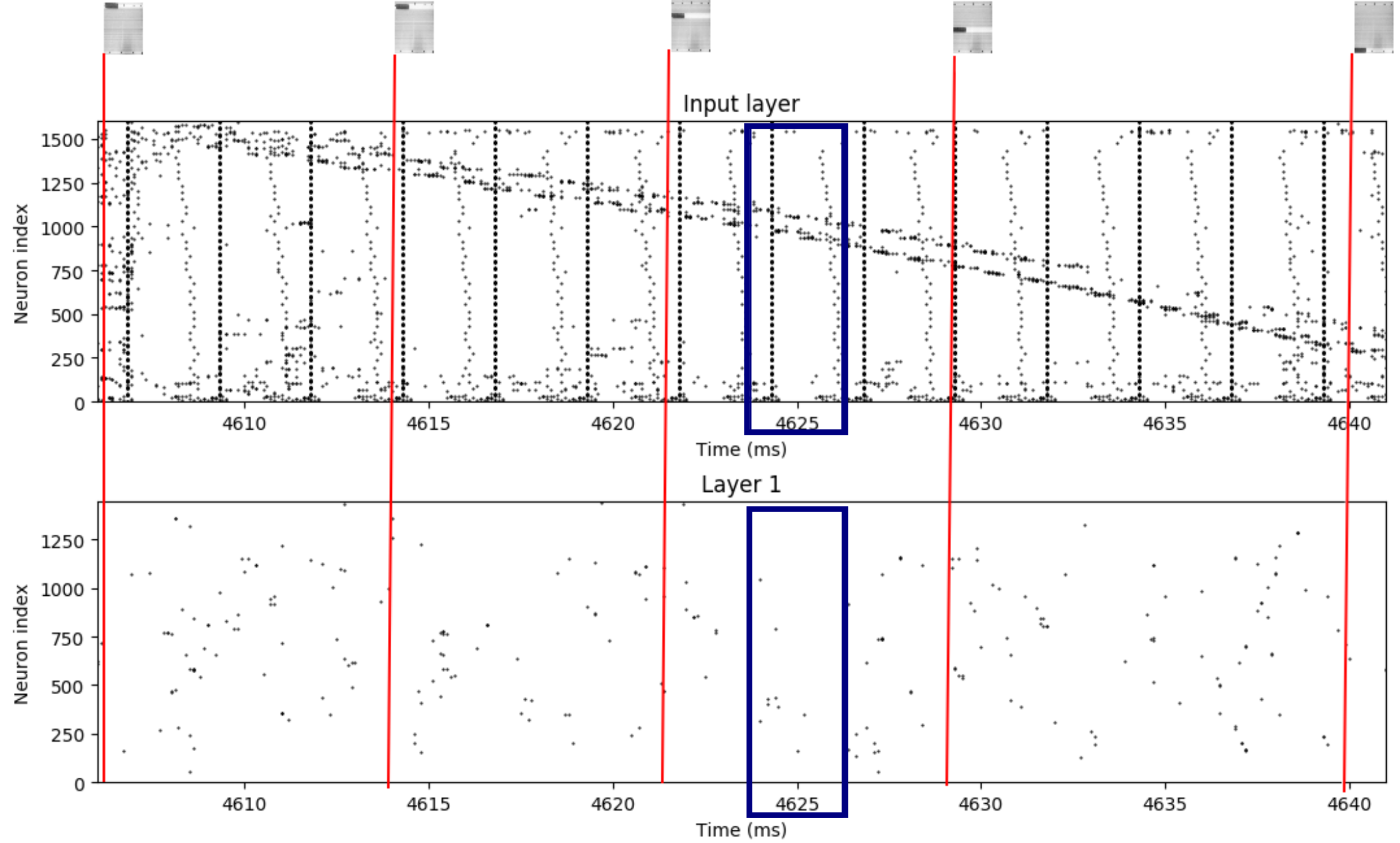}
	\end{center}
	\caption{Raster plot of the spikes obtained during the period [4605,4640]ms of the vertical test and generated by the input layer (after converting the graded values into spikes) and Layer 1 neurons. The blue rectangle is used to track the spike events generated during the period [4624,4625]ms.} \label{fig:inl1Vertical}
\end{figure}

\begin{figure}[H]
	\begin{center}
	\includegraphics[width=1.0\textwidth]{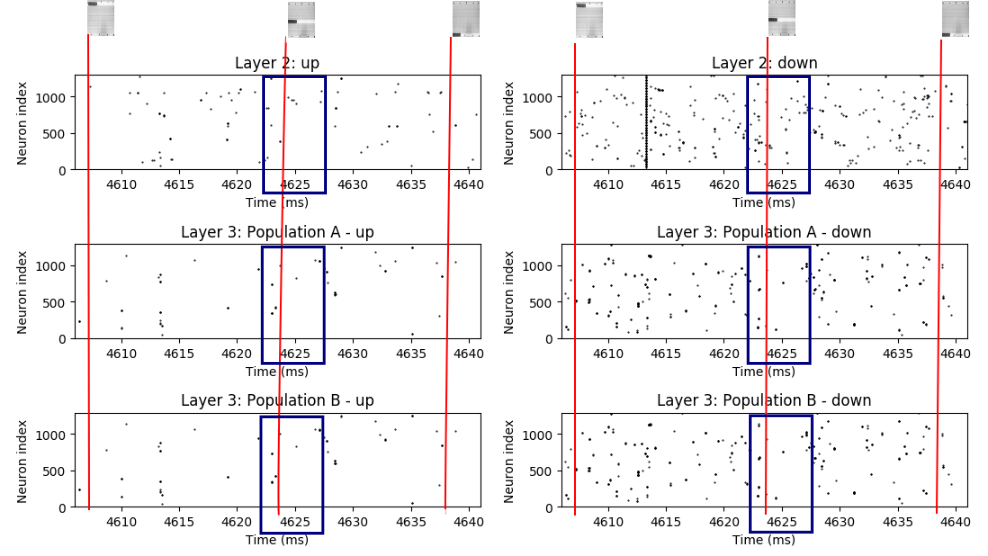}
	\end{center}
	\caption{Raster plot of the spiking pattern obtained during the period [4605,4640]ms of the vertical test and generated by the neurons in Layers 2 and 3. The blue rectangle is used to track the spike events generated during the period [4624,4625]ms.} \label{fig:l2l3Vertical}
\end{figure}

\begin{figure}[H]
	\begin{center}
	\includegraphics[width=1.0\textwidth]{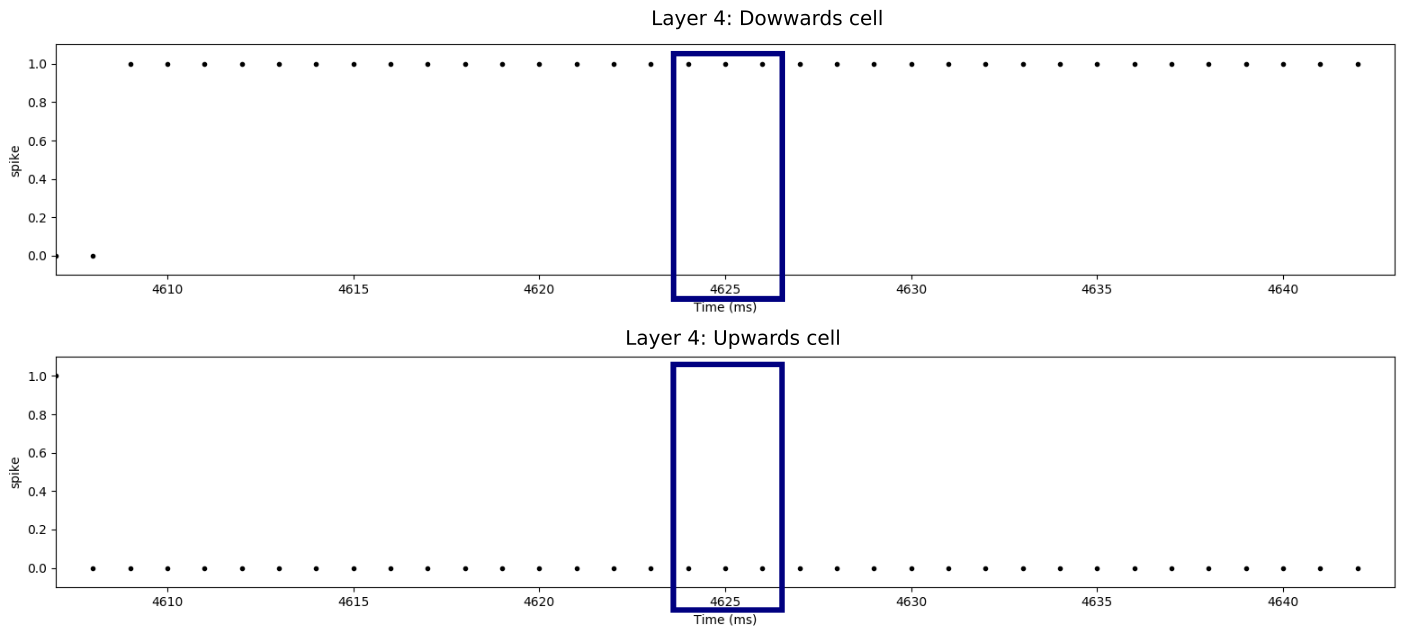}
	\end{center}
	\caption{Raster plot of the spiking pattern obtained during the period [4605,4640]ms of the vertical test and generated by the vertical sensitive cells. The blue rectangle is used to track the spike events generated during the period [4624,4625]ms.} \label{fig:l4Vertical}
\end{figure}

The results obtained from the vertical test sequences are shown in Figures~\ref{fig:inl1Vertical},  \ref{fig:l2l3Vertical} and \ref{fig:l4Vertical}.

Referring to Fig.~\ref{fig:inl1Vertical}, during the period [4605,4640]ms the object is moving downward. The input layer shows the graded values after the conversion to spike events (values above 0.85 are considered spike events). In Layer 1 the spike pattern generated from the edge extraction is shown. The spiking pattern in Figure~\ref{fig:inl1Vertical} is clearly distinct from the spike pattern in Figure~\ref{fig:inl1Horizontal}. It is possible to infer from the spike pattern of the input layer that the object is moving downward.

Referring to Fig.~\ref{fig:l2l3Vertical}, the spike activity of the down cells is more prominent than the up cells, which is a consequence of the type of movement. Again, the spike patterns obtained from population A ($frame[t]-frame[t-1]$) are very similar to those obtained from population B ($frame[t]-frame[t-2]$), which is a consequence of having slow movements that are detected by both populations of neurons.

Figure~\ref{fig:l4Vertical} shows that each cell is spiking at the correct time because the first set of images are upwards movements and then downwards movements. It is also seen in Figure~\ref{fig:l4Vertical} that the right cell generated spike events, while the left cell did not generate spike events. The \gls{pwc} was 6.5\% for the vertical cells. The errors occur when the last image of a sequence is followed by the first image of a new sequence (i.e. when a new sequence starts and the object moves from the end position to the start position of two sequential images).

\subsection{Results per category} \label{results per category}

The \gls{mhsnn} was tested against the dataset before applying \gls{pca} whitening (MHSNNv1) and after applying \gls{pca} whitening (MHSNNv2) (see Section \ref{dataset} for further details). Although the \gls{codd} variants (i.e. MOG, MOG2, CNT, GMG, KNN, LSBP, GSOC) were also tested against the dataset before applying \gls{pca} whitening, the \gls{pcc}, \gls{pwc} and processing results were the same for the dataset after applying \gls{pca} whitening.
Table \ref{tab:leftwards} shows leftwards movements classification and processing time per method.

\begin{table}[H]\caption{Leftwards movements classification and processing time per method.}\label{tab:leftwards}
\resizebox{14.5cm}{!}{
\begin{tabular}{|c|c|c|c|c|}
\hline
Method & movement type & \gls{pcc} & \gls{pwc} & processing time \\ \hline\hline
MHSNNv2 & leftwards & 93.6\% & 6.4\% & 1731.352 ms \\ \hline
\gls{codd}-MOG & leftwards & 90.5\% & 9.5\% & 1.304 ms \\ \hline
\gls{codd}-KNN & leftwards & 90.1\% & 9.9\% & 1.385 ms \\ \hline
MHSNNv1 & leftwards & 87.0\% & 13.0\% & 1760.069 ms \\ \hline
\gls{codd}-MOG2 & leftwards & 86.2\% & 13.8\% & 1.195 ms \\ \hline
\gls{codd}-CNT & leftwards & 61.1\% & 38.9\% & 1.4 ms \\ \hline
\gls{codd}-GSOC & leftwards & 46.1\% & 53.9\% & 3.427 ms \\ \hline
\gls{codd}-LSBP & leftwards & 44.4\% & 55.6\% & 3.665 ms \\ \hline
\gls{codd}-GMG & leftwards & 0.0\% & 100.0\% & 1.655 ms \\ \hline
\end{tabular}}
\end{table}

The results show that the MHSNNv2 shows the best results in terms of highest \gls{pcc} and lowest \gls{pwc} when compared with the other \gls{codd} variants. Nevertheless, the \gls{codd}-MOG shows the second-best result with an average processing time of 1.304ms which is much faster than the MHSNNv2. It is also possible to infer that the \gls{pca} whitening has contributed to an improvement of the \gls{mhsnn} \gls{pcc} by 6.6\%.

Table \ref{tab:rightwards} shows rightwards movements classification and processing time per method.

\begin{table}[H]\caption{Rightwards movements classification and processing time per method.}\label{tab:rightwards}
\resizebox{14.5cm}{!}{
\begin{tabular}{|c|c|c|c|c|}
\hline
Method & movement type & \gls{pcc} & \gls{pwc} & processing time \\ \hline\hline
MHSNNv2 & rightwards & 92.4\% & 7.6\% & 1775.007 ms \\ \hline
\gls{codd}-MOG & rightwards & 93.1\% & 6.9\% & 1.262 ms \\ \hline
\gls{codd}-KNN & rightwards & 92.5\% & 7.5\% & 1.716 ms \\ \hline
\gls{codd}-MOG2 & rightwards & 88.8\% & 11.2\% & 1.545 ms \\ \hline
MHSNNv1 & rightwards & 88.2\% & 11.8\% & 1803.407 ms \\ \hline
\gls{codd}-CNT & rightwards & 61.1\% & 38.9\% & 1.451 ms \\ \hline
\gls{codd}-GSOC & rightwards & 47.9\% & 52.1\% & 2.605 ms \\ \hline
\gls{codd}-LSBP & rightwards & 45.1\% & 54.9\% & 3.38 ms \\ \hline
\gls{codd}-GMG & rightwards & 0.0\% & 100.0\% & 1.604 ms \\ \hline
\end{tabular}}
\end{table}

When compared to the other \gls{codd} variants, the results demonstrate that the MHSNNv2 exhibits the highest \gls{pcc} and lowest \gls{pwc}. Again, the MHSNNv2 is outperformed by the \gls{codd}-MOG, which has a processing time average of 1.262 milliseconds. It is also feasible to conclude that the \gls{pca} whitening has helped to improve the \gls{mhsnn} \gls{pcc} by 4.2 percent.

Table \ref{tab:downwards} shows downwards movements classification and processing time per method.

\begin{table}[H]\caption{Downwards movements classification and processing time per method.}\label{tab:downwards}
\resizebox{14.5cm}{!}{
\begin{tabular}{|c|c|c|c|c|}
\hline
Method & movement type & \gls{pcc} & \gls{pwc} & processing time \\ \hline\hline
MHSNNv2 & downwards & 93.9\% & 6.1\% & 1898.088 ms \\ \hline
\gls{codd}-MOG & downwards& 93.1\% & 6.9\% & 1.213 ms \\ \hline
\gls{codd}-KNN & downwards& 92.3\% & 7.7\% & 1.491 ms \\ \hline
\gls{codd}-MOG2 & downwards& 88.8\% & 11.2\% & 1.228 ms \\ \hline
MHSNNv1 & downwards & 86.1\% & 13.9\% & 1928.457 ms \\ \hline
\gls{codd}-CNT & downwards& 61.1\% & 38.9\% & 1.412 ms \\ \hline
\gls{codd}-GSOC & downwards& 47.7\% & 52.3\% & 2.761 ms \\ \hline
\gls{codd}-LSBP & downwards& 45.3\% & 54.7\% & 3.488 ms \\ \hline
\gls{codd}-GMG & downwards& 0.0\% & 100.0\% & 1.642 ms \\ \hline
\end{tabular}}
\end{table}

The results show that the MHSNNv2 exhibits the greatest \gls{pcc} and lowest \gls{pwc} values when compared to the other \gls{codd} variants. Once more, the \gls{codd}-MOG outperforms the MHSNNv2 with an average processing time of 1.213 milliseconds. It is also possible to extrapolate that the \gls{pca} whitening contributed to an improvement of the \gls{mhsnn}'s \gls{pcc} by 7.8\%.

Table \ref{tab:upwards} shows upwards movements classification and processing time per method.

\begin{table}[H]\caption{Upwards movements classification and processing time per method.}\label{tab:upwards}
\resizebox{14.5cm}{!}{
\begin{tabular}{|c|c|c|c|c|}
\hline
Method & movement type & \gls{pcc} & \gls{pwc} & processing time \\ \hline\hline
MHSNNv2 & upwards & 93.1\% & 6.9\% & 1749.643 ms \\ \hline
\gls{codd}-MOG & upwards & 90.5\% & 9.5\% & 1.2 ms \\ \hline
\gls{codd}-KNN & upwards & 90.4\% & 9.6\% & 1.377 ms \\ \hline
\gls{codd}-MOG2 & upwards & 86.2\% & 13.8\% & 1.154 ms \\ \hline
MHSNNv1 & upwards & 87.6\% & 12.4\% & 1905.820 ms \\ \hline
\gls{codd}-CNT & upwards & 61.1\% & 38.9\% & 1.446 ms \\ \hline
\gls{codd}-GSOC & upwards & 46.1\% & 53.9\% & 2.749 ms \\ \hline
\gls{codd}-LSBP & upwards & 44.3\% & 55.7\% & 3.557 ms \\ \hline
\gls{codd}-GMG & upwards & 0.0\% & 100.0\% & 1.553 ms \\ \hline
\end{tabular}}
\end{table}

The results indicate that, when compared to the other \gls{codd} variations, the MHSNNv2 exhibits the highest \gls{pcc} and lowest \gls{pwc} values.
The \gls{codd}-MOG performs better than the MHSNNv2 once more, with an average processing time of 1.2  milliseconds.
It is also reasonable to deduce that the improvement of the \gls{mhsnn}'s \gls{pcc} by 5.5 percent was caused by the \gls{pca} whitening. 

\section{Discussion} \label{Ch3.5:discussion}

The \gls{mhsnn} architecture was designed to detect horizontal and vertical movements. The \gls{mhsnn} detected leftwards, rightwards, downwards, and upwards movements in 93.6\%, 92.4\%, 93.9\% and 93.1\%, respectively  when tested against the custom semisynthetic dataset. The \glspl{pwc} were consequence of \glspl{fp} at the beginning and end of the image sequences because the sequences were collated together on a single \textit{TimedArray}. The issue occurred because the first frame of another sequence was followed by the last frame of a given sequence. The \gls{mhsnn} was one of the first \gls{snn} capable of detecting horizontal and vertical movements when tested on semisynthetic datasets at the time of its publication back in 2018 \cite{Machado_2018}.\\

A \gls{codd} algorithm that combined the seven \gls{bs} available in the \gls{opencv} library was implemented to benchmark against the \gls{mhsnn}. Although the \gls{mhsnn} was ranked first in terms of the \gls{pcc} and exhibited the lowest \gls{pwc} detecting the motion direction when tested against the semisynthetic dataset, follow-up tests on natural datasets demonstrated that the \gls{mhsnn} would have to be scaled up, which would increase the latency even more. Furthermore, the Brian 2 simulator has proven not to be suitable for real-time applications because i) all the dataset must be loaded into the memory before starting the simulation and ii) Brian 2 is not optimised for running large \glspl{snn} (i.e. above 17000 neurons and 173700 synapses). The speed limitations are very obvious when looking at the processing times of the MHSNNv2 in Tables \ref{tab:leftwards}, \ref{tab:rightwards}, \ref{tab:downwards}, and \ref{tab:upwards}. \\

The lessons learnt from the \gls{mhsnn} architecture implementation for targeting real-time applications using natural datasets were: 1) a better \gls{snn} simulator or a customised \gls{snn} simulator would be required to decrease substantially the latency, 2) robust object motion detection requires the combination of simpler theories \gls{bs} algorithms with \glspl{snn} and 3) the custom \gls{snn} would most likely have to accommodate more than 100,000 neurons to provide the required accuracy. Finally, the \gls{mhsnn} architecture was fundamental for designing the \gls{hsmd} architecture, which will be discussed in the next chapter~\ref{Ch4:object_motion_detection_cells}.

%% file: Chapter4/chapter4.tex
%%%%%%%%%%%%%%%%%%%%%%%%%%%%%%%%%%%%%%%%%%%%%%%%%%%%%%%%%%%%%%%%%%%%%%%%%%%%%%%%
%2345678901234567890123456789012345678901234567890123456789012345678901234567890
% 1 2 3 4 5 6 7 8
% THESIS CHAPTER

\chapter{HSMD: Hybrid Spiking Motion Detection}
\label{Ch4:object_motion_detection_cells}
The \gls{hsmd} algorithm proposed in this chapter was designed to detect object motion in real-time and therefore overcome \acrlong{mhsnn} speed limitations. The speed limitation was dealt by replacing the Brian 2 simulator with an efficient and parallel C++ implementation and through reducing both the number of layers and synapses. Furthermore, the \gls{hsmd} combines a \gls{bs} algorithm with a customised \gls{snn} for detecting object motion as opposed to detecting the direction of motion like in the \gls{mhsnn}. The \gls{hsmd} enhances the \gls{gsoc} \gls{bs} algorithm with a customised 3-layer \gls{snn} that outputs spiking responses akin to the \gls{oms-gc}. The algorithm was compared against existing \gls{bs} approaches available on the \gls{opencv} library, specifically on the \gls{cdnet2012} and the \gls{cdnet2014} benchmark datasets. The results show that the \gls{hsmd} was ranked overall first among the competing approaches and has performed better than all the other algorithms in four of the categories across all the eight test metrics. Furthermore, the \gls{hsmd} proposed in this chapter is the first to use an \gls{snn} to enhance an existing state-of-the-art \gls{gsoc} \gls{bs} algorithm, and the results demonstrate that the \gls{snn} provides near real-time performance in realistic applications.

\section{Introduction} \label{Ch4.1:introduction}
In computer vision, object motion detection is traditionally performed using \gls{bs} methods, where the foreground (pixels or group of pixels whose light intensity values have suffered an abrupt variation) are compared with the previous image or background model \cite{piccardi2004,Maddalena2018,Chapel2020,Garcia-Garcia2020}. \gls{bs} are algorithms for extracting the background from the foreground by modelling the background through the comparison of the current frame with previous frames \cite{Holmes2018, STALIN2014, Vacavant2013, Sharma2018, Rashid2016, Seo2016}. \gls{bs} methods can be implemented using 1) mathematical-based, 2) machine learning, 3) signal processing, and 4) \glspl{dnn} approaches \cite{Chapel2020,Garcia-Garcia2020}. Mathematical-based approaches are the simplest way to model backgrounds using temporal average, temporal median, and histograms, which can be improved using refined models (such as a mixture of Gaussians, kernel density estimation, etc.) and require low computational resources \cite{Chapel2020}. Machine learning models are more robust for performing \gls{bs}, but they must be trained on the target visual features and require significant computational resources \cite{Garcia-Garcia2020}. Signal processing models are used to model the background using the temporal history of pixels as 1D signals and usually require moderate computational resources \cite{Chapel2020}. \gls{dnn} models are by far the most accurate, but they are also the most computationally intensive and therefore not suitable for real-time applications. Although less robust, the classical mathematical \gls{bs} models are better suited for real-time applications. As real-time processing is a key objective of this work, we focus only on mathematical models in this chapter. \\
\gls{bs} methods can be classified as 1) Mathematical, 2) Machine Learning and 3) Signal processing \cite{Chapel2020,Garcia-Garcia2020}. Mathematical theories are the simplest way to model backgrounds using temporal average, temporal median and histograms, which can be improved using refined models (such as a mixture of Gaussians, kernel density estimation, etc.) and require low computational resources \cite{Chapel2020}. Machine learning models are more robust for performing \gls{bs}, but they must be trained on the target visual features and require significant computational resources \cite{Garcia-Garcia2020}. Signal processing models used to model the background using the temporal history of pixels as 1D signals and usually require moderate computational resources \cite{Chapel2020}. Although less robust, the classical mathematical \gls{bs} models are better suited for real-time to near real-time applications. As real-time processing is a key objective of this work, we focus only on mathematical models in this chapter. \\

The \gls{opencv} library \cite{OpenCV2020} is one of the most robust and reliable computer vision libraries, maintained by a wide Open Source community (including high profile companies such as Intel, Microsoft and Google) and is the reference library for computer or robot vision researchers \cite{OpenCV2021}. The \gls{opencv}'s \gls{bs} algorithms (i.e. \gls{mog}, \gls{mog2}, CNT, \gls{knn}, GMG, \gls{lsbp} and \gls{gsoc}) are highly efficient \gls{bs} algorithms that were designed for modelling the dynamic background changes (i.e. about two hundred frames are required to train the background model) and classifying all the background outliers as foreground. The \gls{gsoc} algorithm was selected to perform the first stage of \gls{bs} over the other \gls{bs} available on the \gls{opencv} library because it has demonstrated better accuracy on the \gls{cdnet2012} and \gls{cdnet2014} datasets \cite{Samsonov2017} when compared to other algorithms available on the \gls{opencv} library. \\

The \gls{hsmd} model reported in this chapter was inspired by the object motion functionality exhibited by vertebrate retinas, in which \gls{oms-gc} determine the difference between a local patch's motion trajectory and the background \cite{Gollisch2010}. In fact, the \gls{hsmd} is an improved version of \gls{gsoc} \gls{bs} algorithm \cite{Samsonov2017,samsonovcode2017} is enhanced by a 3-layer \gls{snn}, forming a hybrid architecture. \\

The main contributions of the work reported in this chapter are i) an object motion detection model inspired by the \gls{oms-gc} designed to work with \gls{cots} cameras, ii) enhancement of the dynamic \gls{bs} (mathematical model) using the 3-layered \gls{snn} and iii) optimisation of the proposed method for processing live capture feeds in near real-time. The algorithm was tested on the \gls{cdnet2012} \cite{Goyette2012} and \gls{cdnet2014} benchmark datasets \cite{Wang2014} and compared with the \gls{opencv}'s \gls{bs} algorithms (i.e. \gls{mog}, \gls{mog2}, CNT, \gls{knn}, GMG, \gls{lsbp} and \gls{gsoc}). The \gls{hsmd} can detect motion using commercial-off-the-shelf camera feeds and/or video clips using \gls{snn}, as opposed to cameras exploiting dedicated custom architectures.\\

This chapter is structured as follows: the \gls{hsmd} is described in section \ref{Ch4.2:architecture}; the training details, use-case scenarios and \gls{hsmd} parameterisation are described in section \ref{Ch4.3:methodology}; the results are reported and analysed in section \ref{Ch4.4:results}; and the discussion and future work are discussed in section \ref{Ch4.5:discussion}.

\section{HSMD architecture} \label{Ch4.2:architecture}
The \gls{hsmd} is a combined \gls{bs}/\gls{snn} network to create a hybrid model for detecting motion, emulating the elementary functionalities of the \gls{oms-gc} as described in \cite{Gollisch2010}. \newline 

The architecture of the \gls{hsmd} is shown in Figure~\ref{fig:proposed_architecture}. There are five layers to the overall architecture. Layer 1 performs the \gls{bs} using the \gls{gsoc} algorithm. The resulting \gls{bs} frames are fed into Layer 2 of the \gls{snn}, where the pixel intensity values are converted into currents that are proportional to the light intensity (see \ref{layer_2}). The \gls{bs}-converted currents are fed to the Layer 2 neurons via a 1:1 synaptic connectivity. Layer 2 neurons are synaptically connected to Layer 3 neurons, which perform the first stage of motion analysis; Layer 3 neurons connect to Layer 4 neurons via 1:1 synaptic connectivity. Layer 4 neurons perform precise motion detection. A median filter is used to filter random spikes related to local random illumination variations.
 
\begin{figure*}[hbt!]
\centering
\includegraphics[width=1.0\textwidth]{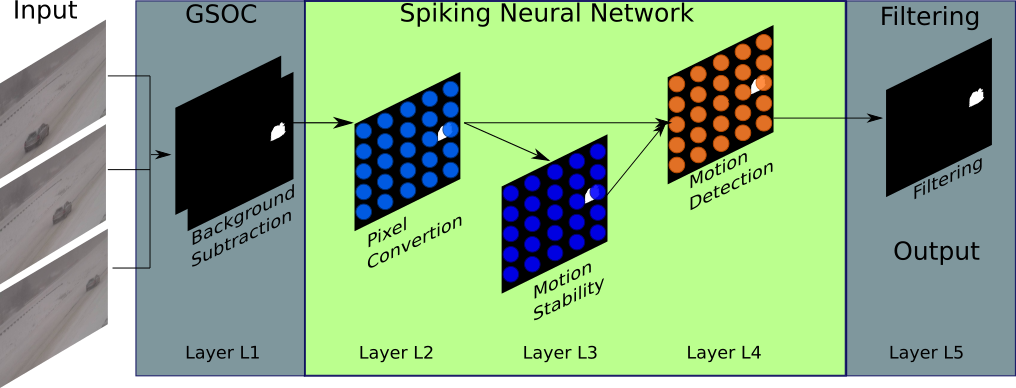}
\caption{\gls{hsmd} with (i) $n\times m$ image input followed by the \gls{bs} using the \gls{gsoc} algorithm, three spiking neuronal layers and filtering. Layer 1: \gls{bs}, Layer 2: pixel intensity to spike events encoding, Layer 3: Motion stability, Layer 4: motion detection and Layer 5: filtering.
\label{fig:proposed_architecture_2}} 
\end{figure*}

The \gls{lif} was the spiking neuron model used in this work because of its simplicity, computational efficiency and suitability for processing images in near real-time. The \gls{lif} spiking neuron model exhibits similar, but less complex, dynamics compared to real biological neurons (see Figure~\ref{fig:LIF}) \cite{Gerstner2002}. The \gls{lif} neuron's dynamics are described by equation \ref{eq:2.1}.

\subsection{Input Layer: background subtraction and reduction}\label{input_layer}

Each $n \times m$ image frame (i.e. camera, video sequence or image sequences) is converted into greyscale.\\

The \gls{gsoc} \cite{GSOC2020} delivers a dynamic and adaptive \gls{bs} using colour descriptors and various stabilisation heuristics \cite{Samsonov2017,samsonovcode2017} while processing the frames pixel-wise and leveraging the parallelism inside \gls{opencv} \cite{Samsonov2017}. 

\subsection{Layer 2: Pixel intensities values to currents encoding}\label{layer_2}
Pixel intensity values are converted into proportional currents and fed into the spiking neurons in Layer 2 via a 1:1 connectivity. The Layer 1 neurons were trained to trigger spike events proportional to the pixel intensity values, as described by equation~\ref{eq:pixel_current}. 
\begin{equation}
 i_c(x,y)=I(x,y). c
 \label{eq:pixel_current}
\end{equation}

\noindent where $i_c(x,y)$ is the corresponding current for the image light intensity value $I(x,y)$ at coordinates $x$ and $y$, and $c$ is a conversion constant obtained experimentally (in our case, $c$=17.5).

\subsection{Layer 3: Motion stability} \label{layer_3}
Layer 3 is used to perform motion stabilisation through the creation of local buffers by delaying the propagation of spike events. A delay is created when a given neuron of layer 2 connects to a neuron in layer 3, before being passed to Layer 4, instead of intra-layer connectivity between direct Layer 2 and Layer 4. Spike events passing through Layer 3 are buffered by neurons in Layer 3 for one simulation time-step ($\delta t$, in this work, $\delta t=10$ ms) and presented to the neurons in Layer 4. N[n] in the following simulation time-step. 

\begin{figure}[H]
\centering
\includegraphics[width=0.45\textwidth]{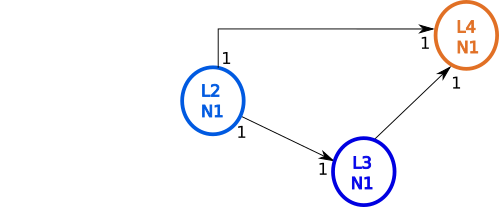}
\caption{\gls{hsmd} connectivity. In this example, it can be seen that the neuron 1 (N1) of each layer connects to the N1 of the subsequent layer.
\label{fig:connectivity}} 
\end{figure}

The neurons in Layer 2 are connected to the Layer 3 neurons via a 1:1 connectivity. Finally, the Layer 3 neurons connect to the Layer 4 neurons via a 1:1 connectivity, as shown in Figure~\ref{fig:connectivity}. All synaptic weights from Layer 2 to Layers 3 and 4 have a value of 1370 (obtained experimentally).

\subsection{Layer 4: Motion detection} \label{layer_4}
The Layer 4 neurons receive synaptic connections from the neurons in Layer 2 and Layer 3 via excitatory synapses and exploit these spiking events to detect motion. Spike events generated by Layer 4 neurons resulted from dynamic changes between sequential image frames. Signals received directly from Layer 2 neurons enable detection of changes between the current image \textit{frame n} and the previous image \textit{frame n-1}. In contrast, those routed via Layer 3 neurons compare the image \textit{frame n-1} with the image \textit{frame n-2}. Layer 4 spike events are mapped into the corresponding area in the original image captured from the camera. The synaptic weights obtained experimentally are 1370 for all the synapses. The Layer 2 to Layer 4 weights were tuned to forward all the spike events generated in Layer 2. The Layer 3 to Layer 4 synaptic weights were tuned to produce spike events from the Layer 4 neurons for each group of two sequential spike events. The main goal is to give high importance (larger weight) to new spike events ($frame [n] - frame [n-1]$) and lower importance to older spike events ($frame [n-1] - frame [n-2]$).

\subsection{Layer 5: Filtering} \label{layer_5}
The Layer 4 neurons' spike events matrix is mapped into a motion matrix $M_d$ of the same size as the captured image (i.e. $n\times m$). The events in the $M_d$ matrix are filtered using an averaging filter described by equations \ref{eq:filter} and ~\ref{eq:conv}:

\begin{equation}
H(u,v)=\frac{1}{u.v}\begin{pmatrix}
\begin{bmatrix}
w_{0,\,0} & ... & w_{0,\,u} \\ 
 ... & ... & ... \\ 
w_{v,\,0} & ... & w_{v,\,u}
\end{bmatrix}
\end{pmatrix}
\label{eq:filter}
\end{equation}

\begin{equation}
Y_d(x,y)=M_d(x,y)*H(u,v)
\label{eq:conv}
\end{equation}

\noindent where $Y_d(x,y)$ is the filtered motion detection matrix, $H(u,v)$ is the averaging Gaussian filter \cite{Deng1993}, $u$ and $v$ are the convolution window length and height respectively, $*$ is the convolution operator, $w$ is the filter window.

\section{Implementation details} \label{Ch4.3:methodology}

The \gls{hsmd} was implemented in C++ using the C++ \gls{stl} 17 (implementation of data structures) \cite{StandardC++2017}, Boost 1.71 (file management) \cite{Boost2020} and \gls{opencv} 4.5.0 (image processing) \cite{OpenCV2020}. The decision not to use any existing \gls{snn} simulator (such as Brian 2 \cite{Brian2021}, and NEST \cite{NEST2019}) was made to ensure that the \gls{snn} could have the lowest latency possible. The \gls{stl} C++17 library offers a collection of C++ algorithms that have been optimised to deliver a higher degree of parallelisation when running on multicore \gls{cpu} systems; The Boost library offers a wide range of reusable algorithms, including file management, time monitoring, and exception handling algorithms; and the \gls{opencv} library offers a collection of algorithms for computer vision applications, including image manipulation and filtering, and \gls{bs} algorithms.

\subsection{HSMD setup} \label{Ch4.3.1:setup}
The \gls{hsmd} initial setup includes the following steps:

\textbf{Step 1 - Select between live capture, video analysis or image sequences:} The user can opt to run the algorithm directly on images being captured by the camera or provide the path of a video or set of image sequences for motion analysis.

\textbf{Step 2 - Create the Layer 2 to Layer 4 neural network:} Read the first image and compute the size of the image. The number of neurons is computed automatically from the dimensions of the first image in a sequence of images.

\textbf{Step 3 - Set the neuronal parameters:} The \gls{lif} parameters recommended in the references \cite{Jolivet2004, Brette2005} were used to configure the \gls{snn}. Therefore, the simulation was configured with a time step of $ \delta t$=10 ms and the default neuron parameters as follows: initial $V_m$=-55.0 mV, $E_L$ = -55.0 mV, $C_m$ = 10.0 pF, $R_m$=1.0 MOhm, $V_{reset}$=-70.0 mV, $V_{min}$=-70.0 ms, $V_{th}$=-70.0 mV, $\tau_m$=10.0 ms, $t_{ref}$=2 ms, $w_{syn}$ = 1555.0 (neurons L3 and L4) and $w_{p2i}$=8.0 (L2 neurons only). 

\textbf{Step 4 - Start the image acquisition :} Images are loaded from folders with sequences of images while the \gls{hsmd} algorithm is being executed. The pseudo-code of the main algorithm is described in Algorithm~\ref{alg:main}.

\begin{algorithm}
 \caption{\gls{hsmd} main algorithm pseudo-code}
 \label{alg:main}
 \begin{algorithmic}[1]
 \STATE $newImage = get\_new\_image\;$
 \STATE $newImageGrey=colour2grey(newImage)\;$
 \STATE $set\_number\_neurons\_from\_newImageGrey\_shape\;$
 \STATE $build\_neuronal\_network\;$
 \STATE $load\_pretrained\_weights\;$
 \WHILE{frames available}
 \STATE $reset\_spike\_events\;$
 \STATE $newImage = capture\_image\_camera\;$
 \STATE $newImageGrey=colour2grey(newImage)\;$
 \STATE $newImageReduced=newImageGrey\;$
 \STATE $dynSubImage=newImageReduced-previousImage\;$
 \STATE $ previousImage=newImageReduced\;$
 \FOR{I\ in\ dynSubImage}
 \IF{dynSubImage[I]$<$ Threshold}
 \STATE $dynSubImage[I]=0.0\;$
 \ENDIF
 \STATE $currents=convPixel2Current(dynSubImage)\;$
 \FOR{i:=0 \TO timestep}
 \STATE $apply\_currents\_to\_neurons\_L2\;$
 \STATE $compute\_L2\_neuron\_spikes;$
 \STATE $convert\_L2\_neuron\_spikes\_to\_currents;$
 \STATE $compute\_L3\_neuron\_spikes;$
 \STATE $convert\_L3\_neuron\_spikes\_to\_currents;$
 \STATE $compute\_L4\_neuron\_spikes;$
 \ENDFOR
 \STATE $spikes=get\_sumSpikeEventsPerL4Neuron()\;$
 \STATE $masked\_spikes=applyAveragingFilter(spikes)\;$
 \STATE $spikes=normalise(spikes)\;$
 \STATE $display(newImage)\;$
 \STATE $display(spikes)\;$
 \ENDFOR
 \ENDWHILE
 \STATE $Display\_spike\_rates;$
\end{algorithmic}
\end{algorithm}

\subsection{Datasets and metrics} \label{Ch4.3.2:datasets_metrics}

\subsubsection{Datasets} \label{Ch4.3.2.1:datasets}

The \gls{cdnet2012} \cite{Goyette2012} (cited more than 379 times and \gls{cdnet2014} \cite{Wang2014} (cited more than 300 times) benchmark datasets were designed for benchmarking \gls{bs} algorithms. While the \gls{hsmd} algorithm has been designed as an object detection algorithm and not a \gls{bs} algorithm, nevertheless these two datasets provide challenging scenarios for robust comparable assessment of the proposed algorithm and network. The \gls{hsmd} was compared against state-of-the-art \gls{bs} algorithms available on the \gls{opencv} library: \gls{mog}~ \cite{Stauffer1999}, \gls{mog2}~\cite{Zivkovic2004}, \gls{knn}~\cite{Zivkovic2006}, GMG~\cite{Godbehere2014}, \gls{lsbp}~\cite{Guo2016}, CNT~\cite{CNT-2016} and \gls{gsoc}~\cite{samsonovcode2017}. The \gls{opencv} \gls{bs} algorithms were used because they are highly optimised, reliable, and publicly available to anyone who wants to test or compare their algorithms.

Each of the benchmark videos in the \gls{cdnet2012} \cite{Goyette2012} and \gls{cdnet2014} \cite{Wang2014} fall into one or more of the challenge categories listed below:\\
\vspace{.1cm}
\textbf{\gls{cdnet2012} and \gls{cdnet2014}}
\begin{itemize}
\item \textbf{Baseline}: reference videos, which are relatively simple to classify; some videos contain very simple movements from the next four categories.
\item \textbf{Dynamic Background}: videos that have both foreground and background motion (e.g. water movement and shaking trees).
\item \textbf{Camera Jitter}: videos captured with cameras installed on unstable structures.
\item \textbf{Shadow}: videos containing narrow shadows from solid structures or moving objects.
\item \textbf{Intermittent Object Motion} videos include objects that are static for most of the time and suddenly start moving.
\item \textbf{Thermal}: videos that exhibit thermal artefacts (i.e. bright spots and thermal reflections on windows and floors).
\end{itemize}

\textbf{\gls{cdnet2014} only}
\begin{itemize}
\item \textbf{Challenging Weather}: outdoor videos recorded during winter storm conditions with extremely low visibility.
\item \textbf{Low Frame-Rate}: videos captured at frame rates ranging from 0.17 to 1 \gls{fps};
\item \textbf{Night}: includes traffic videos with low visibility and strong headlights. 
\item \textbf{\gls{ptz}}: videos recorded with cameras that were subjected to \gls{ptz} movements. 
\item \textbf{Air Turbulence}: videos filmed from distances of 5 to 15 km exhibiting air turbulence and frame distortion.
\end{itemize}

Table~\ref{tab:cat_dataset} lists the categories per each dataset.

\begin{table}[H] \caption{Categories available per each dataset}\label{tab:cat_dataset}
\resizebox{14.5cm}{!}{
\begin{tabular}{|c|c|c|c|c|c|c|c|c|c|c|}
\hline
\begin{tabular}[c]{@{}c@{}}Category /\\ dataset\end{tabular} & \begin{tabular}[c]{@{}c@{}}bad \\ weather\end{tabular} & baseline & \begin{tabular}[c]{@{}c@{}}camera\\ jitter\end{tabular} & \begin{tabular}[c]{@{}c@{}}intermittent\\ object motion\end{tabular} & \begin{tabular}[c]{@{}c@{}}low frame\\ rate\end{tabular} & \begin{tabular}[c]{@{}c@{}}night \\ videos\end{tabular} & \begin{tabular}[c]{@{}c@{}}Pan, Tilt\\ Zoom\end{tabular} & Shadow & thermal & turbulence \\ \hline
\gls{cdnet2012} & & X & X & X &  &  &  & X & X & \\ \hline
\gls{cdnet2014} & X & X & X & X & X & X & X & X & X & X\\ \hline
\end{tabular}}
\end{table}
The \gls{bs} algorithms were configured using the default \gls{opencv} settings \cite{OpenCV2020} and compared against the \gls{hsmd} algorithm. The ground-truth provided by the datasets is composed of the following labels \cite{Goyette2012,Wang2014}:

\begin{itemize}
\item \textbf{Static} - greyscale value 0;
\item \textbf{Shadow} - greyscale value 50;
\item \textbf{non-\gls{roi}} - greyscale value 85;
\item \textbf{Unknown} - greyscale value 170;
\item \textbf{Moving} - greyscale value 255;
\end{itemize}

\begin{figure}[h]
\centering
\includegraphics[width=0.48\textwidth]{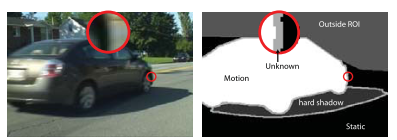}
\caption{Raw image frame (left) and its respective ground-truth (right). The ground-truth images show the annotations using the datasets labels. Adapted from \protect\cite{Goyette2012}}. \label{fig:gt}
\end{figure}

The \textit{static} and \textit{moving} classes contain pixels that belong to the background and foreground, respectively; the \textit{shadows}, one of the most challenging artefacts, should be classified as part of the background. The \textit{unknown} region should not be considered either background or foreground because it contains pixels that cannot be accurately classified as background or foreground. The non-\gls{roi} pixels serve to exclude frames from being classified because some \gls{bs} algorithms require several pixels for the model to stabilise (i.e. create the background model) and for preventing corruption by non-related activities to the considered category \cite{Goyette2012,Wang2014}. Figure~\ref{fig:gt} shows the 5 class regions.

\subsubsection{Metrics} \label{Ch4.3.2.2:metrics}
The average performance obtained for each category using each \gls{bs} method and the \gls{hsmd} algorithms is characterised via eight metrics, as shown below. The four fundamental qualitative metrics are: \gls{tp}, \gls{tn}, \gls{fp} and \gls{fn} \cite{Goyette2012,Wang2014}.

\begin{enumerate}
\item \gls{re}: $\gls{re}=\frac{\gls{tp}}{\gls{tp}+\gls{fn}}$

$\gls{re}$:  ranked by \textbf{descending order};

\item \gls{sp}: $\gls{sp}=\frac{\gls{tn}}{\gls{tn}+\gls{fp}}$;

$\gls{sp}$  ranked by \textbf{descending order};

\item \gls{fpr}: $\gls{fpr}=\frac{\gls{fp}}{\gls{fp}+\gls{tn}}$;

$\gls{fpr}$  ranked by \textbf{ascending order};

\item \gls{fnr}: $\gls{fnr}=\frac{\gls{fn}}{\gls{fn}+\gls{tp}}$;

$\gls{fnr}$  ranked by \textbf{ascending order};

\item \gls{wcr}: $\gls{wcr}=\frac{\gls{fn}+\gls{fp}}{\gls{tp}+\gls{fn}+\gls{fp}+\gls{tn}}$;

$\gls{wcr}$  ranked by \textbf{ascending order};

\item \gls{ccr}: $\gls{ccr}=\frac{\gls{tp}+\gls{tn}}{\gls{tp}+\gls{fn}+\gls{fp}+\gls{tn}}$;

$\gls{ccr}$  ranked by \textbf{descending order};

\item \gls{pr}: $\gls{pr}=\frac{\gls{tp}}{\gls{tp}+\gls{fp}}$;

$\gls{pr}$  ranked by \textbf{descending order};

\item \gls{f1}: $\gls{f1}=2\times \frac{\gls{pr}.\gls{re}}{\gls{pr}+\gls{re}}$

$\gls{f1}$ ranked by \textbf{descending order};

\end{enumerate}

\gls{r}: $\gls{r}=\frac{\overline{\gls{re}}+\overline{\gls{sp}}+\overline{\gls{fpr}}+\overline{\gls{fnr}}+\overline{\gls{wcr}}+\overline{\gls{ccr}}+\overline{\gls{f1}}}{nMet}$;

$\gls{r}$ ranked by \textbf{ascending order};

\gls{arc}:\\
$\overline{\gls{arc}}=\frac{\gls{re}+\gls{sp}+\gls{fpr}+\gls{fnr}+\gls{wcr}+\gls{ccr}+\gls{f1}}{nMet}$;

$\overline{\gls{arc}}$  ranked by \textbf{ascending order};

where $nMet$ is the number of metrics (8 in this case). 

\section{Results} \label{Ch4.4:results}
The \gls{hsmd} was tested on both datasets under the same conditions to ensure an accurate and rigorous comparison. The results are presented both as overall results and per category to better understand the specific performances obtained per method. The overall results for each method are presented in section \ref{Ch4.1:overall results} and the results per method and category are presented in section \ref{Ch4.2:results_category}.

The $\uparrow$ indicates that the highest score is the best result, and the $\downarrow$ that the lowest result is the best result in the tables (\ref{tab:results_overall_12} to \ref{tab:results_category}). The best results are highlighted using light grey for all the methods except the \gls{hsmd} results, which are highlighted in dark grey. $\gls{re}$ stands for recall, $\gls{sp}$ specificity, \gls{fpr} False Positive Rate, \gls{fnr} False Negative Rate, \gls{wcr} Wrong Classifications Rate, \gls{ccr} Correct Classifications Rate, \gls{pr} Precision, \gls{f1} F-score, \gls{r} Average Ranking and $\overline{\gls{arc}}$ Average Ranking across all categories.

\begin{figure*}[htb!]
\centering
\includegraphics[width=1\textwidth]{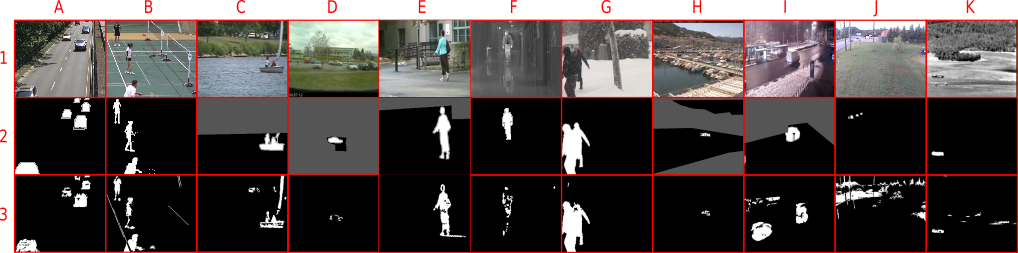}
\caption{Results obtained for each of the eleven of the five categories (columns A to F) are common to both \gls{cdnet2012} and \gls{cdnet2014} datasets, while the remaining six categories (columns G to K) are only available on \gls{cdnet2014} dataset. Column A: baseline; B: camera jitter; C: dynamic background; D: dynamic object motion; E: shadow, F: thermal, G: bad weather, H: low frame rate; I: night videos, J: PTZ and K: turbulence. Row 1: RGB image; 2: ground-truth; and 3: \gls{hsmd} binarised. The raw images, shown in the first row, demonstrate the scenarios that can be found in both datasets. The corresponding ground truth images, presented in the second row, show the 5 labels, namely, i) static [greyscale value 0], ii) shadow [greyscale value 50], iii) non-\gls{roi} [greyscale value 85], iv) unknown [greyscale value 170] and v) moving [greyscale value 255]. The corresponding binarised images generated by the \gls{hsmd} are shown in the third row.} \label{fig:CDnet2012_output}
\end{figure*}

The results for each of the eleven categories shared by both \gls{cdnet2012} and \gls{cdnet2014} are shown in Figure~\ref{fig:CDnet2012_output}.

\subsection{Overall results} \label{Ch4.1:overall results}
Tables~\ref{tab:results_overall_12} and \ref{tab:results_overall_14} present the overall results obtained per method and per metric, ranked by $\overline{\gls{arc}}$ (average ranking across all categories, first column) in ascendant order.

\begin{table*}[htb!]\caption{\gls{cdnet2012} overall results. Results ordered in descendent order by $\overline{\gls{arc}}$} \label{tab:results_overall_12}
\resizebox{14.5cm}{!}{
\begin{tabular}{|l|>{\columncolor[gray]{0.8}}l|l|l|l|l|l|l|l|l|l|}
\hline
\rowcolor[gray]{0.75}
Method & $\overline{\gls{arc}}$ $\downarrow$ & \gls{re} $\uparrow$ & \gls{sp} $\uparrow$ & \gls{fpr} $\downarrow$ & \gls{fnr} $\downarrow$ & \gls{wcr} $\downarrow$ & \gls{ccr} $\uparrow$ & \gls{f1} $\uparrow$ & \gls{pr} $\uparrow$ \\ \hline\hline
\cellcolor{gray}\textcolor{white}{\protect\textbf{\gls{hsmd}}}&\cellcolor{gray}\textcolor{white}{\protect\textbf{2.8}}&0.52&0.994&0.006&0.23&\cellcolor{gray}\textcolor{white}{\protect\textbf{0.024}}&\cellcolor{gray}\textcolor{white}{\protect\textbf{0.976}}&0.77&0.62\\\hline
\gls{gsoc}&3.5&0.54&0.993&0.007&0.25&\cellcolor{gray!20}\textit{0.024}&\cellcolor{gray!20}\textit{0.976}&0.75&\cellcolor{gray!20}\textit{0.63}\\\hline
\gls{mog2}&3.8&0.37&0.995&0.004&0.24&0.026&0.974&0.76&0.50\\\hline
GMG&3.9&0.20&\cellcolor{gray!20}\textit{0.998}&\cellcolor{gray!20}\textit{0.002}&\cellcolor{gray!20}\textit{0.21}&0.033&0.967&\cellcolor{gray!20}\textit{0.79}&0.32\\\hline
\gls{knn}&4.3&0.39&0.995&0.005&0.26&0.025&0.975&0.74&0.51\\\hline
\gls{mog}&4.5&0.32&0.996&0.004&0.26&0.030&0.970&0.74&0.44\\\hline
CNT&6.1&\cellcolor{gray!20}\textit{0.73}&0.927&0.073&0.71&0.081&0.919&0.29&0.41\\\hline
\gls{lsbp}&7.3&0.57&0.90&0.096&0.80&0.109&0.891&0.20&0.29\\\hline
\end{tabular}}
\\
$\uparrow$: the highest score is the best.\\
$\downarrow$: the lowest result is the best. \\
Best \gls{hsmd} results are highlighted using dark grey, while best results per category are highlighted with light grey for other methods.
$\gls{re}$ stands for Recall, $\gls{sp}$ Specificity, \gls{fpr} False Positive Rate, \gls{fnr} False Negative Rate, \gls{wcr} Wrong Classifications Rate, \gls{ccr} Correct Classifications Rate, \gls{pr} Precision, \gls{f1} F-score and $\overline{\gls{arc}}$ Average Ranking across all Categories.
\end{table*}

The \gls{hsmd} algorithm ranks in first place across all eight methods with which it is compared when tested against the \gls{cdnet2012} dataset (see the $\overline{\gls{arc}}$ results in Table~\ref{tab:results_overall_12}). Although the \gls{hsmd} performed very well in five of the eight metrics, it is essential to highlight the results from the \gls{wcr}, \gls{ccr}, and \gls{f1} metrics. The results show that the \gls{hsmd} is sensitive to object motion due to the highest correct counts and lowest wrong count rates that contributed to get the highest F-score and the second-best precision. Furthermore, the \gls{gsoc} algorithm ranks in second place immediately after the \gls{hsmd}.

\begin{table*}[htb!]\caption{\gls{cdnet2014} overall results. Results ordered in descendent order by $\overline{\gls{arc}}$} \label{tab:results_overall_14}
\resizebox{14.5cm}{!}{\begin{tabular}{|l|>{\columncolor[gray]{0.8}}l|l|l|l|l|l|l|l|l|l|}
\hline
\rowcolor[gray]{0.75}
Method & $\overline{\gls{arc}}$ $\downarrow$ & \gls{re} $\uparrow$ & \gls{sp} $\uparrow$ & \gls{fpr} $\downarrow$ & \gls{fnr} $\downarrow$ & \gls{wcr} $\downarrow$ & \gls{ccr} $\uparrow$ & \gls{f1} $\uparrow$ & \gls{pr} $\uparrow$ \\ \hline\hline
\cellcolor{gray}\textcolor{white}{\protect\textbf{\gls{hsmd}}}&\cellcolor{gray}\textcolor{white}{\protect\textbf{2.9}}&0.55&0.993&0.007&0.35&0.018&0.982&0.65&\cellcolor{gray}\textcolor{white}{\protect\textbf{0.60}}\\\hline
\gls{gsoc}&3.0&0.40&0.995&0.005&0.38&\cellcolor{gray!20}\textit{0.017}&\cellcolor{gray!20}\textit{0.983}&0.62&0.48\\\hline
\gls{knn}&3.5&0.34&0.996&0.004&\cellcolor{gray!20}\textit{0.32}&0.019&0.981&\cellcolor{gray!20}\textit{0.68}&0.45\\\hline
GMG&4.3&0.24&\cellcolor{gray!20}\textit{0.997}&\cellcolor{gray!20}\textit{0.003}&0.36&0.022&0.978&0.64&0.35\\\hline
\gls{mog}&4.4&0.58&0.991&0.009&0.39&0.019&0.981&0.61&0.60\\\hline
\gls{mog2}&4.5&0.39&0.994&0.006&0.42&0.018&0.982&0.58&0.47\\\hline
\gls{lsbp}&6.5&0.58&0.945&0.055&0.79&0.064&0.936&0.21&0.31\\\hline
CNT&7.0&\cellcolor{gray!20}\textit{0.72}&0.930&0.070&0.80&0.075&0.925&0.20&0.32\\\hline
\end{tabular}}
$\uparrow$: the highest score is the best.\\
$\downarrow$: the lowest result is the best. \\
Best \gls{hsmd} results are highlighted using dark grey, while best results per category are highlighted with light grey for other methods.
$\gls{re}$ stands for Recall, $\gls{sp}$ Specificity, \gls{fpr} False Positive Rate, \gls{fnr} False Negative Rate, \gls{wcr} Wrong Classifications Rate, \gls{ccr} Correct Classifications Rate, \gls{pr} Precision, \gls{f1} F-score and $\overline{\gls{arc}}$ Average Ranking across all Categories.
\end{table*}

Table~\ref{tab:results_overall_14} shows that the \gls{hsmd} algorithm was ranked in first place in the average ranking $\overline{\gls{arc}}$ across all 11 categories when tested on the \gls{cdnet2014} dataset. The \gls{hsmd} performed very well in seven of the eight metrics, and exceptionally well in the precision metric.

\begin{table*}[htb!]\caption{\gls{hsmd} overall results. Results ordered in descendent order by $\overline{\gls{arc}}$} \label{tab:results_overall_1214}
\resizebox{14.5cm}{!}{
\begin{tabular}{|l|l|l|l|l|l|l|l|l|l|l|}
\hline
\rowcolor[gray]{0.75}
Dataset & $\overline{\gls{arc}}$ $\downarrow$ & \gls{re} $\uparrow$ & \gls{sp} $\uparrow$ & \gls{fpr} $\downarrow$ & \gls{fnr} $\downarrow$ & \gls{wcr} $\downarrow$ & \gls{ccr} $\uparrow$ & \gls{f1} $\uparrow$ & \gls{pr} $\uparrow$ \\ \hline\hline
\cellcolor{gray}\textcolor{white}{\protect\textbf{\gls{cdnet2012}}}&\cellcolor{gray}\textcolor{white}{\protect\textbf{2.8}}&0.52&\cellcolor{gray}\textcolor{white}{0.994}&\cellcolor{gray}\textcolor{white}{0.006}&\cellcolor{gray}\textcolor{white}{0.23}&0.024&0.976&\cellcolor{gray}\textcolor{white}{0.77}&\cellcolor{gray}\textcolor{white}{0.62}\\\hline
\gls{cdnet2014}&2.9&\cellcolor{gray}\textcolor{white}{0.55}&0.993&0.007&0.35&\cellcolor{gray}\textcolor{white}{0.018}&\cellcolor{gray}\textcolor{white}{0.982}&0.65&0.60\\\hline
\end{tabular}}
$\uparrow$: the highest score is the best.\\
$\downarrow$: the lowest result is the best. \\
Best results are highlighted in gray.\\
$\gls{re}$ stands for Recall, $\gls{sp}$ Specificity, \gls{fpr} False Positive Rate, \gls{fnr} False Negative Rate, \gls{wcr} Wrong Classifications Rate, \gls{ccr} Correct Classifications Rate, \gls{pr} Precision, \gls{f1} F-score and $\overline{\gls{arc}}$ Average Ranking across all Categories.
\end{table*}
Table \ref{tab:results_overall_1214} shows that there was a slight decrease in the \gls{hsmd} performance when tested on the eleven categories available in the \gls{cdnet2014} as compared to the original six in the \gls{cdnet2012} dataset. Furthermore, the extra 6 categories, which are only present in the \gls{cdnet2014} contributed to an increase from 2.8 in the \gls{cdnet2012} to 2.9 in the \gls{cdnet2014} of $\overline{\gls{arc}}$ across all categories. The increase of $\overline{\gls{arc}}$ is justifiable with the degradation of 5 metrics, namely, \gls{sp}, \gls{fpr} , \gls{fnr}, \gls{f1} and \gls{pr} as a consequence of the complexity introduced by the video sequences from the extra six categories. This is more evident when analysing the individual \acrfull{r} results per dataset as per listed in Table \ref{tab:results_category}, and Figures~\ref{fig:CDnet2012_results} and \ref{fig:CDnet2014_results}.

Finally, it was anticipated that none of the methods would have excellent performance across all metrics because of the complexity and size of the \gls{cdnet2012} and \gls{cdnet2014} datasets.

\subsection{Results obtained per category} \label{Ch4.2:results_category}

The \gls{r} for each of the methods per category is shown in Table~\ref{tab:results_category}.
\begin{landscape}
\begin{table*}
\begin{center}
\caption{Results per category. Results ordered in descendent order by \gls{r}} \label{tab:results_category}
\resizebox{21cm}{!}{
\begin{tabular}{|l|l|l|l|l|l|l|l|l|l|l|l|l|l|l|l|l|l|l|l|l|l|}
\hline
\rowcolor[gray]{0.75}
\multicolumn{2}{|c|}{IntObjMotion}&
\multicolumn{2}{|c|}{shadow} & 
\multicolumn{2}{|c|}{cameraJitter} &
\multicolumn{2}{|c|}{badWeather} &
\multicolumn{2}{|c|}{dynamicBackground} &
\multicolumn{2}{|c|}{nightVideos} &
\multicolumn{2}{|c|}{\gls{ptz}}&
\multicolumn{2}{|c|}{thermal} & 
\multicolumn{2}{|c|}{baseline} & 
\multicolumn{2}{|c|}{lowFramerate} & 
\multicolumn{2}{|c|}{turbulence} 
\\\hline
\rowcolor[gray]{0.75}
Method & \gls{r}$\downarrow$ & 
Method & \gls{r}$\downarrow$ & 
Method & \gls{r}$\downarrow$ &
Method & \gls{r}$\downarrow$ &
Method & \gls{r}$\downarrow$ &
Method & \gls{r}$\downarrow$ &
Method & \gls{r}$\downarrow$ & 
Method & \gls{r}$\downarrow$ & 
Method & \gls{r}$\downarrow$ & 
Method & \gls{r}$\downarrow$ & 
Method & \gls{r}$\downarrow$ 
\\\hline\hline
\cellcolor{gray}\textcolor{white}{\protect\textbf{\gls{hsmd}}}&
\cellcolor{gray}\textcolor{white}{\protect\textbf{3.875}}&
\cellcolor{gray!20}\textit{\gls{mog2}}&
\cellcolor{gray!20}\textit{2.375}&
\cellcolor{gray!20}\textit{\gls{lsbp}}&
\cellcolor{gray!20}\textit{1.875}&
\cellcolor{gray!20}\textit{CNT}&
\cellcolor{gray!20}\textit{2.125}&
\cellcolor{gray!20}\textit{\gls{lsbp}}&
\cellcolor{gray!20}\textit{1.75}&
\cellcolor{gray}\textcolor{white}{\protect\textbf{\gls{hsmd}}}&
\cellcolor{gray}\textcolor{white}{\protect\textbf{2.625}}&
\cellcolor{gray!20}\textit{CNT}&
\cellcolor{gray!20}\textit{1.75}&
\cellcolor{gray}\textcolor{white}{\protect\textbf{\gls{hsmd}}}&
\cellcolor{gray}\textcolor{white}{\protect\textbf{3.25}}&
\cellcolor{gray!20}\textit{\gls{mog2}}&
\cellcolor{gray!20}\textit{2.625}&
\cellcolor{gray!20}\textit{\gls{gsoc}}&
\cellcolor{gray!20}\textit{2.25}&
\cellcolor{gray}\textcolor{white}{\protect\textbf{\gls{hsmd}}}&
\cellcolor{gray}\textcolor{white}{\protect\textbf{2.875}} 
\\\hline

\gls{lsbp}&4.125&CNT&3.5&CNT&2.625&\cellcolor{gray}\textcolor{white}{\protect\textbf{\gls{hsmd}}}&\cellcolor{gray}\textcolor{white}{\protect\textbf{2.5}}&\gls{mog}&2.375&\gls{lsbp}&3.625& \gls{mog}&3.0&\gls{mog}&3.375&\gls{mog}&3.125&GMG&3.0&\gls{mog2}&2.875\\\hline

GMG&4.125&\cellcolor{gray}\textcolor{white}{\protect\textbf{\gls{hsmd}}}&\cellcolor{gray}\textcolor{white}{\protect\textbf{3.875}}&\gls{gsoc}&4.125&\gls{knn}&3.5&\gls{knn}&3.625&\gls{mog2}&4.0&\gls{lsbp}&4.25&\gls{gsoc}&3.625&\cellcolor{gray}\textcolor{white}{\protect\textbf{\gls{hsmd}}}&\cellcolor{gray}\textcolor{white}{\protect\textbf{3.625}}&CNT&3.375&\gls{lsbp}&3.875\\\hline

\gls{gsoc}&4.5&\gls{knn}&4.25&\cellcolor{gray}\textcolor{white}{\protect\textbf{\gls{hsmd}}}&\cellcolor{gray}\textcolor{white}{\protect\textbf{4.5}}&GMG&3.75&CNT&4.125&\gls{gsoc}&4.25&\gls{mog2}&4.875&\gls{knn}&3.75&GMG&4.0&\gls{lsbp}&4.875&\gls{knn}&4.375\\\hline

\gls{mog2}&4.875&\gls{mog}&4.375&GMG&4.625&\gls{gsoc}&5.5&\gls{mog2}&5.5&\gls{mog}&5.25&\cellcolor{gray}\textcolor{white}{\protect\textbf{\gls{hsmd}}}&\cellcolor{gray}\textcolor{white}{\protect\textbf{4.875}}&CNT&4.75&\gls{gsoc}&4.875&\gls{mog2}&5.125&CNT&4.375\\\hline

\gls{mog}&5.0&GMG&5.5&\gls{knn}&4.875&\gls{lsbp}&5.875&GMG&5.875&CNT&5.25&\gls{knn}&5.5&\gls{lsbp}&5.25&\gls{knn}&5.875&\gls{mog}&5.25&\gls{gsoc}&5.375\\\hline

\gls{knn}&6.375&\gls{gsoc}&6.0&\gls{mog}&6.625&\gls{mog2}&6.125&\gls{gsoc}&6.25&GMG&5.5&\gls{gsoc}&5.5&GMG&5.25&\gls{lsbp}&5.875&\gls{knn}&5.5&\gls{mog}&6.0\\\hline

CNT&6.75&\gls{lsbp}&6.125&\gls{mog2}&6.75&\gls{mog}&6.625&\cellcolor{gray}\textcolor{white}{\protect\textbf{\gls{hsmd}}}&\cellcolor{gray}\textcolor{white}{\protect\textbf{6.5}}&\gls{knn}&5.5&GMG&6.25&\gls{mog2}&6.75&CNT&6.0&\cellcolor{gray}\textcolor{white}{\protect\textbf{\gls{hsmd}}}&\cellcolor{gray}\textcolor{white}{\protect\textbf{6.625}}&GMG&6.25\\\hline
\end{tabular}}
\end{center}
$\downarrow$: the lowest result is the best.\\
The \gls{hsmd} results are highlighted using dark grey for the \gls{hsmd}, and the best results of other methods are highlighted using light grey.\\
\gls{r} is the average ranking.\\
The baseline, cameraJitter, intObjMotion, shadow and thermal categories are shared between the \gls{cdnet2012} and \gls{cdnet2014} datasets. While the badWeather, dynamicBackground, nightVideos, PTZ, lowFrameRate and turbulence categories are specific to the \gls{cdnet2014} dataset.
\end{table*}
\begin{figure*}
\centering
\includegraphics[width=1.5\textwidth]{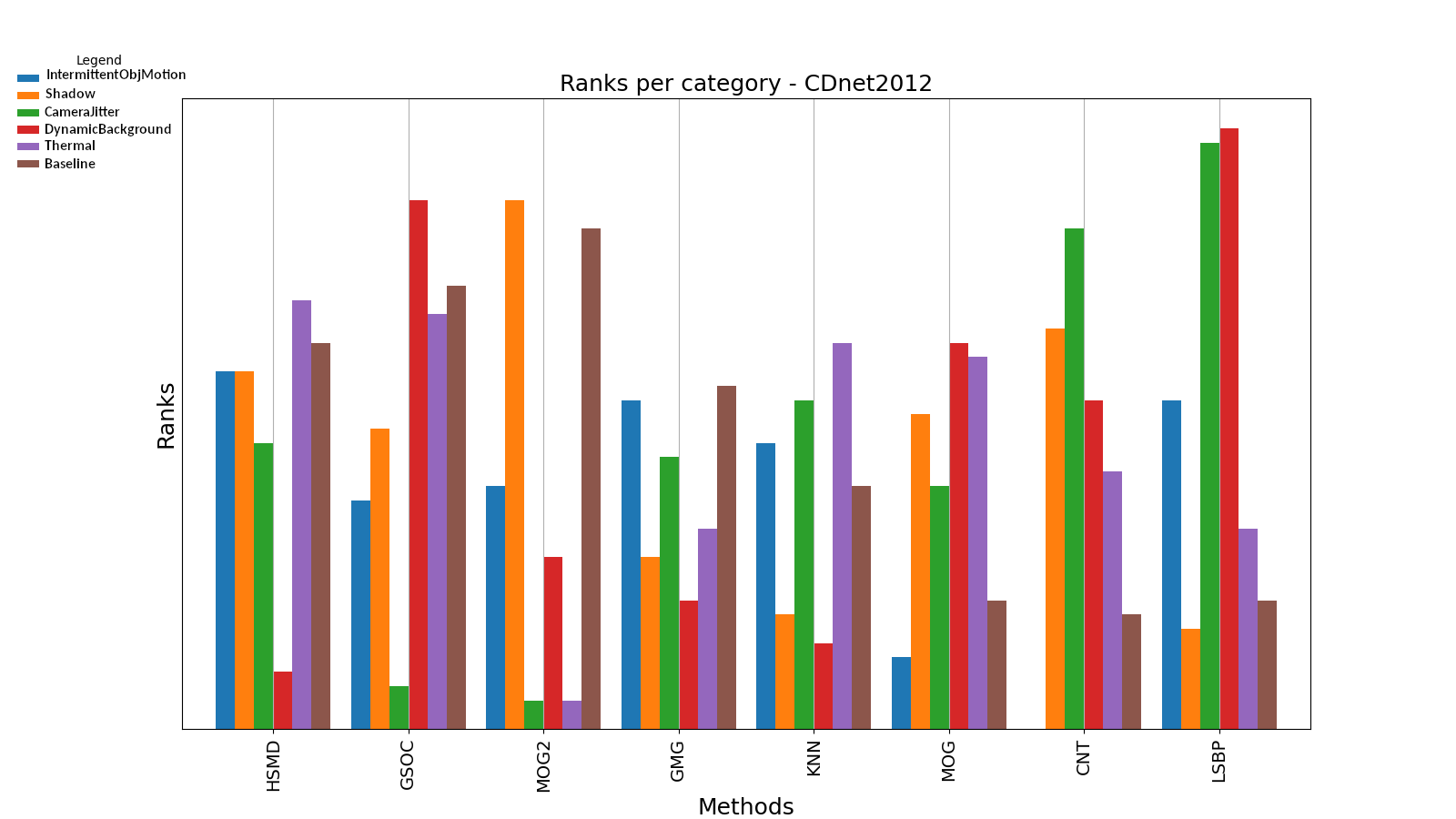}
\caption{\gls{cdnet2012} overall result based on \acrfull{r} per method. The highest bars show the higher ranks, and it is clear that none of the methods had the best ranks in all the categories. Furthermore, it is possible to see that the \gls{hsmd} achieved high ranks across all the categories, except dynamic background.} \label{fig:CDnet2012_results}
\end{figure*}

\begin{figure*}
\centering
\includegraphics[width=1.5\textwidth]{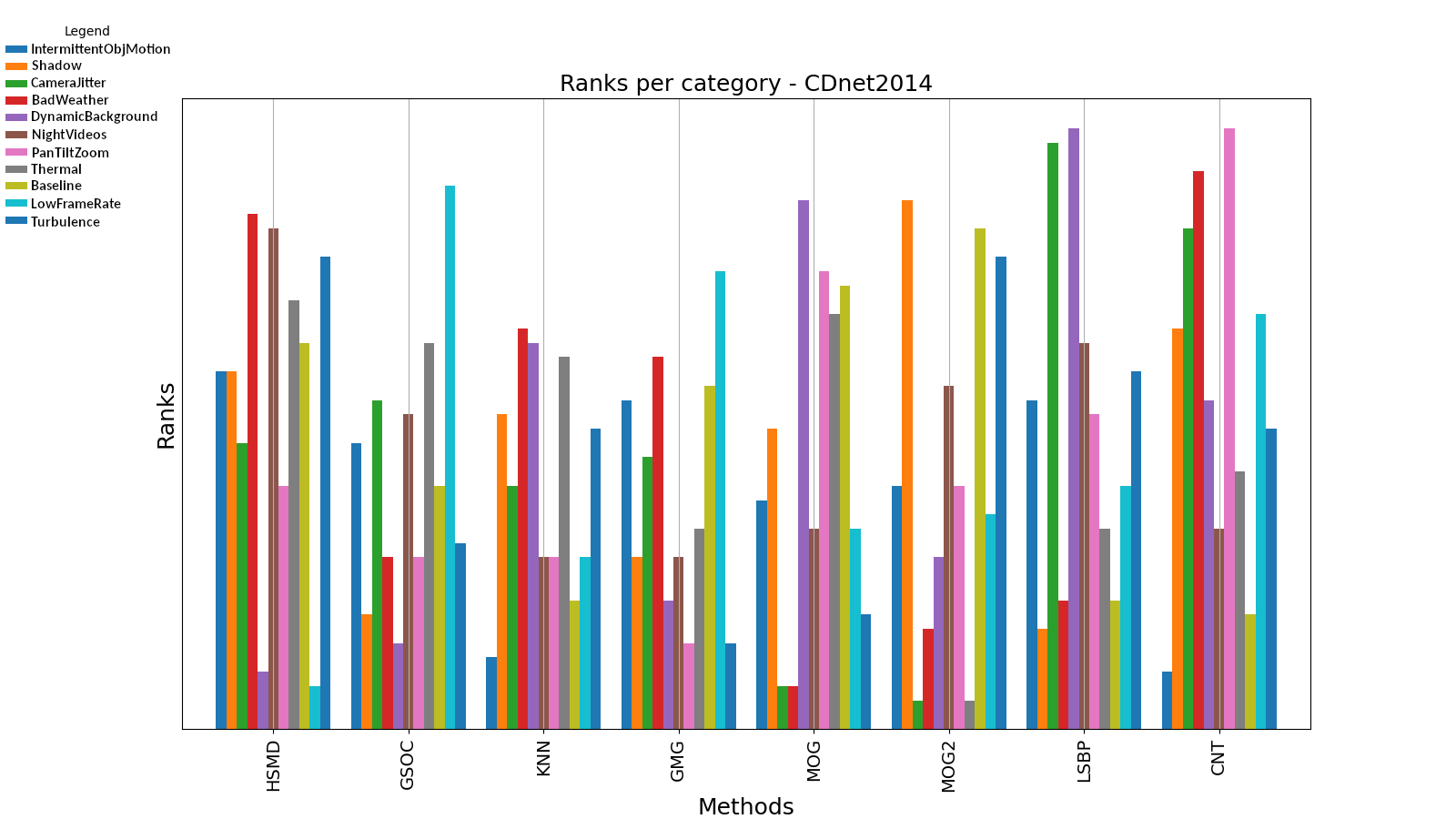}
\caption{\gls{cdnet2014} overall result based on \acrfull{r} per method. The highest bars show the higher ranks, and it is clear that none of the methods had the best ranks in all categories. Furthermore, it is possible to see that the \gls{hsmd} achieved high ranks across most of the categories, except dynamic background and low frame rate.} \label{fig:CDnet2014_results}
\end{figure*}
\end{landscape}
Figures~\ref{fig:CDnet2012_results}~and~\ref{fig:CDnet2014_results} show the variation of the ranks obtained per category and per method. 

Based on the results shown in Table \ref{tab:results_category}, Figure \ref{fig:CDnet2012_results} and Figure \ref{fig:CDnet2014_results}, the \gls{hsmd} performs better when processing image sequences from intermittent object motion, night vision, baseline, and turbulence categories. These categories contain moving objects with high contrast ration, which is ideal for sensing by the spiking neurons. Overall, the \gls{hsmd} has improved the results of the \gls{gsoc} in 8 of the 11 categories (see Table \ref{tab:results_category}). It is also easy to infer that the \gls{hsmd} exhibits the lowest \gls{r} variation, which justifies the \gls{hsmd} being ranked first.

\subsection{Results analysis}
\label{Ch4.4.1:results_analysis}

The \gls{hsmd} performed very poorly in the dynamic background and low frame rate categories, suggesting that the spiking neuron model is not ideal for distinguishing the type of motion. i.e. the spiking neurons detect motion but are unable to distinguish between a shadow or the object itself. This result is probably because, in vertebrate retinas, only the ganglion cells are spiking cells, suggesting that distinction between the main object and shadows is probably performed by other non-spiking cells. Nevertheless, the creation of the new approach incorporating both the \gls{gsoc} algorithm and the \gls{snn}, which emulates the basic \gls{oms-gc} functionality, clearly improves the accuracy of the \gls{gsoc} algorithm.\\
The \gls{cdnet2012} and \gls{cdnet2014} datasets are composed of image files of different resolution, and accordingly, the processing times vary. The \gls{hsmd} takes approximately 72.4ms (\gls{cdnet2014}) and 71.9ms (\gls{cdnet2012}) to process images of 720$\times$480 on a 96-cores Intel(R) Xeon(R) Platinum 8160 CPU @ 2.10GHz equipped with 792 GB of DDR4 and 12.7 TB of disk space. The slight variations are related to other applications running in the background. Therefore, the \gls{hsmd} is capable of processing images of 720$\times$480 at an average speed of 13.82fps (\gls{cdnet2014}) and 13.92fps (\gls{cdnet2012}). Finally, the \gls{hsmd} is the first hybrid \gls{snn} algorithm capable of processing images at such a frame rate, as far as the authors are aware.

\section{Discussion} \label{Ch4.5:discussion}

A bio-inspired \gls{hsmd} has been proposed to detect object motion and assess against the \gls{cdnet2012} and \gls{cdnet2014} datasets. The \gls{hsmd} was written in C++ using the C++ \gls{stl} 17 (implementation of data structures), Boost 1.71 (file management), and \gls{opencv} 4.5.0 (image processing). These incorporate video sequences of many moving objects under various challenging environmental conditions and are widely used for benchmarking \gls{bs} algorithms. The \gls{cdnet2012} is composed of 6 categories of movements, and the \gls{cdnet2014} augments the initial 6 to 11 categories of movements. Eight metrics, utilised as standard in the Change Detection datasets, were used to assess and compare the quality of the \gls{hsmd} algorithm. \\

The \gls{hsmd} performed poorly in the dynamic backgrounds and low frame rate categories, indicating that the spiking neuron model is not the best for classifying these types of object motion (i.e. the spiking neurons are able to detect motion but are unable to distinguish a shadow from the actual object). This deficiency is most likely caused by the fact that only the ganglion cells in vertebrate retinas are spiking cells, indicating that other non-spiking cells are most likely responsible for differentiating between the main object and shadows. However, the development of the new method that combines the \gls{gsoc} algorithm with a customised \gls{snn}, which mimics the fundamental \gls{oms-gc} functionality, significantly increases the \gls{gsoc} algorithm's accuracy. Nevertheless, the \gls{hsmd} algorithm performed overall best in both the \gls{cdnet2012} and \gls{cdnet2014} while performing better than all the tested \gls{bs} algorithms in the intermittent object motion, night videos, thermal and turbulence categories, second best in the bad weather category, and third-best in the baseline and shadow categories. The comparatively good results are a consequence of using the \gls{snn} for emulating the basic functionality of \gls{oms-gc}, which improves the sensitivity of the \gls{hsmd} to object motion. The \gls{hsmd} is also the first hybrid \gls{snn} algorithm capable of processing video/image sequences with near real-time performance (i.e. 720$\times$480@13.82fps [\gls{cdnet2014}] and 720$\times$480@13.92fps [\gls{cdnet2012}]). \\
The main lesson learnt from the implementation of the \gls{hsmd} was that the \gls{snn} latency need to be improved for targetting real-time applications. As already mentioned in Section~\ref{Ch2.2:snn}, \glspl{snn} are massively parallel and can be accelerated using dedicated hardware such as \glspl{fpga} (see~\ref{Ch2.6.3:fpga}). The hardware implementation of the \gls{hsmd} is covered in Chapter~\ref{Ch5:neuromorphic_object_motion_detector}.\\

%% file: Chapter5/chapter5.tex
%%%%%%%%%%%%%%%%%%%%%%%%%%%%%%%%%%%%%%%%%%%%%%%%%%%%%%%%%%%%%%%%%%%%%%%%%%%%%%%%
%2345678901234567890123456789012345678901234567890123456789012345678901234567890
% 1 2 3 4 5 6 7 8
% THESIS CHAPTER

\chapter{NeuroHSMD: Neuromorphic Hybrid Spiking Motion Detection} 
\label{Ch5:neuromorphic_object_motion_detector}
The \gls{neurohsmd} is the hardware implementation of the \gls{hsmd} discussed in Chapter~\ref{Ch4:object_motion_detection_cells}. As discussed in section~\ref{Ch2.6.3:fpga}, \glspl{fpga} offer the desired flexibility to accelerate massively parallel \gls{snn} architectures. Therefore, a high-end \gls{fpga} was selected for accelerating the \gls{hsmd}'s \gls{snn}. \gls{opencl} was used to describe the customised \gls{snn} because it provides a higher level of abstraction than other \gls{hdl} tools (see Section \ref{Ch5.2.4:opencl} for more details), increase in productivity, and compatibility with other compatible devices such as non-Intel \glspl{fpga}, \glspl{cpu} and \glspl{gpu}. The results show that the \gls{neurohsmd} was 82\% faster processing $720 \times 480$ than the \gls{hsmd} algorithm. Furthermore, the \gls{neurohsmd} was ranked first alongside the \gls{hsmd} when benchmarked against \gls{cdnet2012} and \gls{cdnet2014}, meaning that there was no degradation in the quality of the foreground extraction.

\section{Introduction} \label{Ch5.1:introduction}

The human brain is characterised by its tolerance to faults/noise, concurrent processing capabilities, flexibility, and high level of parallelisation when processing data. Furthermore, the adult human brain has a power consumption of about 400 Kcal per day, equivalent to 25 Watts \cite{Pastur-Romay2017}. Again, the human brain can reach 10-50 petaflops, outperforming any \gls{cots} \gls{cpu} \cite{Brooks2012}. Despite the fact that \glspl{cpu} outperform the human brain by several orders of magnitude when processing and transmitting sequential signals, the human brain processes millions of signals in parallel using its massively parallel circuits \cite{Pastur-Romay2017, Brooks2012}.

The human brain is composed of millions of interconnected neuron circuits composed of different types of neuron cells and contributing to specific brain computations \cite{Herculano-Houzel2009, Fox2013}. While \glspl{cpu} transmit signals at a few tenths of a gigahertz, neuronal circuits transmit signals at hundreds of gigahertz \cite{Herculano-Houzel2009, Fox2013}. Nevertheless, the human brain can outperform \glspl{cpu} when processing signals from complex systems such as the auditory and visual systems because of its massive parallel structure \cite{Fox2013}

\glspl{gpu} and \glspl{fpga} are parallel processing devices that can be used in conjunction with \glspl{cpu} to accelerate parallellisable algorithms. \glspl{gpu} are specialised electronic circuits with a flexible architecture designed for parallel processing of graphics and video rendering and accelerating some types of \gls{ai} algorithms \cite{Intel2020b}. \glspl{fpga} are integrated circuits composed of built-in interconnected hardware blocks that can be freely reprogrammable after manufacturing \cite{Intel2020}. In contrast to \glspl{gpu}, which have a well-defined architecture, \glspl{fpga} are flexible devices that allow the user to describe new hardware architectures, such as brain-like, neuromorphic architectures.

The \gls{hsmd} algorithm has proven to be very robust in extracting the foreground from the background (see section~\ref{Ch4:object_motion_detection_cells}) as a direct consequence of using a \gls{snn} to emulate the basic functionality observed in \glspl{oms-gc}. Furthermore, the \gls{hsmd} improves the \gls{gsoc} algorithm by employing a customised \gls{snn} composed of three layers of interconnected neurons through a 1:1 synaptic connectivity. Nonetheless, the \gls{hsmd} is substantially slower than the \gls{gsoc} because the customised \glspl{snn} also introduced a substantial delay because they are, by their parallel nature, not optimised for sequential processing architectures such as \gls{cpu} architectures. Therefore, \glspl{fpga} were the obvious choice for accelerating the \gls{hsmd}'s \gls{snn} because of their flexibility to describe massively parallel architectures. \glspl{fpga} are typically reprogrammed using \gls{vhdl} or Verilog, which are flexible but complex \glspl{hdl}. In recent years, the \gls{fpga} manufacturers have invested in \gls{hls} tools to enable users to programme \glspl{fpga} using C-like programming languages. One of the most successful \gls{hls} tools is \gls{opencl}, which allows users to program kernels that can be compiled targeting \glspl{cpu}, \glspl{gpu}, and \glspl{fpga}. The \gls{neurohsmd} presented in this chapter was implemented using \gls{opencl} on \glspl{fpga}.

The methodology is discussed in section~\ref{Ch5.2:methodology}; the results are presented in section~\ref{Ch5.3:results} and the discussion of the \gls{neurohsmd} results is performed in section~\ref{Ch5.4:discussion}.

\section{Implementation details} \label{Ch5.2:methodology}
\glspl{snn} are composed of a variable number of spiking neurons, and each neuron's output will contribute to the generation of spike events. Spiking neuron models are therefore highly parallelisable and not computationally suitable for \glspl{cpu}. In this section, the details of the hardware implementation are discussed. Heterogeneous computing platforms are discussed in section~\ref{Ch5.2.1:hardware_architectures}; the architecture of \gls{fpga} devices is discussed in section~\ref{Ch5.2.2:fpga_architecture}; \gls{hdl} are discussed in section~\ref{Ch5.2.3:hdl}, \gls{opencl} framework is introduced in section~\ref{Ch5.2.4:opencl}; the selected hardware platform is discussed in section~\ref{Ch5.2.5:hardware_platform}; and the details about the host application and its device kernels are described in section~\ref{Ch5.2.6:NeuroHSMD}.

\subsection{Heterogeneous computing platforms} \label{Ch5.2.1:hardware_architectures}
The rise of \gls{ai} and the continuous generation of big data are creating computational challenges. \glspl{cpu} are not enough to efficiently run state-of-the-art \gls{ai} algorithms or process all the data generated by a wide range of sensors. World-leading processing technology companies (such as NVIDIA, AMD, Intel and ARM) have been looking closely into the new requirements. They have been pushing the boundaries of technology to deliver efficient and flexible processing solutions. 

Heterogeneous computing refers to the use of different types of processor systems in a given scientific computing challenge. 

Heterogeneous platforms are composed of different types of computational units and technologies. Such media can be composed of multicore \glspl{cpu}, \glspl{gpu} and \glspl{fpga} acting as computational units and offering the flexibility and adaptability demanded by a wide range of application domains \cite{SicaDeAndrade2018}. These computational units can significantly increase the overall system performance and reduce power consumption by parallelising concurrent operations that require substantial \gls{cpu} resources over long periods. 

Accelerators like \glspl{gpu} and \glspl{fpga} are massive parallel processing systems that enable accelerating portions of code that are parallelisable. Combining \glspl{cpu} with \glspl{gpu} and \glspl{fpga} helps improve the performance by assigning different computational tasks to specialised processing systems. \glspl{gpu} are optimised to perform matrix multiplications in parallel, which is the major bottleneck in video processing and computer graphics. Nevertheless, \glspl{gpu} also introduce hardware and environmental limitations (e.g. high-power consumption and architectural limitations) \cite{Intel2020b}. \glspl{snn} are massively parallel in their nature and not suitable for matrix representation because each neuron can be considered a node containing several sequential mathematics operations. Therefore, a \gls{fpga} device was selected to accelerate the \gls{hsmd}'s customised \gls{snn}.

\gls{opencl} is a C/C++-based programming language specially designed for software developers to write applications targeting heterogeneous computing platforms such as \glspl{cpu}, \glspl{gpu}, and \glspl{fpga} \cite{Kronos2021}. \gls{opencl} provides an abstraction layer allowing compatibility across devices and enabling the same source code to run on different device architectures. Furthermore, \gls{opencl} provides facilities for developers to control parallelism and make effective use of the target device resources \cite{Intel2019}. The \gls{opencl} framework was chosen to ensure that the \gls{neurohsmd} is widely compatible with a wide range of hardware devices (for more information on \gls{opencl}, see section \ref{Ch5.2.4:opencl}).

\subsection{FPGA Architecture} \label{Ch5.2.2:fpga_architecture}
\glspl{fpga} have been used, for many decades, accelerating applications, including edge/cloud computing. \glspl{fpga} are flexible devices because of their flexible architecture, enabling developers to describe customised architectures. Such flexibility comes with a downside because \glspl{fpga} are also known by their associated complexity. There are two main \gls{fpga} manufacturers, namely, Intel\footnote{Available online, \protect\url{https://www.intel.co.uk/content/www/uk/en/products/programmable/fpga.html}, last accessed: 04/03/2021} and AMD-Xilinx\footnote{Available online, \protect\url{https://www.xilinx.com/}, last accessed: 04/03/2021}. The Computational Neurosciences and Cognitive Robotics (CNCR) group, where the PhD programme was developed, has different Intel \gls{fpga} development boards. The target board is fitted with a state-of-the-art Stratix 10 SoC \gls{fpga} device\footnote{Available online, \protect\url{https://www.intel.co.uk/content/www/uk/en/products/programmable/soc/stratix-10.html}, last accessed: 07/04/2021} (see further details in the next section~\ref{Ch5.2.5:hardware_platform}). Therefore, this chapter focus on the Intel Stratix architecture.

\begin{figure}[htb!]
\centering
\includegraphics[width=0.8\textwidth]{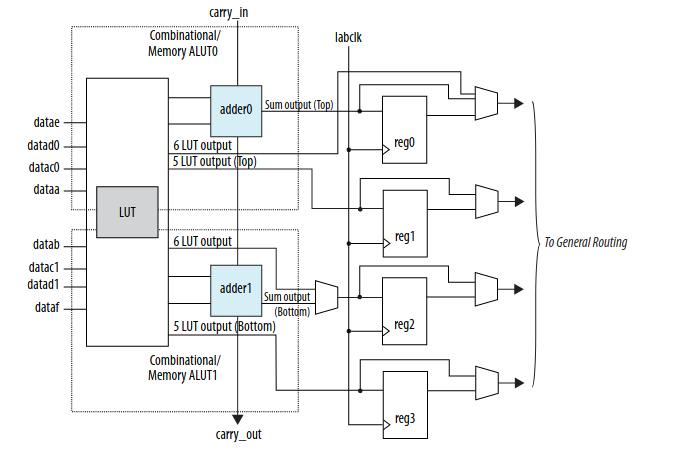}
\caption{\gls{alm} Block Diagram. Each register has the following ports: i) Data in, ii) Data out, iii) Clock, iv) clock enable, v) synchronous clear and vi) asynchronous clear. Adopted from \protect\cite{Intel2020c}.} \label{fig:alm}
\end{figure}

The Intel Stratix family is composed of \gls{lab} made up of 10 basic building blocks called \glspl{alm}. Each \gls{alm} is composed of fractionable Look-Up-Tables, also known as \gls{alut}, two-bit full adder and four registers (see Figure~\ref{fig:alm}).

\glspl{lab} can be freely reconfigured to implement logic and arithmetic functions. Furthermore, up to a quarter of the \glspl{lab} can be used as \gls{mlab}. Each LAB contains dedicated logic elements used to drive control signals to \glspl{alm}. Each \gls{mlab} supports up to 640 bits of simple dual-port \gls{sram}. It is possible to configure each \gls{alm} in an \gls{mlab} as $32 \times 2$ memory blocks equivalent to $32 \times 2 \times 10$ simple dual-port \gls{sram} blocks. Dual-port \glspl{sram} are low-latency memory devices that only take a clock cycle to perform a read/write operation (for example, \gls{sdram} in \glspl{cpu} takes thousands of clock cycles to complete read/write operations).

\subsection{Hardware Description Language} \label{Ch5.2.3:hdl}
The behaviour of \glspl{lab} and \glspl{alm} can be programmed using \gls{hdl}. Although many \glspl{hdl} being available, the \gls{ieee} endorses the \gls{vhdl} and Verilog. Although the \gls{vhdl} syntax is identical to Pascal and the Verilog syntax is identical to C, both languages are easy to learn and hard to master. 

\begin{figure}[htb!]
\centering
\includegraphics[width=0.8\textwidth]{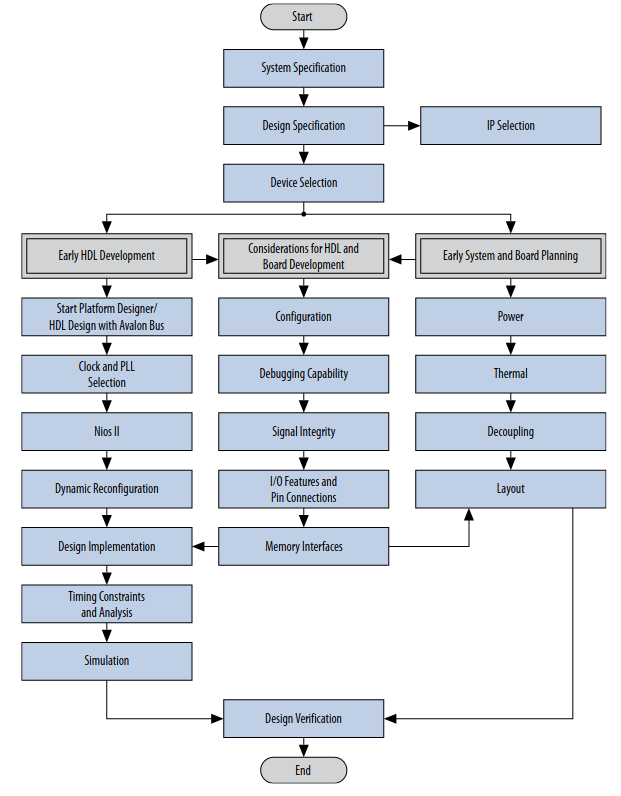}
\caption{Intel Stratix 10 device design flow \protect\cite{Intel2020}. The main stages of the design flow include the system specification, device selection, early system and board planning, pin connection considerations for board design, I/O and Clock planning, design entry, design implementation, analysis and optimisation and verification. Adopted from \protect\cite{Intel2020d}.} \label{fig:design-flow}
\end{figure}

\glspl{hdl} are potent tools because they enable users to programme at the \gls{rtl} (lowest level of coding), which is challenging to master. Hardware developers must have a deep knowledge of the reprogrammed device. Projects designed for a specific \gls{fpga} device might not work in another, even if they belong to the same \gls{fpga} family. Furthermore, the debugging of \gls{hdl} source code is very slow and prone to errors, making applications a long and time-consuming process due to the complex \gls{fpga} architecture (see Figure~\ref{fig:design-flow}).

\gls{fpga} manufacturers have been simplifying the \gls{fpga} design flow through the \gls{hls}. \gls{hls} tools deliver C/C++ like tools providing higher-level hardware abstraction, enabling software developers to use \glspl{fpga}. \gls{opencl}, a \gls{hls} framework, to be discussed in the next section~\ref{Ch5.2.4:opencl} was used to avoid having to use \gls{hdl} when implementing the \gls{neurohsmd}. Moreover, the \gls{neurohsmd}'s \gls{snn} was written in C++ for \gls{opencl} and automatically converted to \gls{hdl} using  Intel \gls{fpga} tools. 

\subsection{OpenCL} \label{Ch5.2.4:opencl}
\begin{figure}[htb!]
\centering
\includegraphics[width=0.8\textwidth]{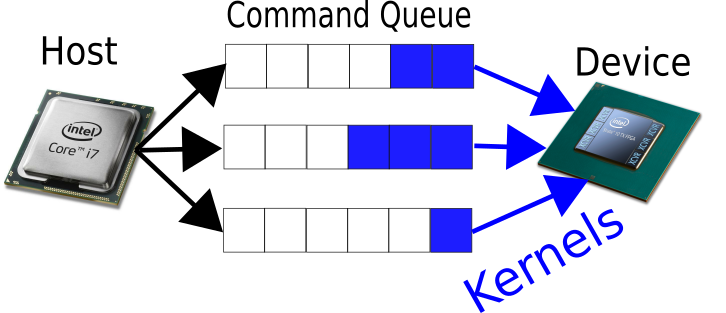}
\caption{Representation of a \gls{opencl} host application and three device kernels} \label{fig:host-device}
\end{figure}

\gls{opencl} applications are split into two parts, namely, \textbf{host} application(s) and \textbf{device} kernel(s) (see Figure~\ref{fig:host-device}). The \textbf{host} application(s) is(are) always compiled on the host Operating System and run on a \gls{cpu}. \textbf{Host} applications are also used to launch the target kernels on the target \textbf{devices}. Kernels are special functions written in \gls{opencl} C/C++ to perform parallelisable computations on accelerators such as \glspl{gpu} and \glspl{fpga} \cite{Intel2019}. For instance, consider two m$\times$n matrices, A and B where it is expected to do the operation C=A+B where C is the third matrix of m$\times$n. In this case, the kernel could just perform, in parallel, the addition of matrices A and B elements and store the result in C. Unlike in \glspl{cpu}, where it would take m$\times$n to do this matrix addition, \glspl{gpu} and \glspl{fpga} could parallelise this operation depending on the resources available per device resulting in the acceleration of the application.

Buffer objects within a context are used in \gls{opencl} to exchange data between the host and device \cite{CodeProjects2011}. The Intel \gls{fpga} \gls{sdk} for \gls{opencl} offline compiler optimises the kernel throughput by adjusting buffer sizes during the kernel compilation process \cite{Intel2021}. \gls{opencl} provides both mapped and asynchronous buffers, enabling the application to continue to run while additional data is exchanged (see Figure~\ref{fig:host-device}).

Software developers have to carefully analyse the code to be optimised and only select the sections that may benefit from hardware acceleration because the maximum speed is always dictated by the \gls{pcie} bus speed. Another big challenge for software developers is the low debugging capabilities available while the code is being executed on the device.

Although it is possible to use \gls{opencl} to program \gls{fpga} and \gls{gpu} devices, \glspl{gpu} are specialised devices designed for video rendering and graphics processing. At the same time, \glspl{fpga} are customisable devices that can be freely reconfigurable. Therefore, \glspl{fpga} offer more flexibility than \glspl{gpu}, which is desirable for accelerating \glspl{snn} because they can be modelled using the \glspl{noc} concept. Each individual spiking neuron in the node interconnects to one or more nodes of the same \gls{snn}. The flexibility offered by both \glspl{fpga} and \gls{opencl} makes selection of \glspl{fpga} over \glspl{gpu} the obvious choice.

\begin{figure}[htb!]
\centering
\includegraphics[width=1\textwidth]{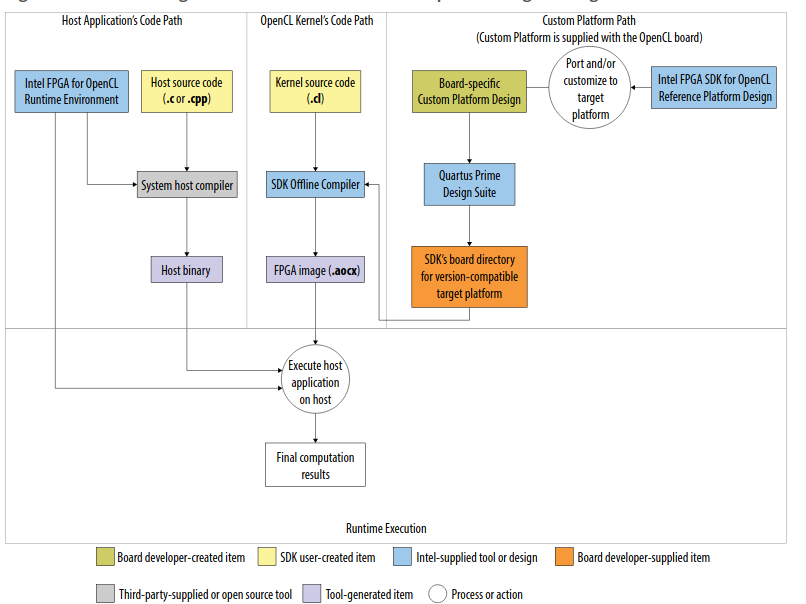}
\caption{Intel \gls{fpga} \gls{sdk} for \gls{opencl} \gls{fpga} design flow. Adopted from \cite{Intel2021b}.} \label{fig:opencl-design}
\end{figure}

Intel \gls{fpga} \gls{sdk} for \gls{opencl} (IOCL) provides a compiler and powerful tools to build and run \gls{opencl} applications targeting Intel \gls{fpga} devices. The IOCL generates two main components: the host application and the \gls{fpga} programming bitstream(s). The IOCL offline compiler (AOC) first compiles the custom kernel(s) to an image file (*.aocx) that will be used to program the \gls{fpga}. In contrast, the host-side C/C++ compiler compiles the host application and then links it to the IOCL runtime libraries (see Figure~\ref{fig:opencl-design}).

\begin{figure}[htb!]
\centering
\includegraphics[width=1\textwidth]{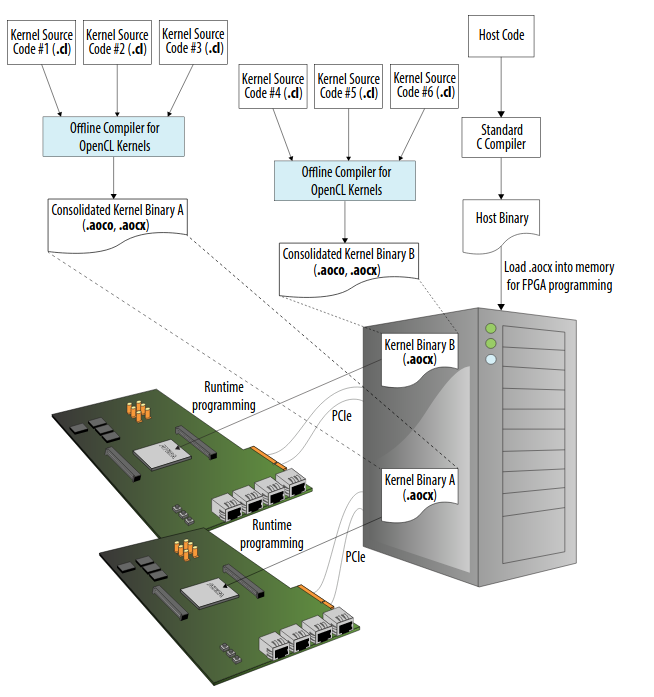}\caption{Intel \gls{fpga} \gls{sdk} for \gls{opencl} \gls{fpga} programming flow. Adopted from \cite{Intel2021b}.} \label{fig:opencl-programming}
\end{figure}

The DE10-pro platform provides the board support package (*.bsp). The AOC targets the DE10-pro platform when compiling an \gls{opencl} kernel to generate the \textbf{*.aocx} object that is \textbf{only compatible} with the DE10-pro's Intel Stratix 10 \gls{fpga} device. The IOCL utility programme (aocl) is used to programme the \gls{fpga} device using the image file \textbf{*.aocx} is then used to programme the \gls{fpga} to enable the host application to exchange data with the kernel using \gls{opencl} buffers via the \gls{pcie} bus. Multiple \glspl{fpga} can be used by the same host application (see Figure~\ref{fig:opencl-programming}).

\begin{figure}[htb!]
\centering
\includegraphics[width=1\textwidth]{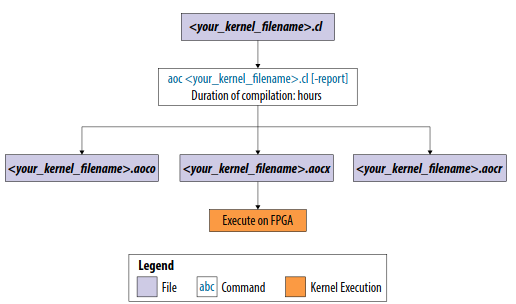}\caption{\gls{opencl} compilation flow. Adopted from \cite{Intel2021b}.} \label{fig:opencl-compilation-flow}
\end{figure}

The IOCL compiles one or more \gls{opencl} kernels and creates a hardware configuration file. After a successful compilation, the files \textbf{*.aocr}, \textbf{*.aoco}, \textbf(*.aocx), and reports/report.html are generated (see Figure \ref{fig:opencl-compilation-flow}). The report.html contains the estimated resource usage and a preliminary assessment of area usage. The intermediary \textbf{*.aoco} and \textbf{*.aocr} are only used in the generation of the \textbf{*.aocx} which is then used to programme the \gls{fpga}.

\subsection{Hardware platform} \label{Ch5.2.5:hardware_platform}
Although there are hundreds of \gls{fpga} boards, only a few boards provide board support packages (BSP) for \gls{opencl}. The creation of the \gls{opencl} BSP for a given \gls{fpga} is a complex and time-consuming process that considers the \gls{fpga} device and how the Inputs/Outputs are routed on the physical board. Therefore, the selection of the \gls{fpga} board should be made taking into consideration the size of the \gls{fpga} device, the \gls{opencl} BSP, the version of the BSP to ensure compatibility with recent Linux distributions and the reference/user manuals.

\begin{figure}[htb!]
\centering
\includegraphics[width=0.45\textwidth]{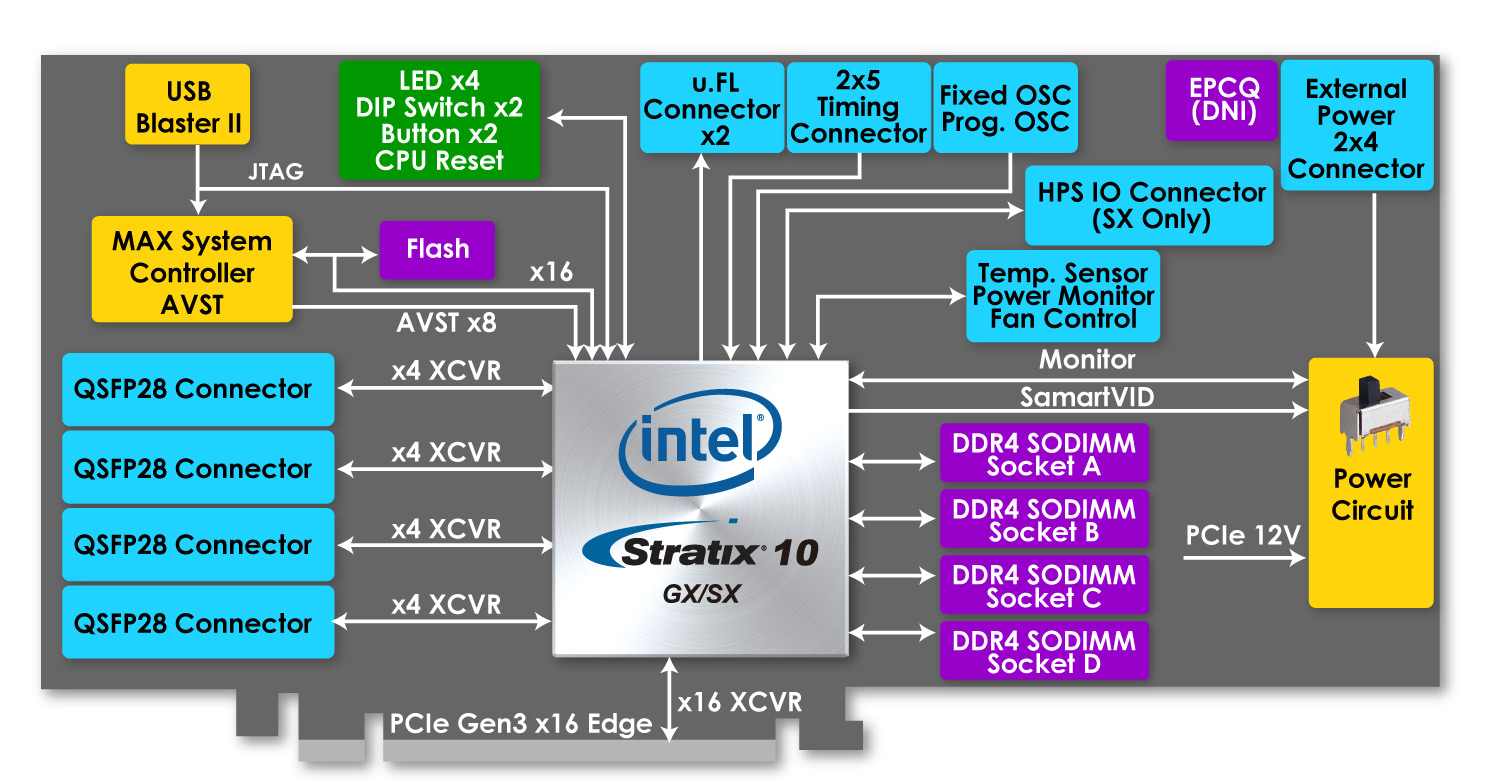}
\includegraphics[width=0.45\textwidth]{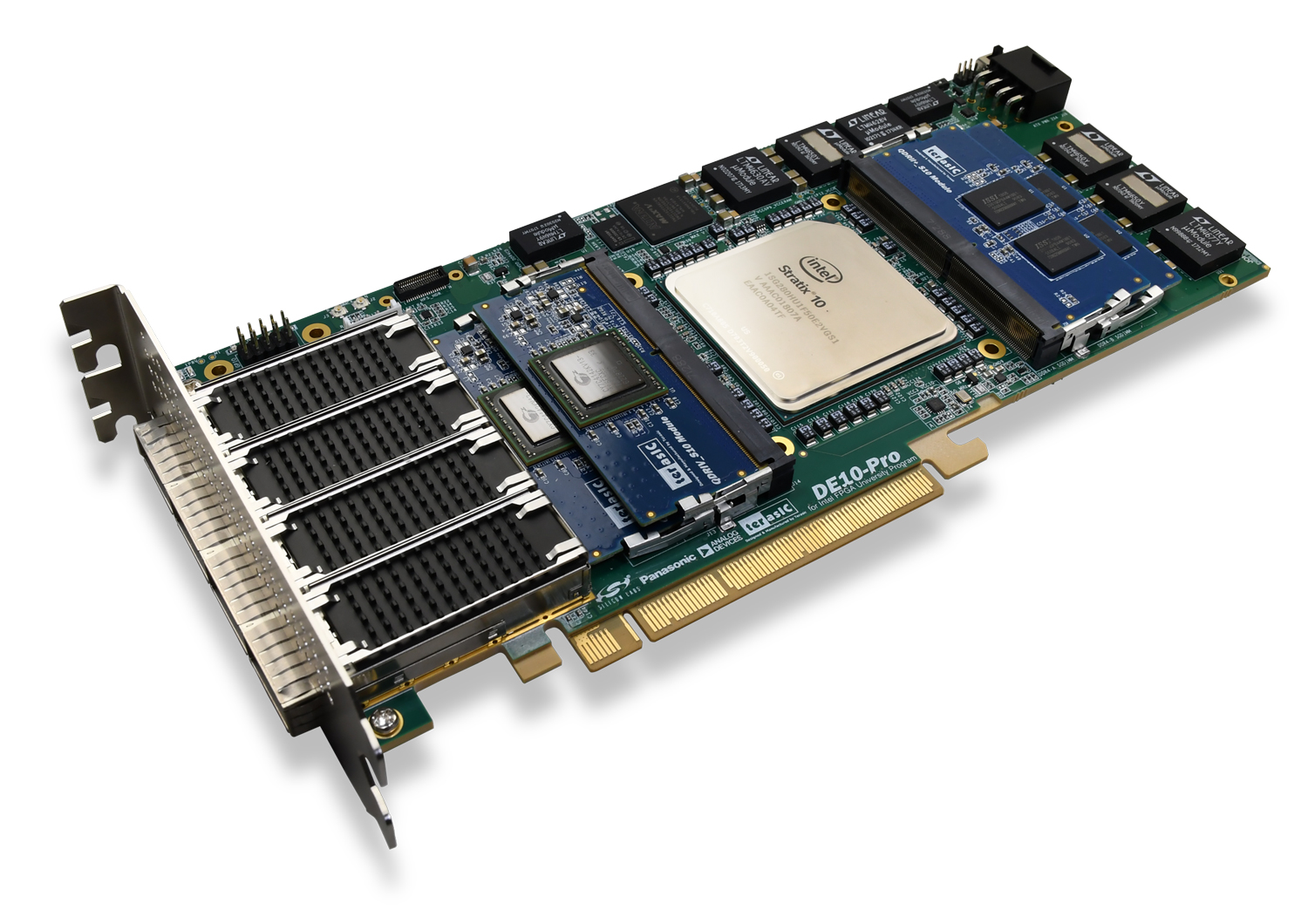}
\caption{Terasic DE10 pro development kit. (left) block diagram and (right) DE10 pro board. Adopted from \protect\cite{Terasic2021}} \label{fig:de10-pro}
\end{figure}

The Terasic DE10-pro development board \cite{Terasic2021} (see Figure~\ref{fig:de10-pro}) was used for implementing the \gls{neurohsmd} discussed in this chapter. The Terasic DE10-pro development board is equipped with a state-of-the-art high-end Intel Stratix 10 \gls{fpga} device. According to Terasic, the DE10 pro was created to meet the needs of \gls{ai}, data centers, and high-performance computing. Furthermore, the DE10-pro development board takes advantage of the latest Intel Stratix 10 to obtain high-speed and low-power (with up to 70\% lower power). It is equipped with a 32GB DDR4 memory module running at over 150 Gbps, up to 15.754 GB/s data transfer via \gls{pcie} Gen 3 x16 edge between \gls{fpga} and the host PC, and 4 onboard QSFP28 (100GbE) connectors.

The DE10 pro was installed on the host PC equipped with an Intel(R) Core(TM) i7-4770 \gls{cpu} @ 3.40GHz and 16 GB of DDR3 using the PCIe slot.

\subsection{NeuroHSMD implementation} \label{Ch5.2.6:NeuroHSMD}
\begin{figure}[htb!]
\centering
\includegraphics[width=0.5\textwidth]{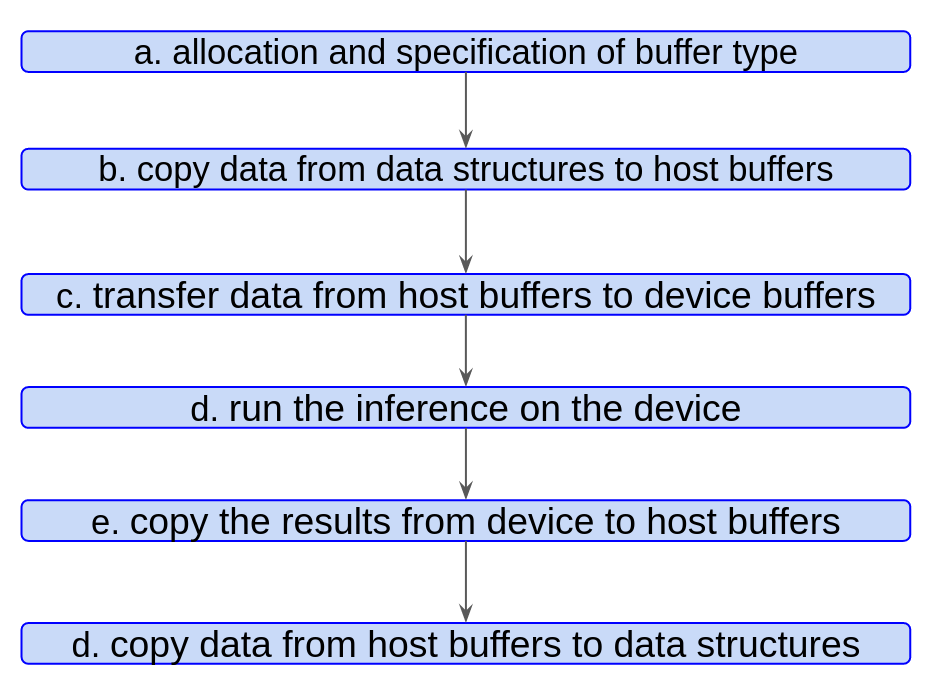}\caption{\gls{opencl} computation stages. The stages include: a) allocation and specification of buffer types on the host and device; b) copying data from the application data structures to host buffers; c) transferring data from host buffers to device buffers; d) running the inference on the device; e) copying the results from the device to host buffers; f) copying data from the host buffers to application data structures.} \label{fig:opencl_stages}
\end{figure}

The \gls{neurohsmd}'s \gls{snn} was written in C++ for \gls{opencl}. In \gls{opencl}, the data flow between host application and \gls{fpga} kernel is as follows: a) allocate and specify buffer types (read/write) on the host and device; b) copy \gls{fpga} data from application data structures to host buffers; c) transfer data from host buffers to device buffers; d) run the inference on the device; e) copy the results from the device to host buffers; f) copy data from host buffers to application data structures (see Figure~\ref{fig:opencl_stages}).\\

The \gls{neurohsmd} algorithm performs the following computation stages: 1) image capture, 2) conversion from colour to grey scale, and 3) background subtraction using the \gls{opencl}'s \gls{gsoc} algorithm, buffer the results and transfer buffered results to the \gls{fpga} device; 4) run the \gls{fpga} inference and wait for the spike results; 5) run the \gls{snn} kernel; 6) apply average filter; and 7) display and save the output image.

\begin{figure} [htb!]
\centering
\includegraphics[width=0.5\textwidth]{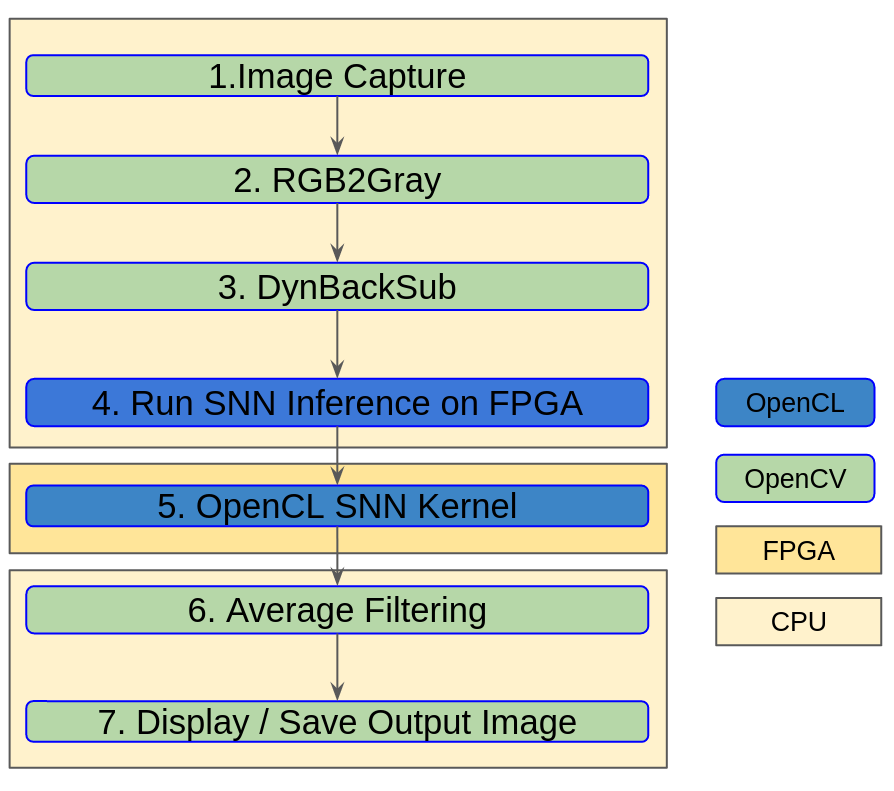}\caption{\gls{neurohsmd} computation stages. The \gls{opencl} implementation is represented in blue and the \gls{opencv} in green. The light yellow background represents the computation stages that run on the \gls{cpu} (i.e. steps 1, to 4, and 6 to 7), and in light orange, the stage that runs on the \gls{fpga} device (i.e. step 5)} \label{fig:neurohsmd_stages}
\end{figure}

\begin{figure}[htb!]
\centering
\includegraphics[width=1.0\textwidth]{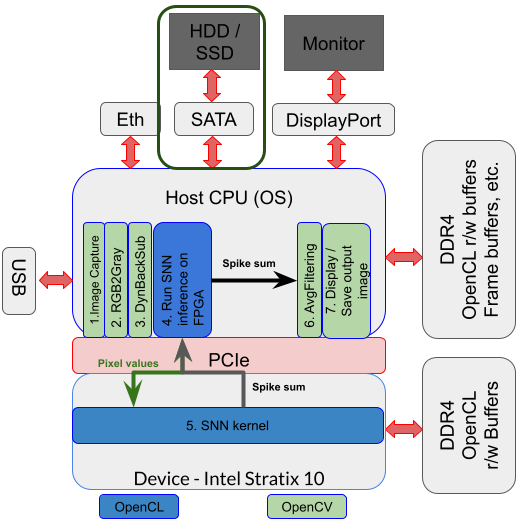}\caption{\gls{neurohsmd} architecture. The diagram represents the computation stages that run both on the \gls{cpu} and \gls{fpga}. Shows that the \gls{fpga} is connected to the host \gls{cpu} via the \gls{pcie} bus. It also shows the dedicated memory of the \gls{cpu} and \gls{fpga} device. This also includes how external devices (e.g. image sensors, monitor, and \gls{hdd}/\gls{ssd}) connect to the host \gls{cpu} via different interfaces (e.g. Ethernet (eth), \gls{sata}, display port, and us\gls{usb}b). The computation stages implemented in \gls{opencl} are in blue and \gls{opencv} in green.} \label{fig:NeuroHSMD_arc}
\end{figure}

The \gls{neurohsmd}'s processing stages are summarised in the diagram \ref{fig:neurohsmd_stages} and the \gls{neurohsmd} architecture is depicted in Figure~\ref{fig:NeuroHSMD_arc}.
 
Furthermore, the \gls{neurohsmd} implementation can be split into the \gls{nha} and the \gls{ndk}. The \gls{nha} and \gls{ndk} are described in detail in the sections \ref{Ch5.2.6.1:host} and \ref{Ch5.2.6.2:kernels}, respectively.

\subsubsection{Host application} \label{Ch5.2.6.1:host}
The \gls{nha} is used to interface the two \gls{ndk}. The \gls{nha} performs the first stages of pre-processing and performs inference of the SNN kernel (described on the \gls{fpga}), and uses the inference results to compute the \gls{bs} . Furthermore, the \gls{nha} can process both live images captured from camera devices or extracted from videos stored on \gls{usb} or \gls{sata} devices. The \gls{nha} is used to interface the \glspl{ndk}. The \gls{nha} performs the first stages of pre-processing and performs inference of the \gls{ndk} (described on the \gls{fpga}), and uses the inference results to compute the \gls{bs}. Furthermore, the \gls{nha} can process both live images captured from camera devices or extracted from videos stored on \gls{usb} or \gls{sata} devices.\\

Algorithm~\ref{alg:host} summarises each of the computation stages that occur in the \gls{nha}. 

\begin{algorithm}
\SetAlgoLined
\KwIn{\\
img: image frame;}
\KwOut{\\
post\_proc\_img:\ post\ processed\ image;\\
stats:\ computation\ statistics;}
\KwAlg{
\begin{algorithmic} [1]
 \STATE initialise\_opencl()\;

 \FOR{$iterator\gets list\_folders.begin()$ \TO list\_folders.end()}
 \STATE $(x,y) \gets get\_image\_size()$; 
 \STATE $num\_layers \gets 3$\;
 \STATE $tot\_neurons \gets x.y.num\_layers$\;
 \STATE reset\_opencl\_buffers(tot\_neurons )\;
 \STATE $gsoc \gets initialise\_gsoc\_back\_subtraction$\;
 \FOR{$iterator2\gets files\_list.begin()$ \TO files\_list.end()}
 \STATE $img \gets read\_image(iterator2);$\;
 \STATE $<pixel\_values> \gets gsoc.compute(img)$\;
 \STATE \textbf{aocl\_snn\_v1($<pixel\_values>$) $\rightarrow <spike\_sum>$}\; \COMMENT{\protect\textbf{Alg.~\ref{alg:NeuroHSMD-v1}}}
 \STATE \textbf{OR}
 \STATE \textbf{aocl\_snn\_v2($<pixel\_values>$) $\rightarrow <spike\_sum>$}\; \COMMENT{\protect\textbf{Alg.~\ref{alg:NeuroHSMD-v2}}}
 \STATE $post\_proc\_img \gets get\_spikes\_sum\_l3(<spike\_sum>)$ \;
 \STATE $save(post\_proc\_img)$ \;
 \STATE $stats \gets compute\_stats(time)$ \;
 \ENDFOR
 \STATE $save(stats)$ \;
 \ENDFOR
 \end{algorithmic}}
\caption{\gls{nha}}\label{alg:host}
\end{algorithm} 

The communication between the \gls{nha} and the \gls{ndk} is limited by the \gls{pcie} bus speed (~16 GB/s \footnote{Available online, \protect\url{https://www.trentonsystems.com/blog/pcie-gen4-vs-gen3-slots-speeds}, last accessed: 21/06/2021}). In the \gls{hsmd}, the limitations are only dictated by the \gls{cpu} speed and the DDR4 memory speed (~34.1 GB/s\footnote{Available online, \protect\url{https://www.crucial.com/support/articles-faq-memory/understanding-cpu-limitations-with-memory}, last accessed: 21/06/2021}) which two times faster than the \gls{pcie} bus speed. 

\subsubsection{Device kernels} \label{Ch5.2.6.2:kernels}
The \glspl{ndk} section covers the two kernels (i.e. NeuroHSMDv1 and NeuroHSMDv2) that have been implemented.
The only difference between the two \glspl{ndk} is that the \textit{$aocl\_snn\_v2$} only computes the spike sums for neurons that have the pixel intensity values above 0.0.\\

\begin{algorithm}
\SetAlgoLined
\setcounter{AlgoLine}{0}
\KwCircuits{16}
\KwConst{\\$R_m$: membrane\ resistance;\\
$\tau_m$: membrane time constant;\\
 $dt$: time step;\\
 $p2c$: pixel\ values\ to\ current;\\
 $steps$: number of steps; \\
 $s2c$: spike\ to\ current;}
\KwIn{\\$<pixel\_val>$: pixel values; \\
$num\_neuron\_layer$: number of neurons per layer;}
\KwOut{\\$<spk\_sum>$: spike sum;}
\KwAlg{
\begin{algorithmic} [1]
 \FOR{$neuron\_idx\gets0$ \TO number\_neurons}
 \STATE $I_s \gets pixel\_val[neuron\_idx].p2c$;\\
 \FOR{$dt1\gets0$ \TO steps}
 \STATE Layer 1:
 \STATE $compute\_V_m\_l1[neuron\_idx](I_s)$;\\
 \STATE $update\_spike\_sum\_l1[neuron\_idx](V_m\_l1)$;\\
 \STATE Layer 2:\\
 \STATE $I_s\_l2 \gets spike\_sum\_l1[[neuron\_idx].s2c$;\\
 \STATE $compute\_V_m\_l2[[neuron\_idx](I_s\_l2)$;\\
 \STATE $update\_spike\_sum\_l2[neuron\_idx](V_m\_l2)$;\\
 \STATE Layer 3:\\
 \STATE $I_s\_l3 \gets spike\_sum\_l2[neuron\_idx].s2c$;\\
 \STATE $compute\_V_m\_l3[neuron\_idx](I_s\_l2+I_s\_l3)$;\\
 \STATE $update\_spk\_sum[neuron\_idx](V_m\_l3)$;\\
 \ENDFOR
 \ENDFOR
 \end{algorithmic}}
\caption{aocl\_snn\_v1}\label{alg:NeuroHSMD-v1}
\end{algorithm} 

The \textit{$aocl\_snn\_v1$} was parallelised by a factor of 16. The \gls{ndk} version 1 is the equivalent implementation of the \gls{hsmd} algorithm described in chapter~\ref{Ch4:object_motion_detection_cells} which is referred to as HSMDv1 in this chapter.

The algorithm~\ref{alg:NeuroHSMD-v2}, inferred by the NHP, summarises the computation steps required to compute the spike sum.

\begin{algorithm}
\SetAlgoLined
\setcounter{AlgoLine}{0}
\KwCircuits{16}
\KwConst{\\$R_m$: membrane\ resistance;\\
$\tau_m$: membrane time constant;\\
 $dt$: time step;\\
 $p2c$: pixel\ values\ to\ current;\\
 $steps$: number of steps; \\
 $s2c$: spike\ to\ current;}
\KwIn{\\$<pixel\_val>$: pixel values; \\
$num\_neuron\_layer$: number of neurons per layer;}
\KwOut{\\$<spk\_sum>$: spike sum;}
\KwAlg{
\begin{algorithmic} [1]
 \FOR{$neuron\_idx\gets0$ \TO number\_neurons}
 \IF{$pixel\_val[neuron\_idx]>0.0$}
 \STATE $I_s \gets pixel\_val[neuron\_idx].p2c$;\\
 \FOR{$dt1\gets0$ \TO steps}
 \STATE Layer 1:
 
 \STATE $compute\_V_m\_l1[neuron\_idx](I_s)$;\\
 \STATE $update\_spike\_sum\_l1[neuron\_idx](V_m\_l1)$;\\
 \STATE Layer 2:\\
 \STATE $I_s\_l2 \gets spike\_sum\_l1[[neuron\_idx].s2c$;\\
 \STATE $compute\_V_m\_l2[[neuron\_idx](I_s\_l2)$;\\
 \STATE $update\_spike\_sum\_l2[neuron\_idx](V_m\_l2)$;\\
 \STATE Layer 3:\\
 \STATE $I_s\_l3 \gets spike\_sum\_l2[neuron\_idx].s2c$;\\
 \STATE $compute\_V_m\_l3[neuron\_idx](I_s\_l2+I_s\_l3)$;\\
 \STATE $update\_spk\_sum[neuron\_idx](V_m\_l3)$;\\
 \ENDFOR
 \ENDIF
 \ENDFOR
 \end{algorithmic}}
\caption{aocl\_snn\_v2}\label{alg:NeuroHSMD-v2}
\end{algorithm} 

The \textit{$aocl\_snn\_v2$} was parallelised by factor of 16. The \gls{ndk} version 2 contains an optimisation where the spike sum for a given neuron of layer 1 is only computed if the pixel intensity value is greater than 0.0. This way, the computations will only be done for neurons that receive stimuli from pixels that belong to the foreground.

This optimisation was also applied to the original \gls{hsmd} algorithm described in chapter~\ref{Ch4:object_motion_detection_cells}. In this chapter, the optimised version of \gls{hsmd} (i.e. where the spike sum for a given neuron of layer 1 is only computed if the pixel intensity value is greater than 0.0) is referred to as HSMDv2.

\subsection{Datasets and benchmark} \label{Ch5.2.7:datasets_comparison}
The NeuroHSMDv1, NeuroHSMDv2, HSMDv1 and HSMDv2 were tested against the \gls{cdnet2012} \cite{Goyette2012} and \gls{cdnet2014} \cite{Wang2014}. The scripts provided by Nil Goyette et al. \cite{Goyette2012} were used with the same protocol used to test the HSMDv1 algorithm reported in chapter~\ref{Ch4:object_motion_detection_cells}. The scripts provided by Nil Goyette et al. \cite{Goyette2012} enables the computation of the eight metrics (\gls{re}, \gls{sp}, \gls{fpr}, \gls{fnr}, PWC, PCC, \gls{pr} and \gls{f1}) and rank the results (\gls{arc}). More details about the eight metrics and the ranks' procedure can be found in section~\ref{Ch4.3.2:datasets_metrics}.

The benchmark of the four algorithms is required to ensure that the four algorithms produce comparative results to that of the original HSMDv1 results when tested against \gls{cdnet2012} and \gls{cdnet2014} datasets. Furthermore, the \gls{opencl} calls the Intel Quartus, which performs several hardware optimisations that may include converting from floating-point to fixed-point representation, which might affect the accuracy of the \gls{neurohsmd} algorithms during the synthesis step (one of the steps of the \gls{opencl} design flow). 

\section{Results} \label{Ch5.3:results}
The HSMDv1, HSMDv2, NeuroHSMDv1 and NeuroHSMDv2 were all tested on the same computer equipped with a quad-core Intel(R) Core(TM) i7-4770 \gls{cpu} @ 3.40GHz, 16GB of DDR3 @ 1600 MHz and 1TB of \gls{hdd}.
The results section is divided into three subsections. Namely, section \ref{Ch5.3.1:resources_usage} shows the resources' usage to enable the comparison between the two kernels' complexity, the speed performance results are presented in sections \ref{Ch5.3.2:performance} and section \ref{Ch5.3.3:benchmark} shows the benchmark results when tested against the \gls{cdnet2012} and \gls{cdnet2014} datasets.

\subsection{Resources Usage} \label{Ch5.3.1:resources_usage}
The resources' usage are given the report generated by the \textit{aoc} after the successful completion of the kernel compilation, which can take several hours (typically between 6h and 24h depending on the kernel complexity for the DE10pro). 
The resources' usage for the compilation of the NeuroHSMDv1 kernels is given in table~\ref{tab:resources_usage_v1} and the NeuroHSMDv2 kernel in table~\ref{tab:resources_usage_v2}. 

\begin{table}[H] \caption{NeuroHSMDv1 resources usage\strut} \label{tab:resources_usage_v1}
\resizebox{14.5cm}{!}{
\begin{tabular}{|l|l|l|l|l|l|}
\hline
\hline
\multicolumn{6}{||c||}{\huge{\textit{\textbf{Summary}}}} \\\hline\hline
\multicolumn{6}{|l|}{\Large\textbf{Info}} \\\hline
Project Name & \multicolumn{5}{|l|}{snn\_pc\_v1} \\\hline
Target Family, Device, Board & \multicolumn{5}{|l|}{Stratix 10, 1SG280LU3F50E1VGS1, de10\_pro:s10\_sh2e1\_4Gx2} \\\hline
AOC Version & \multicolumn{5}{|l|}{19.1.0 Build 240} \\\hline
Quartus Version & \multicolumn{5}{|l|}{19.1.0 Build 240 Pro} \\\hline
Command & \multicolumn{5}{|l|}{aoc device/snn\_pci\_v1.cl -o bin\_acl/snn\_pci\_v1.aocx } \\
& \multicolumn{5}{|l|}{-v -report -board=s10\_sh2e1\_4Gx2 -incremental}\\\hline\hline
\multicolumn{6}{|l|}{\Large{\textbf{Quartus Fit Clock Summary}}}\\\hline
Frequency (MHz) & \multicolumn{5}{|l|}{306.25 (fmax)}\\\hline\hline
\multicolumn{6}{|l|}{\Large{\textbf{Quartus Fit Resource Utilisation Summary}}} \\\hline
 & \textbf{\glspl{alm}} & \textbf{FFs} & \textbf{RAMs} & \textbf{\glspl{dsp}} & \textbf{\glspl{mlab}}\\\hline
 Full design (all kernels) & 387168.7 &	985558 & 2231 & 1408 & 4766 \\\hline
snn & 488257.1 & 1060084 & 2484 & 1024 & 3871 \\\hline\hline
\multicolumn{6}{|l|}{\Large{\textbf{Kernel Summary}}} \\\hline
Kernel Name & Kernel Type & Autorun & Workgroup Size& \# Compute Units & \begin{tabular}[c]{@{}l@{}}Hyper-Optimised\\ Handshaking\end{tabular} \\\hline
snn & 488257.1 & 1060084 & 2484 & 1024 & 3871 \\\hline\hline

\multicolumn{6}{|l|}{\Large{\textbf{Estimated Resource Usage}}} \\\hline
\multicolumn{1}{|c|}{\textbf{Kernel Name}} & \multicolumn{1}{c|}{\textbf{\glspl{alut}}} & \multicolumn{1}{c|}{\textbf{FFs}} & \multicolumn{1}{c|}{\textbf{RAMs}} & \multicolumn{1}{c|}{\textbf{\glspl{dsp}}} & \multicolumn{1}{c|}{\textbf{\glspl{mlab}}} \\\hline
snn & 488257.1 & 1060084 & 2484 & 1024 & 3871 \\\hline
\textbf{Global Interconnect} & 10629 & 16485 & 61 & 0 & 0 \\\hline
\textbf{Board Interface} & 13132 & 20030 & 112 & 0 & 0 \\\hline
\textbf{System} description ROM & 2 & 71 & 2 & 0 & 0 \\\hline
\textbf{Total} & 567523 (30\%) & 907449 (24\%) & 2963 (25\%) & 976 (17\%) & 3786 \\\hline
\textbf{Available} & 1866240 & 3732480 & 11721 & 5760 & 0 \\\hline\hline
\multicolumn{6}{|l|}{\Large{\textbf{Compile Warnings}}} \\\hline
\multicolumn{6}{|l|}{None} \\\hline\hline 
\end{tabular}}
\end{table}

From the analysis of table~\ref{tab:resources_usage_v1} can be seen that the estimated resource utilisation is more pessimistic than the final resources' utilisation, which is a direct consequence of the Intel Quartus' optimisations during the synthesis and routing phases and to ensure that the circuit fits in the \gls{fpga} device. Nevertheless, it takes about 5 minutes to get the \textit{estimated resources usage and between 6 and 24 hours to get the resource usage summary}. Therefore, it is a good practise to define the coefficient \textbf{N} in the statement \textbf{\# PRAGMA UNROLL N} based on the \textit{estimated resources' usage}.

\begin{table}[H] \caption{NeuroHSMDv2 resources usage\strut} \label{tab:resources_usage_v2}
\resizebox{14.5cm}{!}{
\begin{tabular}{|l|l|l|l|l|l|}
\hline
\hline
\multicolumn{6}{||c||}{\huge{\textit{\textbf{Summary}}}} \\\hline\hline
\multicolumn{6}{|l|}{\Large\textbf{Info}}\\\hline
Project Name & \multicolumn{5}{|l|}{snn\_pci\_v2} \\\hline
Target Family, Device, Board & \multicolumn{5}{|l|}{Stratix 10, 1SG280LU3F50E1VGS1, de10\_pro:s10\_sh2e1\_4Gx2}\\\hline
AOC Version & \multicolumn{5}{|l|}{19.1.0 Build 240} \\\hline
Quartus Version & \multicolumn{5}{|l|}{19.1.0 Build 240 Pro} \\\hline 
Command & \multicolumn{5}{|l|}{aoc device/snn\_pci\_v2.cl -o bin\_acl/snn\_pci\_v2.aocx }\\
& \multicolumn{5}{|l|}{-board=s10\_sh2e1\_4Gx2 -v -report -incremental}
\\\hline\hline
\multicolumn{6}{|l|}{\Large\textbf{Quartus Fit Clock Summary}} \\\hline
Frequency (MHz) & \multicolumn{5}{|l|}{300 (Kernel fmax)} \\\hline\hline
\multicolumn{6}{|l|}{\Large\textbf{Quartus Fit Resource Utilisation Summary}}\\\hline
 & \textbf{\glspl{alm}} & \textbf{FFs} & \textbf{RAMs} & \textbf{\glspl{dsp}} & \textbf{\glspl{mlab}} \\\hline
snn	& 486808.4	& 1034661 &	2486 &	1024 &	3933 \\\hline\hline
\multicolumn{6}{|l|}{\Large\textbf{Kernel Summary}} \\\hline
Kernel Name & Kernel Type & Autorun & Workgroup Size & \# Compute Units & \begin{tabular}[|c|]{@{}l@{}}Hyper-Optimised\\ Handshaking\end{tabular} \\\hline
snn & Single work-item & No & 1,1,1 & 1 & Off \\\hline\hline
\multicolumn{6}{|l|}{\Large\textbf{Estimated Resource Usage}}\\\hline
\multicolumn{1}{|c|}{\textbf{Kernel Name}} & \multicolumn{1}{|c|}{\textbf{\glspl{alut}}} & \multicolumn{1}{|c|}{\textbf{FFs}} & \multicolumn{1}{|c|}{\textbf{RAMs}} & \multicolumn{1}{|c|}{\textbf{\glspl{dsp}}} & \multicolumn{1}{|c|}{\textbf{\glspl{mlab}}} \\\hline
\textbf{snn} & 530436 & 844793 & 2868 & 976 & 3844 \\\hline
\textbf{Global Interconnect} & 10629 & 16485 & 61 &	0 & 0 \\\hline
\textbf{Board Interface} & 13132 & 20030 & 112 & 0 & 0 \\\hline
\textbf{System description ROM} & 2 & 71 & 2 & 0 & 0 \\\hline 
\textbf{Total} & 554199 (30\%) & 881379 (24\%) & 3043 (26\%) & 976 (17\%) & 3844 \\\hline
\textbf{Available} & 1866240 & 3732480 & 11721 & 5760 & 0 \\\hline
\multicolumn{6}{|l|}{\Large\textbf{Compile Warnings}} \\\hline
\multicolumn{6}{|l|}{None}\\\hline\hline
\end{tabular}}.
\end{table}

Again, from the analysis of table~\ref{tab:resources_usage_v2} it can be seen that the estimated resource utilisation is higher than the actual resource utilisation as a consequence of the Intel Quartus's optimisations during the synthesis and routing phases.

The \gls{ndk} v2 consumes 1448.7 \glspl{alm} less, 25423 FFs less, 2 RAMs less and the same number of \glspl{dsp} as the \gls{ndk} v1. Nevertheless, the \gls{ndk} v2 kernel max frequency is 300MHz, while the \gls{ndk} v1 kernel max frequency is 306.25 MHz. The \gls{ndk} v2 enables the saving of less than 1\% of resources and introduces a 2\% increase in latency.

To ensure the best use of \gls{fpga} resources, the coefficient N should always be a multiple of $2n$. For example, the resources' usage of $N=48$ is equivalent to $N=64$. Moreover, both \glspl{ndk} had failed to compile when $N=32$ because there was not enough \glspl{alut}. The design required more \glspl{alut} than those available on the device, violating the compilation rules because the design would not fit on the device.

\subsection{Speed performance} \label{Ch5.3.2:performance}

Table \ref{tab:cdnet2012_speed_results} displays the speed results obtained for the four algorithms tested against the \gls{cdnet2012}.
\begin{table}[H] \caption{\gls{cdnet2012} speed result \strut} \label{tab:cdnet2012_speed_results}
\resizebox{14.5cm}{!}{
\begin{tabular}{|l|c|c|c|c|c|c|c|}
\hline
Category & no. imgs & height & width & NeuroHSMDv2 & NeuroHSMDv1 & HSMDv2 & HSMDv1 \\ \hline
baseline/PETS2006 & 1199 & 576 & 720 & \cellcolor[HTML]{C0C0C0}23.66 & 20.45 & 8.40 & 9.73 \\ \hline
cameraJitter/badminton & 1149 & 480 & 720 & \cellcolor[HTML]{C0C0C0}28.33 & 25.48 & 10.01 & 11.50 \\ \hline
dynamicBackground/fall & 3999 & 480 & 720 & \cellcolor[HTML]{C0C0C0}27.28 & 25.01 & 9.87 & 11.08 \\ \hline
shadow/copyMachine & 3399 & 480 & 720 & \cellcolor[HTML]{C0C0C0}28.56 & 25.86 & 10.04 & 10.98 \\ \hline
dynamicBackground/fountain01 & 1183 & 288 & 432 & 55.17 & \cellcolor[HTML]{C0C0C0}58.30 & 27.40 & 29.99 \\ \hline
dynamicBackground/fountain02 & 1498 & 288 & 432 & 56.12 & \cellcolor[HTML]{C0C0C0}58.13 & 27.64 & 30.37 \\ \hline
intermittentObjectMotion/abandonedBox & 4499 & 288 & 432 & 55.93 & \cellcolor[HTML]{C0C0C0}58.45 & 26.88 & 29.89 \\ \hline
intermittentObjectMotion/tramstop & 3199 & 288 & 432 & 55.35 & \cellcolor[HTML]{C0C0C0}58.66 & 27.05 & 29.59 \\ \hline
thermal/park & 599 & 288 & 352 & 55.72 & \cellcolor[HTML]{C0C0C0}59.96 & 28.99 & 33.02 \\ \hline
shadow/peopleInShade & 1198 & 244 & 380 & 66.85 & \cellcolor[HTML]{C0C0C0}74.06 & 36.05 & 38.74 \\ \hline
baseline/highway & 1699 & 240 & 320 & 72.48 & \cellcolor[HTML]{C0C0C0}85.70 & 46.97 & 53.00 \\ \hline
baseline/office & 2049 & 240 & 360 & 69.49 & \cellcolor[HTML]{C0C0C0}78.86 & 40.24 & 46.71 \\ \hline
baseline/pedestrians & 1098 & 240 & 360 & 69.59 & \cellcolor[HTML]{C0C0C0}78.73 & 40.18 & 47.00 \\ \hline
cameraJitter/boulevard & 2499 & 240 & 352 & 68.96 & \cellcolor[HTML]{C0C0C0}78.81 & 41.21 & 46.98 \\ \hline
cameraJitter/sidewalk & 1199 & 240 & 352 & 69.04 & \cellcolor[HTML]{C0C0C0}78.54 & 39.49 & 47.32 \\ \hline
cameraJitter/traffic & 1569 & 240 & 320 & 72.17 & \cellcolor[HTML]{C0C0C0}85.29 & 43.95 & 51.40 \\ \hline
dynamicBackground/boats & 7998 & 240 & 320 & 71.86 & \cellcolor[HTML]{C0C0C0}84.29 & 45.33 & 51.70 \\ \hline
dynamicBackground/canoe & 1188 & 240 & 320 & 71.98 & \cellcolor[HTML]{C0C0C0}83.86 & 44.15 & 52.95 \\ \hline
dynamicBackground/overpass & 2999 & 240 & 320 & 71.98 & \cellcolor[HTML]{C0C0C0}84.56 & 43.69 & 49.31 \\ \hline
intermittentObjectMotion/parking & 2499 & 240 & 320 & 72.98 & \cellcolor[HTML]{C0C0C0}85.21 & 44.68 & 48.55 \\ \hline
intermittentObjectMotion/sofa & 2749 & 240 & 320 & 73.56 & \cellcolor[HTML]{C0C0C0}86.46 & 44.37 & 47.96 \\ \hline
intermittentObjectMotion/streetLight & 3199 & 240 & 320 & 72.01 & \cellcolor[HTML]{C0C0C0}84.95 & 44.31 & 47.76 \\ \hline
intermittentObjectMotion/winterDriveway & 2499 & 240 & 320 & 72.98 & \cellcolor[HTML]{C0C0C0}85.88 & 44.72 & 49.21 \\ \hline
shadow/backdoor & 1999 & 240 & 320 & 72.06 & \cellcolor[HTML]{C0C0C0}85.77 & 44.21 & 48.87 \\ \hline
shadow/bungalows & 1699 & 240 & 360 & 68.61 & \cellcolor[HTML]{C0C0C0}78.36 & 39.80 & 42.25 \\ \hline
shadow/busStation & 1249 & 240 & 360 & 68.79 & \cellcolor[HTML]{C0C0C0}78.78 & 39.85 & 42.91 \\ \hline
shadow/cubicle & 7399 & 240 & 352 & 70.87 & \cellcolor[HTML]{C0C0C0}80.47 & 41.07 & 44.52 \\ \hline
thermal/corridor & 5399 & 240 & 320 & 73.80 & \cellcolor[HTML]{C0C0C0}87.01 & 44.58 & 48.10 \\ \hline
thermal/diningRoom & 3699 & 240 & 320 & 73.38 & \cellcolor[HTML]{C0C0C0}86.60 & 44.38 & 47.82 \\ \hline
thermal/lakeSide & 6499 & 240 & 320 & 73.91 & \cellcolor[HTML]{C0C0C0}86.80 & 45.62 & 49.64 \\ \hline
thermal/library & 4899 & 240 & 320 & 74.83 & \cellcolor[HTML]{C0C0C0}87.05 & 44.40 & 49.36 \\ \hline
\end{tabular}}
\\
Best results are highlighted using grey.
\end{table}
From Table~\ref{tab:cdnet2012_speed_results} can be seen that both the NeuroHSMDv1 and NeuroHSMDv2 have performed better than the software versions (i.e. HSMDv1 and HSMDv2). It is also apparent that the NeuroHSMDv2 performs better in images with higher resolution (i.e. $720 \times 480$ and $720 \times 576$) while the NeuroHSMDv1 in lower resolutions (i.e below $720 \times 480$). Although the HSMDv2 is always faster than the HSMDv1 (non-opimised version), the NeuroHSMDv2 is only more efficient for resolutions above $288\times 432$ and NeuroHSMDv2 is faster for lower size images. This is because \gls{fpga} optimisations require the utilisation of more resources, which might increase latency.

Overall, the NeuroHSMDv1 had an average frame rate of 71.50 \gls{fps}, NeuroHSMDv2 63.20 \gls{fps}, HSMDv1 40.26 \gls{fps}, HSMDv2 36.11 \gls{fps}. Finally, the average frame rate for processing images with the native resolution of $720 \times 480$ per algorithm is i) NeuroHSMDv2 28.06 \gls{fps}, NeuroHSMDv1 25.45 \gls{fps}, HSMDv2 11.19 \gls{fps} and HSMDv1 9.97 \gls{fps}.

The speed results obtained for the four algorithms when tested against the \gls{cdnet2014} are depicted in table~\ref{tab:cdnet2014_speed_results}
\begin{table}
\caption{\gls{cdnet2014} speed results \strut} \label{tab:cdnet2014_speed_results}
\resizebox{14.5cm}{!}{
\begin{tabular}{|l|c|c|c|c|c|c|c|}
\hline
Category & no. imgs & height & width & NeuroHSMDv2 & NeuroHSMDv1 & HSMDv2 & HSMDv1 \\ \hline
badWeather/blizzard & 6999 & 480 & 720 & \cellcolor[HTML]{C0C0C0}29.70 & 24.55 & 10.12 & 11.21 \\ \hline
badWeather/snowFall & 6499 & 480 & 720 & \cellcolor[HTML]{C0C0C0}29.61 & 24.31 & 10.16 & 11.11 \\ \hline
badWeather/wetSnow & 3499 & 540 & 720 & \cellcolor[HTML]{C0C0C0}26.54 & 21.76 & 8.93 & 10.19 \\ \hline
baseline/PETS2006 & 1199 & 576 & 720 & \cellcolor[HTML]{C0C0C0}25.04 & 20.36 & 8.52 & 9.53 \\ \hline
cameraJitter/badminton & 1149 & 480 & 720 & \cellcolor[HTML]{C0C0C0}28.58 & 25.07 & 9.80 & 11.00 \\ \hline
dynamicBackground/fall & 3999 & 480 & 720 & \cellcolor[HTML]{C0C0C0}27.48 & 24.75 & 13.90 & 10.96 \\ \hline
shadow/copyMachine & 3399 & 480 & 720 & \cellcolor[HTML]{C0C0C0}28.76 & 25.66 & 14.10 & 11.51 \\ \hline
turbulence/turbulence0 & 4999 & 480 & 720 & \cellcolor[HTML]{C0C0C0}28.59 & 25.07 & 14.26 & 11.62 \\ \hline
turbulence/turbulence1 & 3999 & 480 & 720 &  \cellcolor[HTML]{C0C0C0}28.24 & 25.21 & 14.28 &  11.35 \\ \hline
turbulence/turbulence3 & 2199 & 486 & 720 &  \cellcolor[HTML]{C0C0C0}28.72 & 25.15 & 14.07 & 11.49 \\ \hline
PTZ/continuousPan & 1699 & 480 & 704 & \cellcolor[HTML]{C0C0C0}28.49 & 23.91 & 9.88 & 10.85 \\ \hline
lowFramerate/tunnelExit\_0\_35fps & 3999 & 440 & 700 & \cellcolor[HTML]{C0C0C0}31.45 & 28.23 & 15.94 & 12.47 \\ \hline
nightVideos/fluidHighway & 1363 & 450 & 700 & \cellcolor[HTML]{C0C0C0}31.17 & 27.64 & 15.27 & 12.10 \\ \hline
turbulence/turbulence2 & 4499 & 315 & 645 &  \cellcolor[HTML]{C0C0C0}41.85 & 39.59 & 23.90 & 18.93 \\ \hline
lowFramerate/port\_0\_17fps & 2999 & 480 & 640 &  \cellcolor[HTML]{C0C0C0}31.35 & 28.17 & 15.75 & 12.39 \\ \hline
lowFramerate/tramCrossroad\_1fps & 899 & 350 & 640 & \cellcolor[HTML]{C0C0C0}39.51 & 36.73 & 21.37 & 16.71 \\ \hline
nightVideos/busyBoulvard & 2759 & 364 & 640 &  \cellcolor[HTML]{C0C0C0}38.86 & 36.00 & 20.90 & 16.43 \\ \hline
nightVideos/bridgeEntry & 2499 & 430 & 630 &  \cellcolor[HTML]{C0C0C0}34.52  & 31.85 & 17.90 & 14.20 \\ \hline
nightVideos/winterStreet & 1784 & 420 & 624 & \cellcolor[HTML]{C0C0C0}35.92 & 32.57 & 18.26 & 14.52 \\ \hline
nightVideos/streetCornerAtNight & 5199 & 245 & 595 & \cellcolor[HTML]{C0C0C0}52.86 & 51.56 & 32.83 & 25.35 \\ \hline
PTZ/twoPositionPTZCam & 2299 & 340 & 570 & \cellcolor[HTML]{C0C0C0}43.30 & 41.11 & 17.91 &     18.92 \\ \hline
PTZ/intermittentPan & 3499 & 368 & 560 &  \cellcolor[HTML]{C0C0C0}40.65 & 38.52 & 16.39 & 17.71 \\ \hline
badWeather/skating & 3899 & 360 & 540 &  \cellcolor[HTML]{C0C0C0}43.00 & 40.79 & 17.20 & 19.08 \\ \hline
nightVideos/tramStation & 2999 & 295 & 480 & \cellcolor[HTML]{C0C0C0}53.15 & 52.62 & 33.46 & 26.15 \\ \hline
dynamicBackground/fountain01 & 1183 & 288 & 432 & 55.38 & \cellcolor[HTML]{C0C0C0}57.06 & 37.95 & 30.16 \\ \hline
dynamicBackground/fountain02 & 1498 & 288 & 432 & 56.56 & \cellcolor[HTML]{C0C0C0}56.97 & 38.22 & 30.49 \\ \hline
intermittentObjectMotion/abandonedBox & 4499 & 288 & 432 & 55.64 & \cellcolor[HTML]{C0C0C0}58.03 & 38.19 & 30.26 \\ \hline
intermittentObjectMotion/tramstop & 3199 & 288 & 432 & 56.58 & \cellcolor[HTML]{C0C0C0}57.56 & 38.03 & 28.99 \\ \hline
shadow/peopleInShade & 1198 & 244 & 380 & 67.31 & \cellcolor[HTML]{C0C0C0}71.54 & 49.84 & 41.11 \\ \hline
baseline/office & 2049 & 240 & 360 & 71.72 & \cellcolor[HTML]{C0C0C0}75.86 & 37.85 & 45.39 \\ \hline
baseline/pedestrians & 1098 & 240 & 360 & 70.75 & \cellcolor[HTML]{C0C0C0}75.87 & 40.13 & 45.66 \\ \hline
shadow/bungalows & 1699 & 240 & 360 & 69.25 & \cellcolor[HTML]{C0C0C0}75.88 & 55.53 & 45.12 \\ \hline
shadow/busStation & 1249 & 240 & 360 & 69.51 & \cellcolor[HTML]{C0C0C0}74.76 & 55.41 & 46.24 \\ \hline
cameraJitter/boulevard & 2499 & 240 & 352 & 70.44 & \cellcolor[HTML]{C0C0C0}75.44 & 39.33 & 42.60 \\ \hline
cameraJitter/sidewalk & 1199 & 240 & 352 & 69.44 & \cellcolor[HTML]{C0C0C0}75.95 & 38.54 & 42.36 \\ \hline
shadow/cubicle & 7399 & 240 & 352 & 71.78 & \cellcolor[HTML]{C0C0C0}77.42 & 57.46 & 46.65 \\ \hline
thermal/park & 599 & 288 & 352 & 57.07 & \cellcolor[HTML]{C0C0C0}58.39 & 43.26 & 36.78 \\ \hline
PTZ/zoomInZoomOut & 1129 & 240 & 320 & 72.70 & \cellcolor[HTML]{C0C0C0}80.89 & 41.94 & 44.97 \\ \hline
baseline/highway & 1699 & 240 & 320 & 74.12 & \cellcolor[HTML]{C0C0C0}82.40 & 42.85 & 46.89 \\ \hline
cameraJitter/traffic & 1569 & 240 & 320 & 72.87 &  \cellcolor[HTML]{C0C0C0}81.54 & 40.97 & 45.95 \\ \hline
dynamicBackground/boats & 7998 & 240 & 320 & 73.14 & \cellcolor[HTML]{C0C0C0}82.14 & 60.35 & 47.99 \\ \hline
dynamicBackground/canoe & 1188 & 240 & 320 & 72.64 & \cellcolor[HTML]{C0C0C0}81.62 & 61.19 & 46.91 \\ \hline
dynamicBackground/overpass & 2999 & 240 & 320 & 72.85 & \cellcolor[HTML]{C0C0C0}82.26 & 61.72 & 50.00 \\ \hline
intermittentObjectMotion/parking & 2499 & 240 & 320 & 73.56 & \cellcolor[HTML]{C0C0C0}81.88 & 61.94 & 51.23 \\ \hline
intermittentObjectMotion/sofa & 2749 & 240 & 320 & 73.57 & \cellcolor[HTML]{C0C0C0}82.95 & 62.15 & 47.82 \\ \hline
intermittentObjectMotion/streetLight & 3199 & 240 & 320 & 72.77 & \cellcolor[HTML]{C0C0C0}82.48 & 62.27 & 48.13 \\ \hline
intermittentObjectMotion/winterDriveway & 2499 & 240 & 320 & 74.19 & \cellcolor[HTML]{C0C0C0}82.81 & 62.96 & 48.48 \\ \hline
lowFramerate/turnpike\_0\_5fps & 1499 & 240 & 320 &  73.10 & \cellcolor[HTML]{C0C0C0}82.38 & 60.99 &  46.52 \\ \hline
shadow/backdoor & 1999 & 240 & 320 & 73.73 & \cellcolor[HTML]{C0C0C0}83.84 & 62.82 & 51.77 \\ \hline
thermal/corridor & 5399 & 240 & 320 & 74.86 &  \cellcolor[HTML]{C0C0C0}83.65 & 62.48 & 51.11 \\ \hline
thermal/diningRoom & 3699 & 240 & 320 & 74.42 & \cellcolor[HTML]{C0C0C0}83.86 & 62.34 & 50.98 \\ \hline
thermal/lakeSide & 6499 & 240 & 320 & 74.93 & \cellcolor[HTML]{C0C0C0}83.60 & 63.23 & 51.68 \\ \hline
thermal/library & 4899 & 240 & 320 & 75.79 & \cellcolor[HTML]{C0C0C0}83.72 & 62.17 & 47.23 \\ \hline
\end{tabular}}
\\
Best results are highlighted using grey.
\end{table}

Table~\ref{tab:cdnet2014_speed_results} shows that the NeuroHSMDv1 and NeuroHSMDv2 have performed better than the software versions (i.e. HSMDv1 and HSMDv2) when tested against the \gls{cdnet2014} dataset. Once again, the NeuroHSMDv2 performs better in images with higher resolution (i.e. equal or higher than $480 \times 295$) while the NeuroHSMDv1 in lower resolutions (i.e below $480 \times 295$). Again, the NeuroHSMDv2 is only more efficient for resolutions above $288\times 432$ because of the complexity and latency introduced by the optimisation circuit.

Overall, the NeuroHSMDv1 has an average frame rate of 43.51 \gls{fps}, NeuroHSMDv2 39.94 \gls{fps}, HSMDv2 29.30 \gls{fps}, and HSMDv1 25.18 \gls{fps}.  It is essential to highlight that the \gls{cdnet2014} has more categories and image sequences, leading to different frame rates for the \gls{cdnet2012} and \gls{cdnet2014} datasets.  

The average frame rate for processing images with the native resolution of $720 \times 480$ per algorithm was i) NeuroHSMDv2 28.71 \gls{fps}, NeuroHSMDv1 24.95 \gls{fps}, HSMDv2 12.37 \gls{fps} and HSMDv1 11.25 \gls{fps}. These results are in line with the results obtained for the 4 algorithms when tested against the \gls{cdnet2012} dataset.

\subsection{Benchmark} \label{Ch5.3.3:benchmark}
Table~\ref{tab:cdnet2012_ranks} shows the results obtained after testing the 4 methods against the \gls{cdnet2012} ground-truth images using the scripts provided by Nil Goyette et al. \cite{Goyette2012}.

\begin{table}[H] \caption{\gls{cdnet2012} Overall ranks \strut} \label{tab:cdnet2012_ranks}
\resizebox{14.5cm}{!}{
\begin{tabular}{|l|>{\columncolor[gray]{0.8}}c|c|c|c|c|c|c|c|c|}
\hline
Method & $\overline{\gls{arc}}$ $\downarrow$ & \gls{re} $\uparrow$ & \gls{sp} $\uparrow$ & \gls{fpr} $\downarrow$ & \gls{fnr} $\downarrow$ & \gls{wcr} $\downarrow$ & \gls{ccr} $\uparrow$ & \gls{f1} $\uparrow$ & \gls{pr} $\uparrow$ \\ \hline\hline
HSMDv1 & \cellcolor[HTML]{C0C0C0}1 & 0.52 & 0.994 & 0.006 & 0.23 & 0.024 & 0.976 & 0.77 & 0.62 \\ \hline
HSMDv2 & \cellcolor[HTML]{C0C0C0}1 & 0.52 & 0.994 & 0.006 & 0.23 & 0.024 & 0.976 & 0.77 & 0.62 \\ \hline
NeuroHSMDv1 & \cellcolor[HTML]{C0C0C0}1 & 0.52 & 0.994 & 0.006 & 0.23 & 0.024 & 0.976 & 0.77 & 0.62 \\ \hline
NeuroHSMDv2 & \cellcolor[HTML]{C0C0C0}1 & 0.52 & 0.994 & 0.006 & 0.23 & 0.024 & 0.976 & 0.77 & 0.62 \\ \hline
\end{tabular}}
\\
$\uparrow$: the highest score is the best.\\
$\downarrow$: the lowest result is the best. \\
All the 4 methods were ranked first because no changes were made to the customised \gls{snn}.
$\gls{re}$ stands for Recall, $\gls{sp}$ Specificity, \gls{fpr} False Positive Rate, \gls{fnr} False Negative Rate, \gls{wcr} Wrong Classifications Rate, \gls{ccr} Correct Classifications Rate, \gls{pr} Precision, \gls{f1} F-score and $\overline{\gls{arc}}$ Average Ranking across all Categories.
\end{table}

From the results shown in Table~\ref{tab:cdnet2012_ranks} it is possible to infer that the results obtained with the four methods are equivalent because all the methods were ranked in first place with the same values per metric. Indexing all the algorithms in the first place was expected because the speed optimisation in version 2 of the \gls{neurohsmd} and \gls{hsmd} should not interfere with the model dynamics.

Table~\ref{tab:cdnet2014_ranks} depicts the results obtained after testing 4 methods against the \gls{cdnet2014} ground-truth images using the scripts provided by Nil Goyette et al. \cite{Goyette2012}.
\begin{table}[H] \caption{\gls{cdnet2014} Overall ranks \strut} \label{tab:cdnet2014_ranks}
\resizebox{14.5cm}{!}{
\begin{tabular}{|l|>{\columncolor[gray]{0.8}}c|c|c|c|c|c|c|c|c|}
\hline
Method & $\overline{\gls{arc}}$ $\downarrow$ & \gls{re} $\uparrow$ & \gls{sp} $\uparrow$ & \gls{fpr} $\downarrow$ & \gls{fnr} $\downarrow$ & \gls{wcr} $\downarrow$ & \gls{ccr} $\uparrow$ & \gls{f1} $\uparrow$ & \gls{pr} $\uparrow$ \\ \hline\hline
HSMDv1 & \cellcolor[HTML]{C0C0C0}1 & 0.55 & 0.993 & 0.007 & 0.35 & 0.018 & 0.982 & 0.65 & 0.60\\ \hline
HSMDv2 & \cellcolor[HTML]{C0C0C0}1 & 0.55 & 0.993 & 0.007 & 0.35 & 0.018 & 0.982 & 0.65 & 0.60 \\ \hline
NeuroHSMDv1 & \cellcolor[HTML]{C0C0C0}1 & 0.55 & 0.993 & 0.007 & 0.35 & 0.018 & 0.982 & 0.65 & 0.60 \\ \hline
NeuroHSMDv2 & \cellcolor[HTML]{C0C0C0}1 & 0.55 & 0.993 & 0.007 & 0.35 & 0.018 & 0.982 & 0.65 & 0.60 \\ \hline
\end{tabular}}
\\
$\uparrow$: the highest score is the best.\\
$\downarrow$: the lowest result is the best. \\
All the 4 methods were ranked first because no changes were made to the customised \gls{snn}.
$\gls{re}$ stands for Recall, $\gls{sp}$ Specificity, \gls{fpr} False Positive Rate, \gls{fnr} False Negative Rate, \gls{wcr} Wrong Classifications Rate, \gls{ccr} Correct Classifications Rate, \gls{pr} Precision, \gls{f1} F-score and $\overline{\gls{arc}}$ Average Ranking across all Categories.
\end{table}

From the results shown in Table~\ref{tab:cdnet2014_ranks} it is possible to infer that the results obtained with the four methods are probably the same because the four methods were ranked, again, in first place with the same values per metric. These results are important because it is possible to infer that there has been no degradation in accuracy as a consequence of the hardware acceleration.

\section{Discussion}\label{Ch5.4:discussion}

Two bio-inspired \gls{neurohsmd} have been proposed to accelerate the \gls{hsmd} algorithm that was discussed in Chapter\ref{Ch4:object_motion_detection_cells}. The NeuroHSMDv1 and NeuroHSMDv2 (speed optimisation) were tested against the \gls{cdnet2012} and \gls{cdnet2014} datasets. The NeuroHSMDv1 has lower latency when processing images with resolutions equal to or greater than $480 \times 295$. The NeuronHSMDv2 (speed optimisation) has a lower latency when processing images with resolutions smaller than $480 \times 295$. To ensure a fair comparison between the software and hardware implementations, two \gls{hsmd} versions were used (HSMDv1 is the same algorithm described in Chapter \ref{Ch4:object_motion_detection_cells} and HSMDv2 with speed optimisation).

The HSMDv1, HSMDv2, NeuroHSMDv1 and NeuroHSMDv2 were all tested on the same computer equipped with a quad-core Intel(R) Core(TM) i7-4770 \gls{cpu} @ 3.40GHz, 16GB of DDR3 @ 1600 MHz and 1TB of \gls{hdd}. The average frame rate for processing images with the native resolution of $720 \times 480$ per algorithm was:\\

\textbf{\gls{cdnet2012}:}\\

i) NeuroHSMDv2 28.06 \gls{fps}, NeuroHSMDv1 25.45 \gls{fps}, HSMDv2 11.19 \gls{fps} and HSMDv1 9.97 \gls{fps}\\

\textbf{\gls{cdnet2014}:}\\

i) NeuroHSMDv2 28.71 \gls{fps}, NeuroHSMDv1 24.95 \gls{fps}, HSMDv2 12.37 \gls{fps} and HSMDv1 11.25 \gls{fps}\\

The four methods were also tested against the ground-truth images available in the \gls{cdnet2012} and \gls{cdnet2014} datasets using the eight metrics, which were used to assess and compare the quality of the \gls{hsmd} algorithm. The four methods obtained the same values for all the metrics and were all ranked first. The first place acquired by the four methods is an indication that there was no degradation in the hardware acceleration.\\

Future work includes optimising the \gls{hsmd} algorithm to detect and track motion in challenging scenarios (e.g. low frame rate, dynamic background, and camera jitter) and an investigation to verify if the \gls{snn} improves the remaining methods' output. It is also planned to implement and evaluate more complex retinal cells (such as predictive cells) using \glspl{snn}, as well as assess the effectiveness of \gls{bs} algorithms in optimising and accelerating such complex retinal cells using \gls{fpga} technology.\\

%% file: Chapter6/chapter6.tex
%%%%%%%%%%%%%%%%%%%%%%%%%%%%%%%%%%%%%%%%%%%%%%%%%%%%%%%%%%%%%%%%%%%%%%%%%%%%%%%%
%2345678901234567890123456789012345678901234567890123456789012345678901234567890
%        1         2         3         4         5         6         7         8
% THESIS CONCLUSIONS
\def\baselinestretch{1}
\chapter{Discussion and Future work} 
\label{Ch6:conclusions}

\def\baselinestretch{1.0}

\section{Main contributions} \label{Ch6.1:results}
Retinal cells are highly efficient in performing primary steps in processing of natural images. The retina performs visual tasks such as object movement sensitivity, looming, fast response to fast stimuli, and even prediction.

The first goal of the PhD research programme (see chapter \ref{Ch2:lr}) was to investigate and replicate the basic functionalities of \glspl{dsgc} (such as detecting horizontal and vertical movements and more generic object motions). This PhD research programme's second objective was to explore the use of \glspl{snn} for implementing bio-inspired object motion detectors inspired by the \gls{oms-gc}. The third and final objective was to optimise the object motion detector, using \gls{fpga} technology to improve the latency by accelerating \gls{snn} and reducing the dynamic power consumption associated with high-frequency clocks.\\
    
The \gls{mhsnn} architecture presented in Chapter~\ref{Ch3:mhsnn} contributes to the first and second objectives of this PhD research programme. \gls{mhsnn}'s main contributions are:

\begin{enumerate}
    \item Layer 1 edge detection
    \item Extraction of direction motion features in Layer 2;
    \item Extraction of movement features in Layer 3;
    \item Movement detection in Layer 4.
\end{enumerate}

Therefore, the \gls{mhsnn} outputs reflect object motion detection in vertebrate retinas. The most straightforward behaviour shown by \glspl{dsgc}, horizontal and vertical movement detection, was modelled. Experiments were carried out using a semisynthetic dataset of a black cylinder performing leftwards, rightwards, downwards and upwards movement. A \gls{codd} algorithm that combined the seven \gls{bs} available in the \gls{opencv} library was implemented to benchmark against the \gls{mhsnn}. The \gls{mhsnn} was ranked first in terms of the \gls{pcc} and exhibited the lowest \gls{pwc} detecting the motion direction when tested against the semisynthetic dataset. Although the \gls{mhsnn} could be adapted for modelling other types of ganglion cells, such as looming (sensitive to approach and recede movements), fast-response (sensitive to fast movements) and predictive (predicting movement trajectories) cells, the \gls{mhsnn} is not a scalable architecture because of the required number of neurons and synapses (above 17000 neurons and 173700 synapses for processing $40\times40$ pixels image, see Chapter~\ref{Ch3:mhsnn}). Nevertheless, the number of neurons and synapses can be substantially decreased by combining \gls{snn} with existing \gls{bs} algorithms.\\

The \gls{mhsnn} architecture lead to the development of a bio-inspired \gls{hsmd} discussed in Chapter \ref{Ch4:object_motion_detection_cells}, to detect object motion and assess against the \gls{cdnet2012} and \gls{cdnet2014} datasets. The \gls{hsmd} contributes to the first and second objectives of this PhD research programme. These incorporate video sequences of many moving objects under various challenging environmental conditions and are widely used for benchmarking background subtraction algorithms. The \gls{cdnet2012} is composed of five categories of movements, and the \gls{cdnet2014} augments the initial five to eleven categories of movements. Eight metrics, utilised as standard in the change detection datasets, were used to assess and compare the quality of the \gls{hsmd} algorithm. The \gls{hsmd} algorithm performed overall best in both the \gls{cdnet2012} and \gls{cdnet2014} while performing better than all the tested \gls{bs} algorithms in the intermittent object motion, night videos, thermal and turbulence categories, it performs second-best in the bad weather category and the third-best in the baseline and shadow categories. The comparatively good results are a consequence of using the \gls{snn} for emulating the basic functionality of \gls{oms-gc}, which improves the sensitivity of the \gls{hsmd} to object motion. The \gls{hsmd} is also the first hybrid \gls{snn} algorithm capable of processing video/image sequences in near real-time (i.e. 720$\times$480@13.82fps [\gls{cdnet2014}] and 720$\times$480@13.92fps [\gls{cdnet2012}]). \\

Finally, in Chapter~\ref{Ch5:neuromorphic_object_motion_detector}, two bio-inspired \gls{neurohsmd} have been proposed to accelerate the \gls{hsmd} algorithm that was discussed in Chapter\ref{Ch4:object_motion_detection_cells}. The \gls{neurohsmd} algorithm contributes to all the objectives of this PhD research programme. The NeuroHSMDv1 and NeuroHSMDv2 (speed optimisation) were tested against the \gls{cdnet2012} and \gls{cdnet2014} datasets. The NeuroHSMDv2 (speed optimisation) has lower latency when processing higher resolution images (equal to or greater than $480 \times 295$). The NeuronHSMDv1 has a lower latency when processing images with lower resolutions (smaller or equal to $480 \times 295$). Both versions of the \gls{neurohsmd} algorithms have produced the same results as the original \gls{hsmd} algorithm reported in Chapter \ref{Ch4:object_motion_detection_cells}. The average frame rate for processing images with the native resolution of $720 \times 480$ per algorithm was\\
\textbf{\gls{cdnet2012}:}\\
i) NeuroHSMDv2 28.06 fps, NeuroHSMDv1 25.45 fps, HSMDv2 11.19 fps and HSMDv1 9.97 fps\\
\textbf{\gls{cdnet2014}:}\\
i) NeuroHSMDv2 28.71 fps, NeuroHSMDv1 24.95 fps, HSMDv2 12.37 fps and HSMDv1 11.25 fps.\\

\section{Future work} \label{Ch6.2:discussion}
Future work on the \gls{mhsnn} includes replacing the first two layers by \gls{bs} algorithms to reduce the number of neurons and synapses and test it on natural images. It is also planned to adapt the \gls{mhsnn} to detect approach and receding movements. At the same time, it is also planned to benchmark the \gls{mhsnn}'s \gls{snn} inference on \glspl{cpu}, \glspl{fpga} and \glspl{gpu} and compare the power consumption profiles of each solution. The power consumption and the \gls{pcc} will be used to justify the technology (or technologies) to be utilised to design intelligent cameras to run variants of the \gls{mhsnn}. 

\gls{hsmd} future work includes optimising the \gls{hsmd} algorithm to detect and track motion in challenging scenarios (e.g. low frame rate, dynamic background, and camera jitter). It is further planned to investigate if the \gls{snn} used in the \gls{hsmd} also improves other \gls{opencv} \gls{bs} methods. Furthermore, the author will also test if the \gls{snn} can enhance other \gls{bs} algorithms available on the \gls{opencv} library. Simultaneously, it is also planned to benchmark the \gls{hsmd}'s \gls{snn} inference on \glspl{cpu}, \glspl{fpga} and \glspl{gpu} and compare the different power consumption profiles. The same methodology that was used to develop the \gls{hsmd} algorithm will also be used to model more complex retinal cells such as the fast-response and predictive cells \cite{Gollisch2010}.\\

Future work for \gls{neurohsmd} includes optimising the custom \gls{snn} to fit into low-cost \gls{ps} and \gls{fpga} on the same chip (\gls{ps}-\gls{fpga}), which could deliver the next-generation of intelligent cameras. The author also aims to convert the \gls{opencl} kernels into \gls{cuda} kernels and evaluate the \gls{neurohsmd} performance on a processor system and \gls{gpu} on the same chip (\gls{ps}-\gls{gpu}). This will lead to development of new intelligent motion detectors which can be used in different fields from crowd monitoring to \gls{adas}. The knowledge gathered in this PhD research programme about porting complex \gls{snn} into \gls{fpga}s using \gls{opencl} will also be used for porting complex retinal cells (e.g. fast-response and predictive \glspl{gc}) and other parallelisable algorithms that can be accelerated using dedicated hardware.